\documentclass[isre,nonblindrev]{informs3}

\OneAndAHalfSpacedXI

\usepackage{natbib}

 \bibpunct[, ]{(}{)}{,}{a}{}{,}

 \def\bibfont{\small}

 \def\bibsep{\smallskipamount}

\TheoremsNumberedThrough

\EquationsNumberedThrough

\MANUSCRIPTNO{1}

\usepackage[ruled]{algorithm2e}

\newenvironment{proof*}[1][Proof]{\noindent\textbf{#1.} }{\hfill$\square$}

\usepackage{todonotes}
\usepackage{tabularx}
\usepackage{threeparttable}
\usepackage{xcolor}
\usepackage{tcolorbox}
\usepackage{colortbl}
\usepackage{float}
\usepackage[colorlinks=true, linkcolor=black, citecolor=black, urlcolor=black]{hyperref}
\usepackage{booktabs}
\usepackage{enumitem}

\tcbuselibrary{most}

\usepackage{dsfont}
\usepackage{setspace}

\usepackage{amsfonts}

\usepackage{bibunits}

\defaultbibliography{ref}
\defaultbibliographystyle{informs2014}

\begin{document}

\RUNAUTHOR{Cheng, Mayya, and Sedoc}

\RUNTITLE{Toward Robust Reproducibility in LLM Annotation}

\TITLE{To Err Is Human; To Annotate, SILICON? Toward Robust Reproducibility in LLM Annotation}

\ARTICLEAUTHORS{
\AUTHOR{Xiang Cheng}
\AFF{Robert H. Smith School of Business, University of Maryland\\xccheng@umd.edu}
\AUTHOR{Raveesh Mayya, João Sedoc}
\AFF{Leonard N. Stern School of Business, New York University\\\{raveesh, jsedoc\}@stern.nyu.edu}
\vspace{0.35cm}
\AUTHOR{Updated: 2 April 2026. \\(For the Latest Version: \href{https://arxiv.org/abs/2412.14461}{\color{blue}{Click Here}})}
}

\ABSTRACT{
\begin{center} \textbf{ABSTRACT}\\ \end{center}
Unstructured text data annotation is foundational to management research. LLMs offer a cost-effective and scalable alternative to human annotation, but they introduce a novel challenge: the annotator itself can be retired. Proprietary models undergo regular deprecation cycles, threatening long-term reproducibility. Hence, the ability to reproduce annotation results when the original model becomes unavailable, i.e., robust reproducibility, is a central methodological challenge for LLM-based annotation.
Achieving robust reproducibility requires first controlling measurement error. We develop an analytical framework that decomposes measurement error into four sources: guideline-induced error from inconsistent annotation criteria, baseline-induced error from unreliable human references, prompt-induced error from suboptimal meta-instruction, and model-induced error from architectural differences across LLMs.
We develop the SILICON workflow that instantiates the analytical framework, prescribing targeted interventions at each error source. Empirical validation across nine management research tasks confirms that these interventions reduce measurement error, and simulations show that the resulting error reduction yields more accurate downstream statistical estimates.
With measurement error controlled, we address two further aspects of robust reproducibility. First, we propose a regression-based methodology to establish backup open-weight models, which are permanently accessible. Every tested task has at least one open-weight model with no statistically detectable performance difference.
Second, we quantify the upper bound of annotation quality attainable from the current set of available models by proposing a routing procedure that selectively sends low-confidence items to auxiliary models, revealing when model aggregation improves performance and when that may adversely affect labeling quality.
An open-source toolkit will be released upon publication.

}


\thispagestyle{empty}
\maketitle
\begin{bibunit}
\section{Introduction}\label{intro}
Annotation of unstructured text data, using either humans or artificial intelligence tools, is an integral part of management research \citep{ghose_estimating_2011, humphreys_automated_2018, loughran_measuring_2014}.
In particular, machine learning and Natural Language Processing tools are widely employed to extract insights from unstructured text data and explain the mechanisms underlying business phenomena via sentiment and emotional content analysis \citep{chakraborty_attribute_2022, liu_large-scale_2019, melumad_selectively_2019, yang_understanding_2019}, product attribute classification \citep{banerjee_interacting_2021, kwark_spillover_2021}, advertisement analysis \citep{lee_advertising_2018, shi_hype_2022}, managerial response analysis \citep{deng_managerial_2023}, consumer review analysis \citep{hong_just_2021,mayya_who_2021}, and product categorization \citep{lee_how_2021}.

Large Language Models (LLMs) promise to transform text data annotation, offering improved time-efficiency and cost-effectiveness compared to human annotators. Researchers are exploring LLMs for text annotation tasks \citep{huang_algorithm-enabled_2024,pangakis_automated_2023, yeverechyahu_impact_2024, zhang_comprehensive_2024}, with studies showing LLMs can match or exceed crowdsourced worker performance \citep{gilardi_chatgpt_2023}. However, this rapid adoption proceeds without addressing a novel challenge absent in traditional annotation: the annotator itself can be retired. Proprietary LLMs undergo regular deprecation cycles or version updates that can drastically alter model behavior \citep{chen_how_2023}.
Human annotators may also become unavailable, but they can in principle be re-recruited or replaced with comparably trained substitutes.
A deprecated model, by contrast, is irrecoverably inaccessible, and no successor is guaranteed to replicate its behavior.
Building on the \cite{national_academies_of_sciences_engineering_and_medicine_reproducibility_2019} framework for reproducibility and replicability in science, and \cite{roper_testing_2022} distinction between reproducibility and robustness, we define robust reproducibility in the LLM annotation context as the ability to reproduce annotation results when the original model becomes unavailable. This is a key methodological challenge for LLM-based annotation in research.

Despite growing research interest in LLMs for text annotation tasks \citep{tan_large_2024}, existing studies do not address robust reproducibility. Extant work validates LLM outputs against human-generated labels \citep{pangakis_automated_2023} and offers general implementation guidelines \citep{tornberg_best_2024}, but these efforts examine isolated components of the annotation process. Even structured approaches like \cite{de_kok_chatgpt_2025} and \cite{carlson_use_2025} overlook reproducibility concerns as proprietary models undergo retirement cycles.
This gap motivates our primary research question: \textit{How can researchers achieve and evaluate robust reproducibility in LLM-based text annotation?}

Determining whether a model is robustly reproducible requires first correctly identifying the model's performance. If the performance metric is contaminated by measurement error \citep{yang_mind_2018}, then the performance signal is not credible, and any reproducibility comparison is unreliable.
The first step toward robust reproducibility is therefore a rigorous answer to: \textit{How to minimize measurement error in LLM annotation?} Answering this question requires a framework that decomposes measurement error by explicitly identifying where the errors come from, since every annotation configuration, i.e., guidelines, human baselines, prompts, and models, introduces its own sources of error. This concern has long existed in human annotation but is magnified with LLMs. Unlike human annotators who can seek clarification and apply judgment beyond written instructions, LLMs interpret only what is explicitly stated, not necessarily what the researcher intended \citep{pryzant_automatic_2023}.
While prescriptive and detailed instructions \citep{rottger_two_2022} can address this asymmetry by leveraging the LLM's capacity to absorb comprehensive detail without cognitive overload, no existing work systematically decomposes and controls these sources of error. Without such a decomposition, observed differences under a substitute model cannot be conclusively attributed to the model itself; they could equally reflect uncontrolled errors from guidelines, baselines, or prompts \citep{lee_common_2023}.

\begin{figure}[h]
    \centering
    \caption{Methodological Roadmap Toward Robust Reproducibility in LLM Annotation}
    \includegraphics[width=0.95\linewidth]{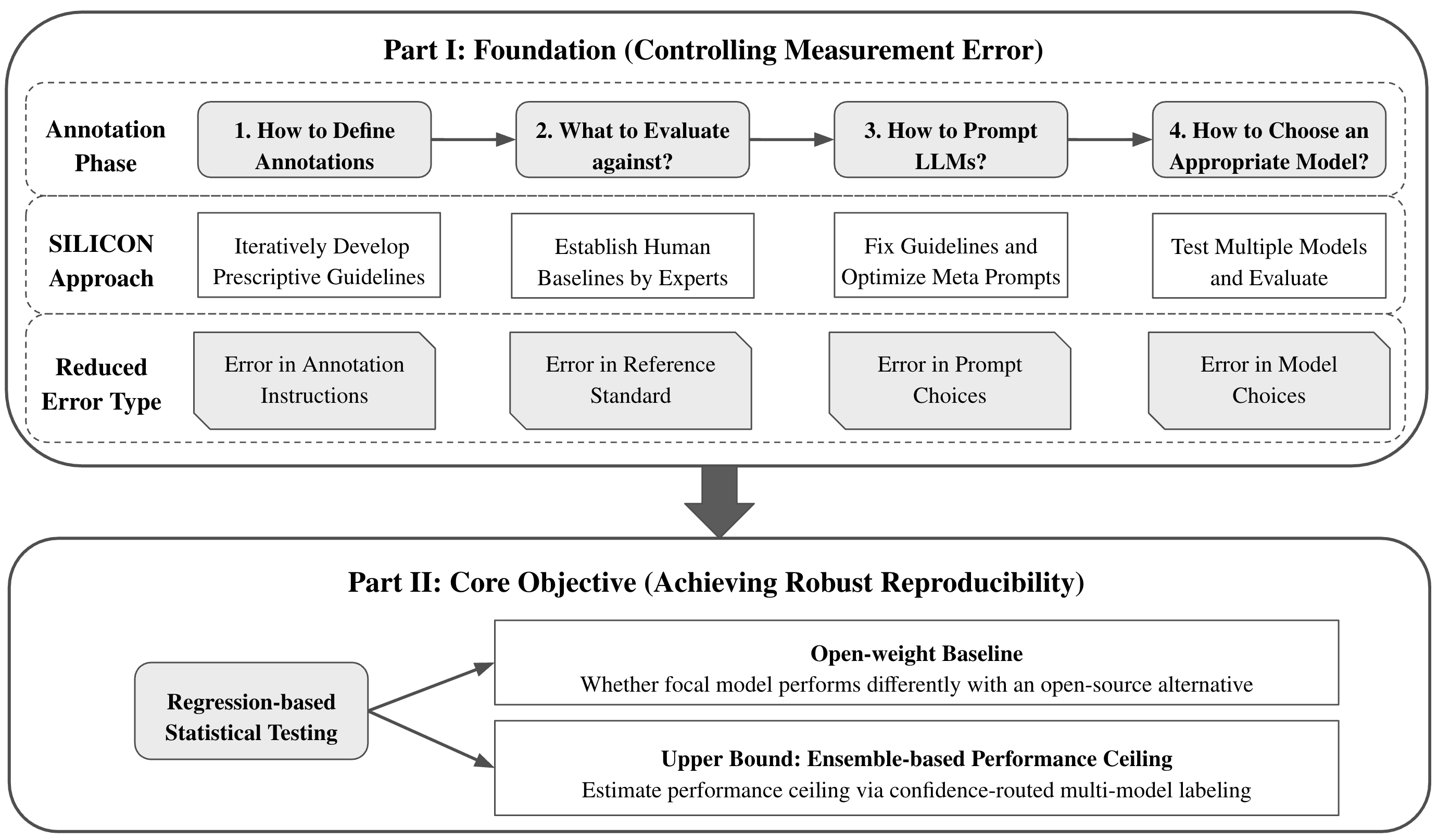}
    \label{fig:how_to_graph}
\end{figure}

To address robust reproducibility in LLM annotation, we first present an analytical framework that decomposes measurement error into four sources: (1) guideline-induced error from inconsistent annotation criteria, (2) baseline-induced error from unreliable human reference standards, (3) prompt-induced error from suboptimal meta-instruction formatting, and (4) model-induced error from architectural differences across LLMs.
We then instantiate this framework in the SILICON workflow (\textbf{S}ystematic \textbf{I}nference with \textbf{L}LMs for \textbf{I}nformation \textbf{C}lassificati\textbf{o}n and \textbf{N}otation).
Part I of Figure~\ref{fig:how_to_graph} visualizes how to achieve measurement error reduction as a prerequisite toward robust reproducibility.

We empirically validate the SILICON workflow by testing multiple state-of-the-art LLMs across nine representative text classification tasks from management research, spanning binary, multi-label, and multi-class classification.
Our empirical analysis reveals that (1) iteratively refined guidelines reduce measurement error by 0.1 to 0.5 agreement level (measured by Cohen's Kappa between LLM and human baselines) compared to one-shot guidelines across six tasks for which we developed the guidelines; (2) expert-generated baselines exhibit a 0.1 to 0.4 point higher inter-annotator agreement than crowd-generated baselines, indicating higher reliability of expert baselines; (3) system prompt placement generally reduces prompt-induced error relative to user prompt placement, though the improvement does not reach statistical significance; and (4) model performance varies substantially across tasks; no single model dominates all contexts, which necessitates task-specific selection to minimize model-induced error. The simulation results further confirm that the error reduction improves downstream statistical inference.

Once measurement error sources are controlled, we address two further aspects of robust reproducibility.
First, we propose a regression-based methodology to establish backup open-weight models, which are permanently accessible in case of model retirement.
When the focal model is closed-weight, we compare it against available open-weight models for the same task. If the performance gap is not statistically distinguishable, then the task is reproducible using open infrastructure; a large gap indicates dependence on a model that may be deprecated or inaccessible over time.
Second, we quantify the upper bound of annotation quality attainable from the current set of available models.
Drawing on the human annotation literature, where generating multiple labels per instance is standard practice for improving annotation quality \citep{ipeirotis_repeated_2014}, we construct multi-model ensembles by augmenting the focal model with the second- and third-best models. We estimate item-level model confidence using the First-Second Distance (FSD) metric \citep{lyu_calibrating_2025} and propose a routing procedure that selectively sends low-confidence items to auxiliary models. Sweeping the routing threshold from focal-model-only to full integration reveals whether aggregation is helpful for a given task and, if so, which threshold yields the greatest improvement.
Part II of Figure~\ref{fig:how_to_graph} illustrates how these two aspects jointly characterize robust reproducibility.

Empirical results reveal that every task whose best-performing model is closed has at least one open-weight alternative with no statistically detectable performance difference.
Furthermore, the value of multi-model aggregation is task-dependent: in at least one case, auxiliary models provide no benefit whatsoever. The routing curves also inform a practical tradeoff between performance and cost. When aggregate performance increases sharply up to a certain threshold and then improves only marginally, researchers must weigh whether the additional cost of multi-model labeling is justified by the relatively small gain in overall performance.

Our work contributes to management research by providing the first systematic treatment of robust reproducibility in LLM-based annotation. We make three contributions. First, our analytical framework decomposes measurement error into four sources, establishing the prerequisite for credible performance assessment: without controlling these sources, any reproducibility comparison is unreliable. Second, we operationalize robust reproducibility as a lack of statistical difference via regression-based testing, with explicit consideration of the target model---whether an open-weight baseline or an upper-bound ensemble comparison. Third, we propose a regression-based methodology to establish backup open-weight models and a routing procedure to quantify the upper bound of performance from the current set of available models. The routing analysis reveals that model aggregation does not always improve performance and provides researchers with a diagnostic for determining whether and at what confidence threshold to deploy additional models, balancing performance gains against labeling cost. Whereas existing approaches focus on isolated performance benchmarking, SILICON ensures annotation results remain valid and reproducible as models evolve. Practically, SILICON is resource-efficient: by evaluating model performance and measurement error on representative samples, researchers can achieve reliable annotation quality assessments with modest sample sizes (typically 100 to 200 documents). Although our empirical validation focuses on classification tasks, these findings remain applicable as LLMs evolve because they address fundamental sources of measurement error rather than model-specific implementations. To facilitate adoption, we will release an open-source toolkit that implements SILICON for practice and research.

\section{Conceptual Background and Related Literature}

Our work contributes to the design science field within Information Systems in the line of \cite{gregor_positioning_2013}. The main theoretical construct of interest is robust reproducibility,  and in order to achieve it, we propose a methodological framework---SILICON---as a design artifact for decomposing and reducing measurement error. In this section, we begin with background on robust reproducibility, then move to the annotation process, and finally to LLMs for text annotation.

\subsection{Robust Reproducibility}\label{sec:robust_repro_lit}

We introduced robust reproducibility as the ability to reproduce annotation results when the original model becomes unavailable; here, we situate this concept within the broader reproducibility literature.
Reproducibility is a cornerstone of scientific inquiry: independent researchers should be able to obtain the same results using the same methodology \citep{fisar_reproducibility_2024}. For most research tasks, reproducibility relies on transparent reporting of methods, data availability, and the stability of analytical tools. Statistical methods such as regression or factor analysis are mathematically stable once defined. Human annotators, while variable, belong to a stable class of agents whose judgment can be approximated by training successors with the same guidelines.

Reproducibility concerns become qualitatively different, however, when machines perform tasks previously reserved for humans. Unlike statistical methods that are immutably defined or human annotators who can be trained as successors, LLMs are versioned commercial products subject to deprecation without notice \citep{palmer_using_2023}, and their behaviors can vary drastically between adjacent versions \citep{chen_how_2023}. A model that produces strong annotation results today may be retired, and its behavioral characteristics cannot be reproduced by any successor. This makes \textit{robust reproducibility}, the ability to reproduce annotations when the original model becomes unavailable \citep{roper_testing_2022}, a distinct methodological challenge. Achieving it requires first controlling measurement error; without this foundation, any discrepancy between models is uninterpretable.

A critical follow-up question is: \textit{reproducible with respect to what target?} Two natural targets arise. First, a permanently accessible open-weight model provides a conservative baseline when the focal model is closed; if the focal model is statistically indistinguishable from this target, researchers obtain a baseline reproducibility guarantee that survives model retirement.
Second, within the robust reproducibility framework, a natural complement to the open-weight baseline is an upper bound on the focal model's achievable performance.
Multi-model labeling, in which additional models independently label the same items and their outputs are aggregated, provides a natural mechanism for constructing this upper bound by asking whether stronger labeling support can push performance beyond what the focal model achieves alone. Literature on human annotation demonstrates that generating multiple labels per instance can improve annotation quality when human annotator confidence is low \citep{ipeirotis_repeated_2014}. Translating this idea to multi-model labeling raises two questions: does augmenting the focal model with labels from complementary models actually improve aggregate performance, and can labels be introduced selectively to achieve the best outcome? We address both questions in the proposed robust reproducibility protocol.

\subsection{Science of Annotation} \label{sec:prescriptive_guidelines}
In management research using text analysis, the effectiveness of text annotation processes fundamentally depends on establishing consistent interpretation across annotators, thereby minimizing measurement error that stems from heterogeneous mental models. Annotation tasks face challenges when annotator perspectives, influenced by their socio-economic backgrounds, lead to divergent interpretations.
For example, toxicity detection labels can vary significantly based on annotators' racial identities \citep{sap_risk_2019}. Moreover, when researchers themselves serve as annotators, they risk introducing confirmation bias by unconsciously labeling data in ways that support their hypotheses \citep{nickerson_confirmation_1998}; delegating annotation to a team of independent research assistants mitigates this threat. Researchers have addressed this inherent subjectivity by choosing between \textit{descriptive} approaches that capture diverse perspectives or \textit{prescriptive} approaches that enforce consistent interpretation \citep{rottger_two_2022}. While neither approach is inherently superior, prescriptive approaches are particularly valuable in management research, where researchers aim to measure constructs using precise, theoretically grounded, and unified definitions.

The prescriptive approach becomes especially critical when transitioning from human to LLM annotators due to fundamental differences in how the two process information. Human annotators can seek clarification during ambiguous cases and rely on contextual understanding beyond the written guidelines \citep{salehi_communicating_2017}. In contrast, LLMs operate exclusively on textual input without the ability to request clarification or access implicit domain knowledge \citep{pryzant_automatic_2023}. This creates a critical asymmetry: whereas human annotation benefits from concise guidelines to maintain attention and reduce cognitive load \citep{snow_cheap_2008}, LLMs perform optimally with detailed, exhaustive instructions that anticipate edge cases and provide explicit decision rules. Since LLMs cannot seek clarification during the annotation process, a prescriptive approach becomes critical to prevent any guideline ambiguity from translating into increased measurement error.

Researchers have established that the prescriptive approach to any guideline requires an ``iterative'' refinement.
Iterative guideline refinement through multiple rounds of documented discussions among annotators can achieve substantial agreement on the intended interpretation, since discussions resolve edge-cases \citep{lee_common_2023, oortwijn_interrater_2021}.
This iterative refinement process is even more critical when LLMs serve as annotators because, unlike human annotators, models cannot adapt their interpretation mid-task through discussion.

While the principles for human annotation are established in the literature, their application to LLM-based annotation introduces additional challenges, including model selection and the need for carefully crafted ``meta'' system prompts that help models understand tasks and return results in structured, usable formats.

\subsection{LLMs for Text Annotation}
Given that LLMs are trained on vast volumes of human-generated text, researchers have explored their capacity to replicate or match human judgment across various domains (e.g., \citealt{chen_large_2024, goli_frontiers_2024, ye_lola_2024}). Early studies document that LLMs could replace human coders for psychological construct detection \citep{rathje_gpt_2024} or policy interpretation \citep{leek_introducing_2024}, or act as experimental subjects \citep{doshi_generative_2024, xie_can_2024}.

For text annotation specifically, where the objective is to produce consistent labels for clearly specified tasks rather than to mimic human cognition, LLMs have demonstrated strong potential \citep{tan_large_2024}. For instance, \citet{gilardi_chatgpt_2023} found that zero-shot ChatGPT labeling achieved higher accuracy in political science classification tasks than crowd workers. \citet{mousavi_lexicons_2024} demonstrated that GPT-4 matched or surpassed traditional lexicon-based, supervised learning, and embedding approaches in psychometric text analysis.

Indeed, LLMs can serve as cost-effective alternatives to human annotation for well-defined classification tasks.
Yet this promise has also prompted scrutiny of when and how LLMs should be deployed as annotators, and several studies now propose systematic methods to guide this decision. \citet{calderon_alternative_2025} propose a statistical procedure that provides a principled determination of whether an LLM can justifiably replace human annotators for a given task.
\citet{pangakis_automated_2023} highlight the importance of validating LLM annotation results and propose several model selection procedures. \citet{tornberg_best_2024} offers implementation guidelines focusing on prompt engineering and evaluation metrics. \citet{de_kok_chatgpt_2025} presents a framework for GPT-3.5 and GPT-4 deployment for accounting annotation tasks. \citet{carlson_use_2025} develop a five-stage implementation framework for LLM annotation with a focus on comparisons across prompts and models.

Our work builds on the extant literature while addressing a gap in prior studies: robust reproducibility. Beyond establishing that LLMs can match human performance, we ask why certain protocols produce reliable results and whether those results survive model change. The formal decomposition of measurement error that we propose illuminates the mechanisms underlying performance differences documented in prior work, and provides the analytical basis for assessing whether annotation results remain valid after the original model is retired.

\section{Methodology}\label{sec:methodology}

\subsection{Decomposing Measurement Error in LLM Annotation}\label{sec:sources}

Using LLMs to annotate data poses a fundamental measurement challenge: the ground truth labels are unobservable. In practice, we evaluate LLM output by comparing it against human reference labels, which are themselves imperfect. As a result, improvements in observed agreement between LLM and human labels may not reflect genuine gains in the LLM's alignment with the truth. In this section, we develop an analytical framework that decomposes this measurement error and derives empirical predictions.

We describe an annotation setup as a \emph{configuration} $c=(G,H,P,M)$, where $G$ represents annotation guidelines, $H\in\{\text{experts},\ \text{crowds}\}$ indicates the human source of the reference baselines, $P$ represents the meta-prompt, and $M$ represents the LLM model.
Let $x$ be an item to annotate (e.g., a customer review). The annotation task is to assign a label to $x$ that captures a characteristic of interest to the researcher (e.g., sentiment expressed).
We denote the unobserved ground truth label as $y$, the human reference label as $r$, and the LLM label under configuration $c$ as $\hat{y}_c$.

Our goal is to find the configuration that minimizes annotation loss against the ground truth:
\begin{equation}
\min_{c}\, \mathbb{E}\Big[\mathcal{L}\big(\hat y_c,\,y\big)\Big],
\label{eq:risk-ipei}
\end{equation}
where $\mathcal{L}$ is a task-appropriate loss function. For classification with 0--1 loss, this reduces to maximizing accuracy. We approach this optimization through component-wise analysis, examining how variations in guidelines ($G$), human baselines ($H$), meta-prompts ($P$) and models ($M$) individually contribute to measurement error reduction.

To connect this objective to observable quantities, we distinguish two forms of agreement. The \textit{true agreement} measures how often the LLM matches the unobserved ground truth, $T(c)\equiv \Pr(\hat y_c=y)$, while the \textit{reference agreement} measures how often the LLM matches the human reference, $R(c)\equiv \Pr(\hat y_c=r)$.\footnote{In our analytical derivations we use accuracy (proportion of matches) because it maps cleanly to the label-noise model below. In the empirical sections we report Cohen's Kappa for interpretability. The conclusions hold for any agreement statistic that increases monotonically as task loss decreases on the pairwise local comparisons.} Since perfect annotation corresponds to $T(c)=1$, we define \emph{measurement error} as the shortfall from perfect agreement:
\begin{equation}
\varepsilon_c \;\equiv\; 1 - T(c).
\label{eq:shortfall-ipei}
\end{equation}

We introduce two definitions that serve as our central units of analysis:

{\OneAndAHalfSpacedXI
\begin{tcolorbox}[
    colback=white,
    colframe=black!75!white,
    arc=3mm,
    title={Key Definitions},
    boxrule=1pt,
    left=10pt,
    right=10pt,
    top=8pt,
    bottom=8pt
]
A \textit{contrast} is a pairwise comparison between two configurations that differ in exactly one component of $\{G,H,P,M\}$.

A \textit{reference agreement contrast} is a contrast measured on reference agreement $R$.
\end{tcolorbox}
}

Intuitively, a positive reference agreement contrast suggests one configuration is ``better.'' The central analytical question is whether such a positive contrast corresponds to a reduction in measurement error.

\subsubsection{Relation between reference agreement contrast and measurement error reduction.}
Since we can only observe reference agreement $R$ but care about true agreement $T$, we need to relate the two formally. We adopt the following assumption of symmetric $K$-class random classification noise, standard in annotation literature \citep{natarajan_learning_2013}.

\begin{assumption}[Symmetric label noise]\label{assum:sln}
Let $K\ge2$ be the number of classes and $e\in[0,1)$ the error rate of the human reference. For every item with ground truth $y\in\{1,\dots,K\}$,
\[
\Pr(r=y\mid y)=1-e,\qquad
\Pr(r=\ell,\, y\ne \ell \mid y)=\frac{e}{K-1}\quad\text{for each }\ell\ne y.
\]
\end{assumption}
Assumption \ref{assum:sln} implies that human annotators are correct with probability $1-e$, and when they err, each wrong class is equally likely.

Importantly, although the LLM label $\hat y_c$ and the reference $r$ are generated independently of each other, they may still be correlated conditional on the truth $y$. Specifically, they could err identically on the same item, not because one influenced the other, but because the item is ambiguous or the guidelines are unclear.
This conditional correlation creates a key challenge: reference agreement $R$ can improve either because the LLM gets closer to the truth, or because it starts making the \emph{same mistakes} as the human reference. To formalize this distinction, we introduce the co-labeling covariance, denoted $J(c)$, and establish its role in the following result (proof in Appendix~\ref{appendix:proof}).

\begin{lemma}\label{lemma:bias-aware}
Under Assumption~\ref{assum:sln}, for any configuration $c$,
\[
R(c)\;=\; a\,T(c)\;+\;b\;+\;J(c),
\]
where $a = (1{-}e)-\tfrac{e}{K-1}$ and $b = \tfrac{e}{K-1}$ are constants determined by the human error rate $e$ and the number of classes $K$, and
\[
J(c)\;=\; \mathbb E_{y}\!\left[\sum_{\ell=1}^K
\Big(\Pr(\hat y_c{=}\ell,\,r{=}\ell\mid y) \;-\; \Pr(\hat y_c{=}\ell\mid y)\,\Pr(r{=}\ell\mid y)\Big)\right].
\]
\end{lemma}

The lemma decomposes reference agreement into three interpretable parts. The linear term $a\,T(c)+b$ reflects what reference agreement \emph{would} be if LLM errors and human errors were conditionally independent given the truth. The co-labeling covariance $J(c)$ captures the additional agreement that arises because the LLM and the human tend to assign the same label--whether correct or incorrect--beyond what conditional independence would predict. A positive $J(c)$ is problematic when it reflects the LLM and the reference making the \emph{same errors}, inflating $R(c)$ without improving $T(c)$.

Because $a$ and $b$ depend only on $(G,H)$ but not on the prompt or model, any contrast that holds $(G,H)$ fixed yields:
\begin{equation}
R(c_1)-R(c_0) \;=\; a\big(T(c_1)-T(c_0)\big) \;+\; \big(J(c_1)-J(c_0)\big).
\label{eq:diff-id-bias}
\end{equation}
Equation~(\ref{eq:diff-id-bias}) shows that the reference agreement contrast decomposes into the truth-level improvement plus the change in co-labeling covariance.
Rearranging isolates changes in measurement error:

\begin{proposition}\label{prop:err-reduction}
Under Assumption~\ref{assum:sln}, the difference in measurement error between configurations $c_1$ and $c_0$ is:
\begin{equation}
\varepsilon_{c_1}-\varepsilon_{c_0}
\;=\; -\big(T(c_1)-T(c_0)\big)
\;=\; -\tfrac{1}{a}\Big(\big(R(c_1)-R(c_0)\big) - \big(J(c_1)-J(c_0)\big)\Big).
\label{eq:err-reduction}
\end{equation}
Hence, holding the human reference fixed, a positive reference agreement contrast reduces measurement error if it exceeds the increase in co-labeling covariance.
\end{proposition}

Proposition~\ref{prop:err-reduction} establishes that observed improvements translate to measurement error reduction when the change in co-labeling covariance is sufficiently small. Verifying this condition requires empirical validation on two fronts. First, we compare expert versus crowd baselines to determine which human procedure produces more stable identification (Section~\ref{sec:expert_vs_crowd}). Second, when reference agreement is low, we establish empirical cutoffs beyond which reference agreement contrasts reliably signal error reduction. Our analysis in Section~\ref{sec:expert_vs_crowd} identifies a conservative threshold ($\kappa \geq 0.5$) where $J$-induced confounding becomes minimal, providing practical guidance for when LLM annotation achieves sufficient quality for research applications.

\subsubsection{Designing contrasts.}
Each component of $c=(G,H,P,M)$ requires a different comparison protocol: changing guidelines ($G$) or baselines ($H$) affects the human reference itself, whereas changing prompts ($P$) or models ($M$) changes only the generated output being evaluated.
We describe each protocol in turn.

\emph{(i) Guideline changes via anchored evaluation.}
Comparing guidelines requires a stable evaluation target. We select a preferred guideline $G^\star$ as the canonical construct definition, build the human reference $r^\star$ under $G^\star$ (with $H$ fixed), and measure how well LLM labels produced under each candidate guideline agree with $r^\star$. Writing this anchored reference agreement as $R(G\,;\,G^\star)$, we define the guideline contrast as:
\begin{equation}
\Delta^{(G)} \;\equiv\; R(G_1\,;\,G^\star) \;-\; R(G_0\,;\,G^\star).
\label{eq:swap-G-anchored-ipei}
\end{equation}
This anchoring is asymmetric by design: it evaluates which guideline drives the LLM closer to the construct under $G^\star$. In practice, we designate the iteratively refined guidelines as $G^\star$, treating them as the canonical construct definition following annotation science showing that iterative refinement converges toward more consistent interpretations \citep{lee_common_2023}. The contrast then evaluates whether one-shot guidelines can achieve comparable alignment with this established standard. Interpretation follows Equation~\eqref{eq:diff-id-bias}: improvements in $R(G\,;\,G^\star)$ reflect truth-level gains \emph{net of} any change in co-labeling covariance with the anchored reference. We expect that refined guidelines ($G_1$) are clearer about the annotation criteria compared to the base guidelines ($G_0$), which plausibly decreases the probability of LLM and human reference both being wrong, so $J(c_1)-J(c_0)\le 0$ is reasonable. Here, $\Delta^{(G)}$ is conservative as it underestimates the improvement in $T$.

\emph{(ii) Human procedure selection via internal consistency.}
For a fixed $G^\star$, let $\mathrm{IAA}(H)$ denote within-procedure inter-annotator agreement computed on the annotated sample (e.g., Krippendorff's $\alpha$ or mean Cohen's $\kappa$ over annotator pairs). We define:
\begin{equation}
\Delta^{(H)} \;\equiv\; \mathrm{IAA}(H_1) \;-\; \mathrm{IAA}(H_0).
\label{eq:swap-H-IAA-ipei}
\end{equation}
Under Assumption \ref{assum:sln}, higher IAA implies a lower per-annotator error rate $e$, which yields a more accurate reference aggregate \citep{krippendorff_reliability_2004}.\footnote{Since $a = (1-e) - e/(K-1)$ from Assumption~\ref{assum:sln}, lower $e$ yields a larger identification slope, making observed LLM-reference contrasts more faithful to truth-level contrasts.} Intuitively, when annotators using $(G^\star,H)$ agree more with one another, their majority-vote aggregation better approximates $y$, making the observed reference agreement contrasts better reflect actual reductions in measurement error.

IAA alone, however, does not capture whether the model and the reference tend to make the same mistakes on the same items, which is captured by co-labeling covariance $J$. Even with good IAA, a large positive ``both-incorrect'' component can inflate reference agreement without any improvement against the ground truth. This scenario is especially plausible for crowd worker-generated references for two reasons. First, for complex tasks, many crowd workers and current LLMs may lack domain knowledge that experts apply. Second, when guidelines are under-specified, experts draw on tacit annotation criteria not fully captured in the written guidelines. Crowd workers and LLMs, prompted only with written guidelines, do not apply these tacit rules and consequently may make similar classification errors.

We therefore use $\Delta^{(H)}$ as the primary criterion to choose $H^\star$: pick the human procedure with the higher IAA. However, co-labeling covariance $J$ remains unobserved. To assess whether expert baselines provide a more stable identification, we perform a sensitivity analysis in Section~\ref{sec:expert_vs_crowd}, examining how introducing crowd labels into expert baselines affects reference agreement contrasts.

\emph{(iii) Prompt and model changes.}
With $(G^\star,H^\star)$ fixed, the prompt and model contrasts are:
\begin{align}
\Delta^{(P)} &\equiv R(P_1) \;-\; R(P_0), \label{eq:swap-P-ipei}\\
\Delta^{(M)} &\equiv R(M_1) \;-\; R(M_0). \label{eq:swap-M-ipei}
\end{align}
According to Proposition~\ref{prop:err-reduction}, a positive contrast implies improvement in true agreement when $R(c_1)-R(c_0) > J(c_1)-J(c_0)$. With the reference fixed, a superior $P$ or $M$ (as indicated by a higher $R$) implies better alignment with $y$ and reduced measurement error when the change in co-labeling covariance is sufficiently small. We argue this condition is plausible for two reasons. First, with $G^\star$ and $H^\star$ fixed, any residual shift in $J$ arises solely from the LLM side and should be small. Second, when prompts and models improve true agreement, they are more likely to reduce the ``both-incorrect'' component of $J$, making $J(c_1)-J(c_0)\le 0$ more plausible than $>0$. The observed $R$ gains are therefore conservative as they tend to \emph{understate} the improvement in $T$.

\subsubsection{Sequential optimization strategy.} \label{sec:sequential_optimize}

Empirically, we adopt a sequential approach, starting from a baseline configuration $(M_0,P_0)$, where $M_0$ is a widely adopted closed model and $P_0$ embeds guidelines in a standard prompt structure. First, holding $(H,P,M)$ fixed, we compare candidate guidelines using the anchored contrast in Equation~\eqref{eq:swap-G-anchored-ipei} and select $G^\star$. Second, we compare human procedures via $\Delta^{(H)}$ in Equation~\eqref{eq:swap-H-IAA-ipei} to select $H^\star$. As the final two steps, with $(G^\star,H^\star)$ fixed, we evaluate prompts and models using the reference agreement contrasts in Equations~\eqref{eq:swap-P-ipei}--\eqref{eq:swap-M-ipei}, respectively.\footnote{This ordering reflects two dependencies: (i) guideline selection must precede baseline creation since annotators apply guidelines; (ii) baseline establishment must precede prompt/model optimization to ensure stable evaluation reference.} We adopt this sequential design because it surfaces robust, model-agnostic principles about implementation details (e.g., always placing guideline content in the system role, as empirically shown in Table~\ref{tab:six_prompt_four_models}).

\subsection{Measurement Error Reduction Mechanisms}

We propose the SILICON workflow to instantiate the analytical framework from Section~\ref{sec:sources} through a four-stage protocol (Appendix Figure~\ref{fig:workflow}), where each stage targets a specific measurement error source.

Throughout, we measure agreement using Cohen's Kappa ($\kappa$), which adjusts for chance agreement and supports both pairwise LLM-vs-human comparisons and multi-annotator IAA analysis. Unlike raw agreement rates, $\kappa$ accounts for the probability of agreement occurring by chance, making it appropriate for evaluating annotation quality across tasks with varying class distributions. It extends to multi-label tasks through context-dependent weighting (Appendix~\ref{appendix:cohenkappa}). We adopt a conservative target of $\kappa \geq 0.5$, above the standard ``moderate agreement'' threshold of 0.4 \citep{landis_measurement_1977}, providing a buffer for variability and coinciding with the empirical cutoff where co-labeling covariance effects are minimal (Section~\ref{sec:expert_vs_crowd}).

\subsubsection{Guideline-induced error reduction.}\label{sec:iterative_guidelines}
As established in Section~\ref{sec:prescriptive_guidelines}, LLMs cannot seek clarification during annotation, making explicit guidelines the sole mechanism for consistent interpretation \citep{tornberg_best_2024}. Yet many LLM annotation studies adopt existing guidelines directly or draft new ones in a one-shot fashion \citep{gilardi_chatgpt_2023, rathje_gpt_2024}, risking guideline-induced error. We compare iteratively refined guidelines against one-shot guidelines adapted from existing research. The iterative refinement follows a structured protocol based on \citet{lee_common_2023}: annotators independently apply preliminary definitions, discuss disagreements through documented conversations, and refine guidelines until achieving predetermined IAA thresholds on fresh subsamples. Figure~\ref{fig:iterative_annotation} presents the protocol.

{\OneAndAHalfSpacedXI
\begin{tcolorbox}[
    colback=white,
    colframe=black!75!white,
    arc=3mm,
    title={\textbf{Contrast of Interest: Guideline Effect}},
    boxrule=1pt,
    left=10pt,
    right=10pt,
    top=8pt,
    bottom=8pt
]
With $(H,P,M)$ fixed and evaluation anchored at $G^\star$, replacing one-shot guidelines $G_0$ by iteratively refined ones $G_1$ increases agreement against the anchored reference (Equation~\eqref{eq:swap-G-anchored-ipei}):
\[
\Delta^{(G)} > 0.
\]
\end{tcolorbox}
}

{\OneAndAHalfSpacedXI
\begin{figure}[hbt]
    \centering
    \caption{Iterative Process to Create Annotation Guidelines}
    \label{fig:iterative_annotation}
\begin{tcolorbox}[
    colback=white,
    colframe=black!75!white,
    arc=3mm,
    title={\textbf{Iterative Annotation Process Flow}},
    boxrule=1pt,
    left=10pt,
    right=10pt,
    top=8pt,
    bottom=8pt
]
\begin{enumerate}
    \item Research Assistants (RAs) independently annotate an initial sample dataset based on preliminary definitions
    \item Measure inter-annotator agreement (IAA) and compare it with the predefined threshold
    \item If IAA threshold is not met:
    \begin{itemize}
        \item Conduct documented discussions focusing on disagreements
        \item Reshuffle and re-annotate the \textit{same} sample independently
        \item Repeat until threshold is met
    \end{itemize}
    \item Once threshold is met, proceed to annotate a \textit{new} sample
    \item Iteration concludes when RAs achieve the IAA threshold on first pass with a new sample
    \item RAs independently draft annotation guidelines and then collaboratively merge them
\end{enumerate}
\end{tcolorbox}
\end{figure}
}

\subsubsection{Baseline-induced error reduction.}\label{sec:expert_baseline}
Human annotation baselines serve as the reference standard against which LLM annotations are evaluated, making baseline quality crucial for valid performance assessment. Domain experts bring specialized knowledge and nuanced judgment \citep{snow_cheap_2008}, while crowd workers offer scalability but typically lack deep understanding of annotation guidelines \citep{snow_cheap_2008, zhang_needle_2023}. A key challenge is the infeasibility of recruiting domain experts in many scenarios. SILICON addresses this by creating task-specific ``annotation experts'' through its iterative guideline development process: individuals who demonstrated consistent understanding through high IAA. Having internalized consistent interpretations through the collaborative development process, they reduce evaluation-driven measurement error by minimizing disagreements about ambiguous cases. The protocol compares expert-generated baselines against crowd-generated baselines using identical guidelines and evaluation protocols, with sample sizes of 120 to 200 items \citep{lee_common_2023}.

{\OneAndAHalfSpacedXI
\begin{tcolorbox}[
    colback=white,
    colframe=black!75!white,
    arc=3mm,
    title={\textbf{Contrast of Interest: Baseline Effect}},
    boxrule=1pt,
    left=10pt,
    right=10pt,
    top=8pt,
    bottom=8pt
]
Under fixed $(G^\star,P,M)$, annotation expert procedures $(H_1)$ exhibit higher within-procedure agreement than crowd worker procedures $(H_0)$ (Equation~\eqref{eq:swap-H-IAA-ipei}):
\[
\Delta^{(H)} > 0.
\]
\end{tcolorbox}
}

\subsubsection{Prompt-induced error reduction.}\label{sec:system_prompt}
Meta-prompt structure affects LLM annotation performance even when guidelines remain verbatim constant, creating prompt-induced measurement error. As established in Section~\ref{sec:prescriptive_guidelines}, LLMs operate exclusively on textual input, making the structure surrounding the guidelines consequential. We implement controlled meta-prompt optimization focusing on three elements: (1) strategic positioning of annotation guidelines within prompt structure (system prompt versus user prompt), (2) adoption of task-relevant personas to align model perspective with domain expectations \citep{hu_quantifying_2024}, and (3) incorporation of reasoning strategies such as chain-of-thought prompting \citep{wei_chain--thought_2022}.

{\OneAndAHalfSpacedXI
\begin{tcolorbox}[
    colback=white,
    colframe=black!75!white,
    arc=3mm,
    title={\textbf{Contrast of Interest: Prompt Effect}},
    boxrule=1pt,
    left=10pt,
    right=10pt,
    top=8pt,
    bottom=8pt
]
For fixed $(G,H,M)$, an improved meta-prompt reduces measurement error and raises reference agreement (Equation~\eqref{eq:swap-P-ipei}):
\[\Delta^{(P)} > 0.\]
\end{tcolorbox}
}

\subsubsection{Model-induced error assessment.}\label{sec:model_choice_regression}
Different LLMs exhibit varying annotation performance due to architectural differences, training data variations, and alignment strategies \citep{mousavi_lexicons_2024}. The critical question for robust reproducibility is whether annotation results remain stable across different models. We evaluate multiple state-of-the-art LLMs using identical guidelines, baselines, and meta-prompts to isolate model-specific performance differences. The evaluation includes both closed models and open-weight models to address reproducibility concerns (Section~\ref{sec:robust_reproducibilty}). We employ a regression-based method to test for statistical differences, clustering standard errors at the item level; the regression equation is presented in Section~\ref{sec:regression-equation}, with details in Appendix~\ref{appendix:regression}. Beyond identifying which models perform best, Section~\ref{sec:multi-llm} describes how constructing an ensemble-based upper bound on focal-model performance supports this comparison.

{\OneAndAHalfSpacedXI
\begin{tcolorbox}[
    colback=white,
    colframe=black!75!white,
    arc=3mm,
    title={\textbf{Contrast of Interest: Model Effect}},
    boxrule=1pt,
    left=10pt,
    right=10pt,
    top=8pt,
    bottom=8pt
]
 For fixed $(G,H,P)$, model changes yield non-zero contrast (Equation~\eqref{eq:swap-M-ipei}):
\[\Delta^{(M)} \neq 0.\]
\end{tcolorbox}
}

\subsection{Robust Reproducibility Protocol}\label{sec:robust_reproducibilty}

We operationalize robust reproducibility as a lack of statistical difference between a focal model and a target model, assessed via regression-based testing. The natural question is: \textit{reproducible with respect to what target?} Two natural targets arise: (1) an open-weight baseline anchored by a permanently accessible open-weight model, and (2) an ensemble-based upper bound under stronger labeling support. We next describe the protocol for each.

\subsubsection{Open-weight baseline.}\label{sec:regression-equation}
In practice, researchers may choose a focal model for deployment based on performance and cost considerations.\footnote{For illustration purposes, in this paper, we consistently choose the best-performing model (yielding the highest Cohen's Kappa against human baselines) in each task as the focal model. } If the chosen focal model is closed, researchers should run the open-weight baseline test described below before relying on it for large-scale annotation. Open-weight models are permanently accessible and, when the focal model is closed, provide a conservative benchmark for the focal model's performance. If the focal model is statistically indistinguishable from an open-weight alternative, researchers obtain a baseline reproducibility guarantee that survives model retirement. If the focal model is itself open-weight, no open-weight baseline test is required for that task. We test whether open-weight models can substitute closed focal models without significant quality loss using our regression-based statistical testing framework, reporting confidence intervals. When an interval includes zero, we interpret this as insufficient evidence of a difference (i.e., we fail to reject the null of no difference).

Formally, denoting the tested LLMs by $m$ (1, 2, ..., $M$) and the item (each annotation document, such as each business proposal document) by $i$ (1, 2, ..., $I$), we present the logistic regression model below:
\begin{equation}
\text{logit}(P(\mathit{Matched}_{im}=1)) = \alpha_0 + \alpha_1 \mathds{1}(m=1) + \alpha_2 \mathds{1}(m=2) + \dots + \alpha_{M-1}\mathds{1}(m=M-1),
\end{equation}
where the dependent variable ($\mathit{Matched}_{im}$) is an indicator variable equal to 1 if model $m$'s label for item $i$ matches the human reference baselines, and 0 otherwise. Standard errors are clustered at the item level (at the \textit{i} level) to account for the repeated-measures structure. Coefficients are interpreted as log-odds differences relative to the omitted baseline model.
This methodology enables the assessment of whether a focal model is robustly reproducible with respect to a target model, operationalized as the absence of a statistically detectable performance difference. Note that the regression's dependent variable, item-level label accuracy, intentionally differs from the Cohen's Kappa metric used for focal model selection. The regression directly tests whether a substitute model reproduces correct labels at the same rate as the focal model on the same items, which is the quantity of interest for practical reproducibility. By contrast, Kappa additionally corrects for chance agreement: a model that correctly labels ``easy'' items (those whose ground-truth category is common) receives less credit under Kappa than under accuracy. Both metrics are informative, but for the specific question of whether a researcher can switch models without degrading label quality on a given corpus, item-level accuracy is the more direct test.\footnote{As a robustness check, we also estimated fractional logistic regressions using a weighted agreement score as the dependent variable; results are qualitatively unchanged (see Appendix~\ref{app:fractional_logit}).}

\subsubsection{Upper bound: ensemble-based performance ceiling.}\label{sec:multi-llm}
As established in Section~\ref{sec:robust_repro_lit}, multi-model labeling raises two questions: does augmenting the focal model with labels from complementary models actually improve aggregate performance, and can labels be introduced selectively to achieve the best outcome? To answer both questions, we construct a task-specific three-model ensemble anchored on the best-performing model for each task, augmenting it with the second- and third-best models. The highest observed performance across all ensemble configurations is treated as the empirical upper bound for that task.

We operationalize selective labeling through confidence-based routing, which sends low-confidence items to auxiliary models instead of applying all models to every item \citep{ipeirotis_repeated_2014}. We estimate confidence post hoc using the First-Second Distance (FSD) method \citep{lyu_calibrating_2025}. Items with FSD below a threshold $(\tau$) are sent to auxiliary models for majority voting, while high-confidence items retain the focal model's label. Sweeping $\tau$ from 0 to 1 yields a family of routed ensemble configurations, ranging from the focal model alone to full integration.

This design addresses both questions through the shape of the routing curve. Because the focal model is already the best-performing single model, routing all items to majority vote can override correct focal labels on high-confidence cases. The best ensemble outcome may therefore occur at an interior routing threshold rather than at full integration. Whether auxiliary models improve performance at all depends on whether they contribute complementary error profiles; if they replicate the focal model's mistakes, additional labels add cost without information. The routing curve thus serves a diagnostic role: it reveals whether model aggregation is helpful for a given task, and if so, which threshold yields the greatest improvement. The curve also informs a practical tradeoff between performance and cost. If aggregate performance increases sharply up to a certain threshold and then improves only marginally, researchers must weigh whether the additional cost of routing more items to multiple models is justified by the relatively small gain in overall performance. Implementation details of the routing procedure are provided in Appendix~\ref{app:fsd_routing}.

\subsection{Toolkit and Implementation}
We have created an open-source toolkit that operationalizes SILICON, which we will release upon publication. The toolkit includes utilities for agreement calculation for guideline refinement and baseline establishment; automated pipelines to compare different meta-prompt settings and multiple models; modules for model-confidence estimation and multi-LLM annotation; and statistical tests for robust reproducibility.

\section{Empirical Validation Design}\label{sec:validation}

We validate the SILICON workflow across seven annotation cases that are representative of management research, organized by the level of methodological control achievable for each error source.
The first category comprises four cases: business proposal evaluation, review attribute detection, dialog intent classification and breakdown analysis, and affective content evaluation. For these scenarios, we implement the complete SILICON workflow, developing the annotation guidelines iteratively and establishing annotation-expert-validated human baselines with three research assistants per task.
The second category comprises three supplementary scenarios: language toxicity detection, criticism stance detection, and sentiment analysis. For these, we adopted guidelines and baselines directly from published papers, which limits our ability to verify the original IAA or baseline provenance.
Consequently, we treat these scenarios as supplementary evidence and draw our primary conclusions from the first four scenarios.
Appendix \ref{app:sum_cases} presents a summary of original research questions, annotation approaches, SILICON process implementation, and key findings for each case, and Online Appendix presents corresponding prompts.\footnote{Link to Online Appendix for Prompt Details: \url{https://drive.google.com/file/d/14sFKxeSANFX3Xl-umVxGNR_6qKZ5_M5Y/view?usp=sharing}  (This file is anonymous and non-tracking).}

\subsection{Primary Cases with Full Methodological Control.}

\subsubsection{Business proposal evaluation.}
Business documentation analysis (e.g., of business proposals) plays a critical role in organizational decision making and strategic behavior \citep{cao_peer_2019}.
We focus on the classification of decentralized autonomous organization (DAO) business proposals, building on the work of \cite{obermeier_decentralized_2024}. Each ``proposal'' was categorized into one of six categories based on their primary function or intention: Organizational, Business Model, Marketing, Functionality, Security, or Other. The categorization process aims to reveal patterns in how DAOs delegate and structure decision rights across their platforms.

\subsubsection{Review attribute detection.}
Online reviews contain information critical to understanding consumer purchase decisions. We obtained the annotated dataset from \cite{liu_large-scale_2019}, where the authors used a large scale deep learning approach to identify the presence of six attributes in product reviews, including price, performance feature, relatability/durability, conformance, aesthetics, and perceived quality. This is a multi-label classification task which provides insights into how different product attributes influence consumer decision-making.

\subsubsection{Dialog intent classification and breakdown analysis.} Dialog analysis is critical across various domains, particularly in understanding and improving human-computer interaction.
We focus on two critical aspects of dialog analysis: (1) dialog intent classification \citep{stolcke_dialogue_2000}, and (2) dialog breakdown analysis, which identifies points in a conversation where communication fails \citep{higashinaka_dialogue_2016}. We use a sample of customer service dialog data obtained from a Hugging Face dataset.\footnote{\href{https://huggingface.co/datasets/bitext/Bitext-customer-support-llm-chatbot-training-dataset}{https://huggingface.co/datasets/bitext/Bitext-customer-support-llm-chatbot-training-dataset}}
The first task, dialog intent classification, requires human annotators to classify the intention of each utterance into one or multiple categories (i.e., multi-label classification): factual questions, yes/no questions, task commands, invalid commands, appreciations, complaints, comments, non-opinion statement, positive answers, and negative answers.
The second task, the dialog breakdown analysis, focuses on identifying whether an utterance in the conversation leads to a breakdown, classifying each utterance as: (1) not a breakdown, (2) a possible breakdown, or (3) confirmed breakdown.

\subsubsection{Affective content evaluation (connectedness presence and classification).}
Affective content in marketing, including advertisements, packaging, and product descriptions, shapes consumer decision-making \citep{paharia_underdog_2011}.
We use a dataset from \citet{cheng_support_2024}, which focuses on one dimension: the sense of connectedness; specifically the presence and nature of connectedness in online product descriptions.
The first task, connectedness presence detection, is a binary classification task where human annotators assess whether a product description conveys a sense of connectedness or not.
The second task, connectedness classification, is conducted only when connectedness is present. This is a multi-label classification task where each product description may be assigned one or more of the following categories: Emotional Appeal, Storytelling, Relatable Scenarios, Empathy, and Cultural or Community Connection.

\subsection{Complementary Cases with Limited Methodological Control.}

\subsubsection{Language toxicity detection.}
Detecting toxic content in textual data plays a critical role in management research, with applications spanning online communities \citep{matook_user_2022, sibai_why_2024} and organizational behavior \citep{rosette_why_2013}.
We use annotation guidelines, human baseline, and data publicly available from \cite{saha_rise_2023}.
The annotated items are social media posts from Gab. The task is to mark each post as a) fear speech, b) hate speech, c) normal, or d) both fear speech and hate speech (if the post contains elements of both).

\subsubsection{Criticism stance detection.}
We selected this case because the core task---criticism stance detection---can be generalized to address a broad range of organizational and social challenges. Specifically, we use data from \cite{peng_dynamics_2022}, which focuses on detecting the criticism stance of tweets about certain academic papers, framing it as a binary classification task.

\subsubsection{Sentiment analysis.}
Sentiment analysis has broad implications in management research, spanning social media \citep{zhang_large-scale_2016, oh_are_2023, tang_racial_2024}, e-commerce \citep{homburg_measuring_2015},
and firm-to-public communications \citep{choudhury_machine_2019}.
For this task, we utilize annotation guidelines, human baselines, and raw data from \cite{sen_human_2020}.

\subsection{Prompt and Model Specifics}

To keep the focus on methodological insight rather than model-specific benchmarking, from this point forward, we refer to each model by a short code that encodes its provider and weight-access status (closed vs. open). We retain provider and openness because models from the same provider often exhibit correlated behavior, and open-weight versus closed-weight access directly affects reproducibility. The mapping is: OA-CW1 (GPT-4o), OA-CW2 (o3-mini), OA-CW3 (GPT-4.5), OA-CW4 (GPT-4.1), OA-OW1 (GPT-OSS-120B), GO-CW1 (Gemini 1.5 Pro), GO-CW2 (Gemini 2.5 Pro), AN-CW1 (Claude 3.5 Sonnet), AN-CW2 (Claude 3.7 Sonnet), ME-OW1 (LLaMA 3.3 70B), ME-OW2 (LLaMA 3 70B), and DS-OW1 (DeepSeek-R1), where CW denotes closed-weight and OW denotes open-weight.\footnote{More specifically, we use the following versions of models: GPT-4o-2024-08-06, o3-mini-2025-01-31, Claude-3-5-Sonnet-20240620, Gemini-1.5-pro, Llama-3.3-70B-Instruct, DeepSeek-R1, and GPT-OSS-120B.}

The prompt strategies included a base prompt, a prompt with persona settings, and a prompt with chain-of-thought, all unified in their placement within the system role rather than the user role (for reasons that will appear in Section \ref{result:prompt}). These prompts are initially tested using the focal model OA-CW1 to establish a benchmark.
Based on the results from OA-CW1, we identify the optimal prompt by balancing performance and cost considerations.
This optimal prompt is then applied across several models: OA-CW1, OA-CW2, AN-CW1, GO-CW1, ME-OW1, DS-OW1, and OA-OW1.
Acknowledging that the best-performing prompt might vary between models, we further examine this variability in the business proposal evaluation task. Specifically, we test all three prompt strategies, altering their placement (system role versus user role) to generate six distinct prompts. These prompts are applied to a selective subset of four models.
We set the temperature to 1 across all models, aligning with the default value for OA-CW1.

\section{Empirical Results} \label{sec:result}

We first present measurement error reduction results in the same sequence as described in Section \ref{sec:sequential_optimize}: guideline contrast, reference baseline contrast, meta-prompt contrast, and model contrast. We then evaluate robust reproducibility by testing an open-weight substitution and ensemble-based upper bounds.

\subsection{Measurement Error Reduction} \label{sec:result_error}
\subsubsection{Reducing guideline-induced error.}

\begin{figure} [h]
    \centering
    \caption{LLM Performance based on One-shot vs. Iteratively Refined Annotation Guidelines}
    \includegraphics[width=0.8\linewidth]{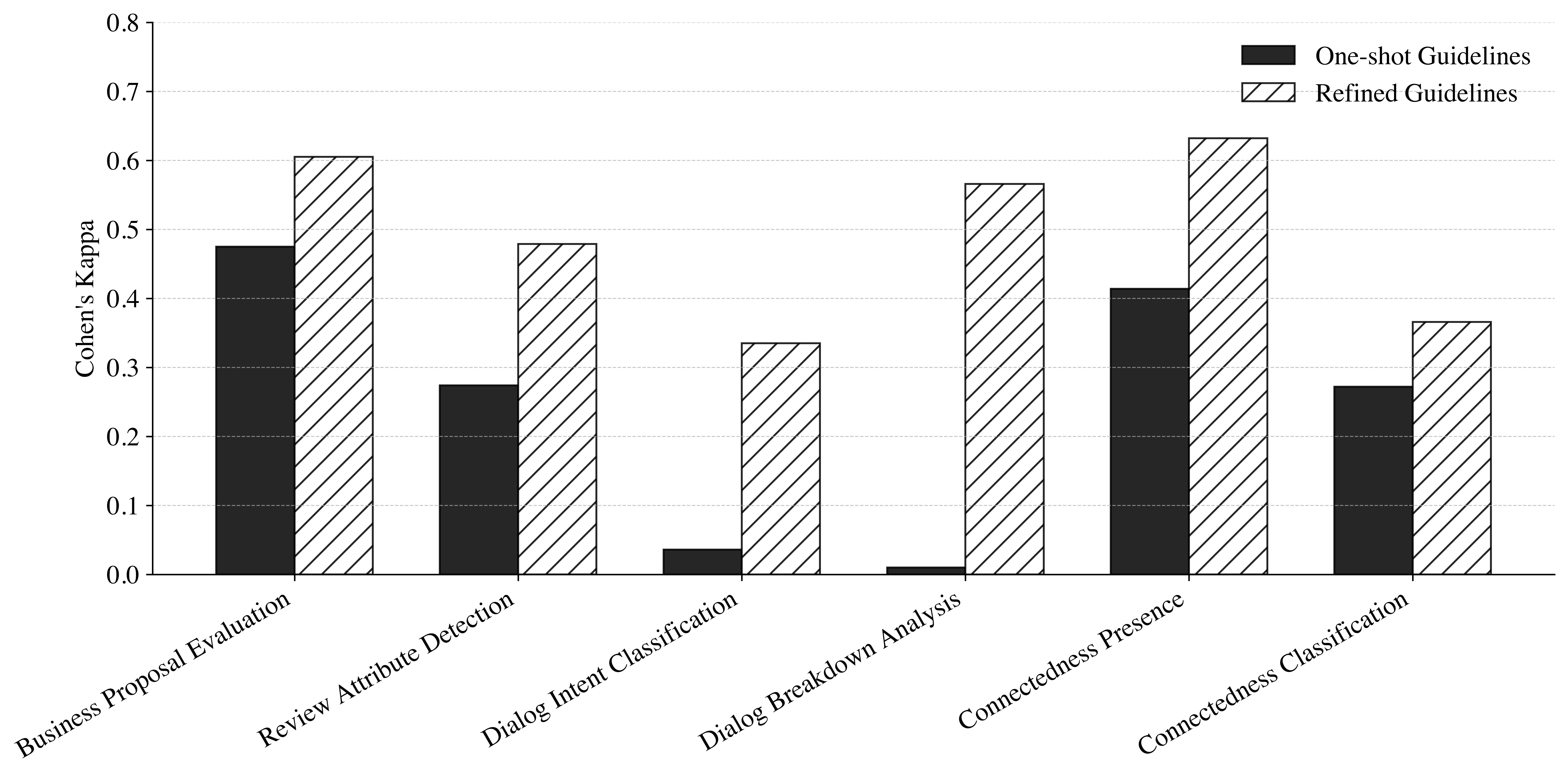}

    \begin{flushleft}
        \footnotesize \parbox{\linewidth}{
     \textit{Notes.} The y-axis shows the $\kappa$ between LLM annotation and human annotation baselines (i.e., the majority votes from experts). The LLM used is OA-CW1.  LLM performance based on one-shot annotation guidelines is always worse than that based on iteratively refined annotation guidelines.
        }
    \end{flushleft}

    \label{fig:initial_vs_refined}
\end{figure}

Ambiguous or incomplete annotation guidelines can create systematic measurement error when annotators apply inconsistent interpretations. We examine whether iterative guideline refinement reduces this source of error by comparing LLM annotation performance using one-shot (pre-existing) guidelines versus iteratively refined ones developed through our structured protocol. Figure~\ref{fig:initial_vs_refined} presents Cohen's Kappa values between LLM annotations and expert-generated ground truth across six tasks. LLM performance is consistently lower when using one-shot guidelines compared to iteratively refined guidelines. The improvement ranges from 0.1 to 0.5 Kappa points, with the largest gains observed in dialog intent classification and breakdown analysis tasks, where initial agreement levels below 0.1 (indicating no meaningful agreement) improve to fair agreement level after guideline refinement.

These findings provide strong support for the argument that when holding human baseline procedures ($H$), meta-prompts ($P$), and models ($M$) constant while anchoring evaluation at refined guidelines (\(G^*=G_1\)), the transition from $G_0$ to $G_1$ increases reference agreement ($\Delta^{(G)}>0$).
Appendix Figure~\ref{fig:ra_iteration} illustrates the gains from the iterative refinement process, showing that IAA among annotation experts generally rises over multiple revision rounds, though occasional non-monotonic dips occur within individual rounds. Achieving high IAA typically requires several iterations for complex tasks, supporting our workflow that prescriptive guidelines must be systematically developed rather than created in isolation.

\subsubsection{Reducing baseline-induced error.}\label{sec:expert_vs_crowd}
Unreliable human reference standards introduce systematic baseline-induced error when the evaluation standard itself diverges from ground truth. We test whether expert-generated baselines reduce this source of error by comparing the internal consistency of annotation experts versus crowd workers using identical annotation guidelines.

\begin{figure} [h]
 \vspace{-.75em}
    \centering
    \caption{Inter-Annotator Agreement Comparison: Expert- and Crowd Worker-labeled Baseline}
    \includegraphics[width=0.8\linewidth]{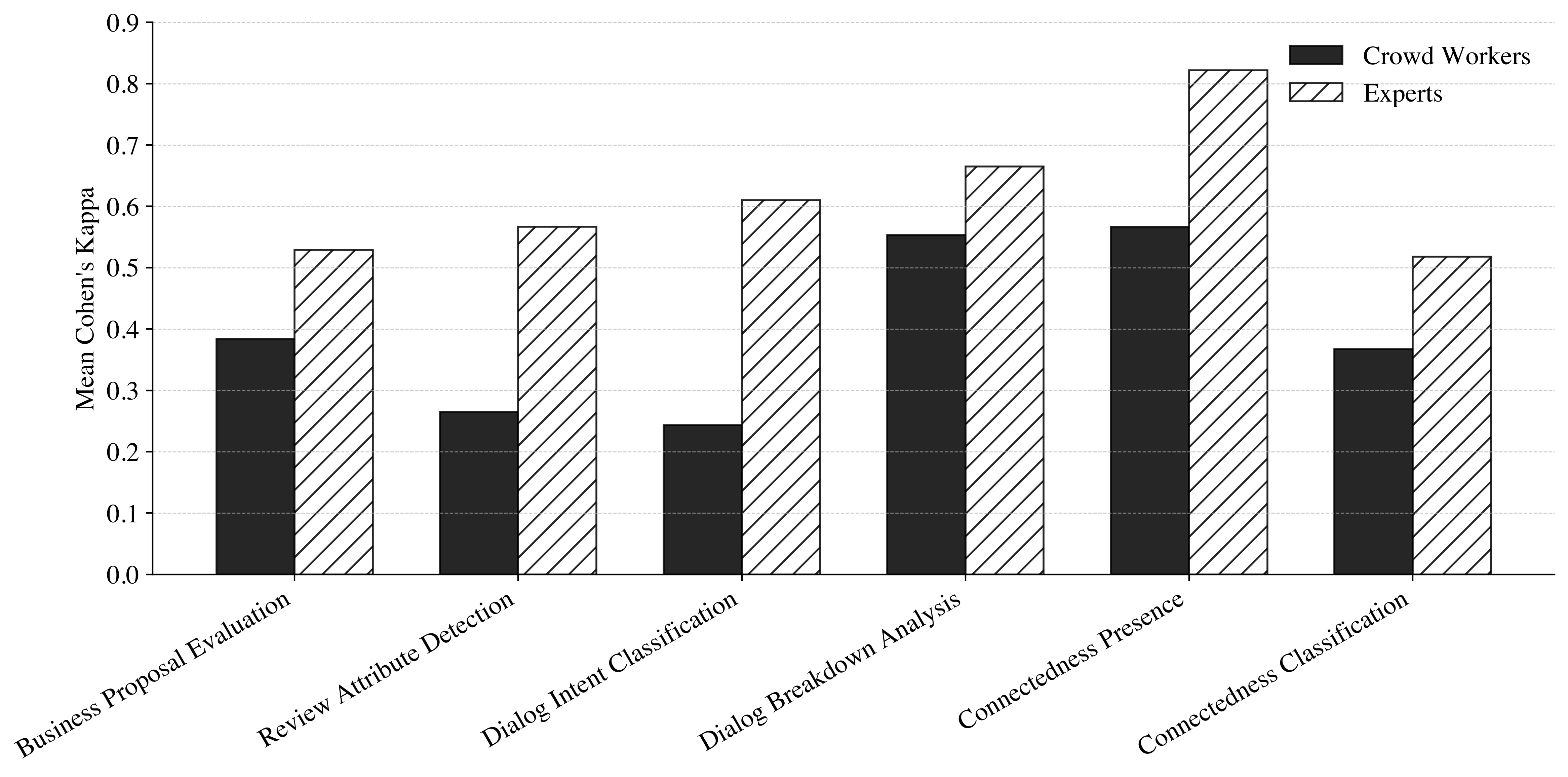}

    \begin{flushleft}
        \footnotesize \parbox{\linewidth}{
     \textit{Notes.} The y-axis shows mean $\kappa$ among three annotators (either annotation experts or crowd workers). Expert-labeled annotations consistently yield higher agreement across all tasks.
        }
    \end{flushleft}
    \label{fig:crowd_vs_llm_iaa}
\end{figure}

For each of the four primary cases (comprising six sub-tasks), we construct two distinct annotation baselines using our finalized annotation guidelines. The expert baseline is obtained from three undergraduate research assistants who participate in the iterative guideline development process, while the crowd worker baseline is obtained from three different research assistants who receive the guidelines without participating in their development.
Both groups annotate 200 to 300 randomly chosen documents using identical protocols. Figure~\ref{fig:crowd_vs_llm_iaa} presents the IAA comparisons between the two baseline procedures. The annotation experts consistently achieve higher consensus compared to crowd workers across all six tasks. For instance, in the dialog intent classification task, the mean Cohen's $\kappa$ among expert annotators is 0.61 (indicating substantial agreement), whereas crowd workers achieve only 0.23, reflecting weaker alignment and less consistent understanding of the annotation criteria.

These findings provide strong support for the argument that under fixed guidelines (\(G^*\)) and models/prompts (M,P), expert annotation procedures (\(H_1\)) exhibit higher within-procedure agreement than crowd worker procedures (\(H_0\)) yielding ($\Delta^{(H)}>0$). The superior consistency occurs because the iterative refinement process allows experts to develop shared mental models of the task, facilitating more consistent interpretations of complex phenomena \citep{ruggeri_let_2024}. This higher baseline reliability provides a more demanding and stable evaluation standard to assess LLM annotation quality.

 \begin{figure} [h]
    \centering
    \caption{LLM Agreement with Expert vs. Crowd Baselines}
    \includegraphics[width=0.75\linewidth]{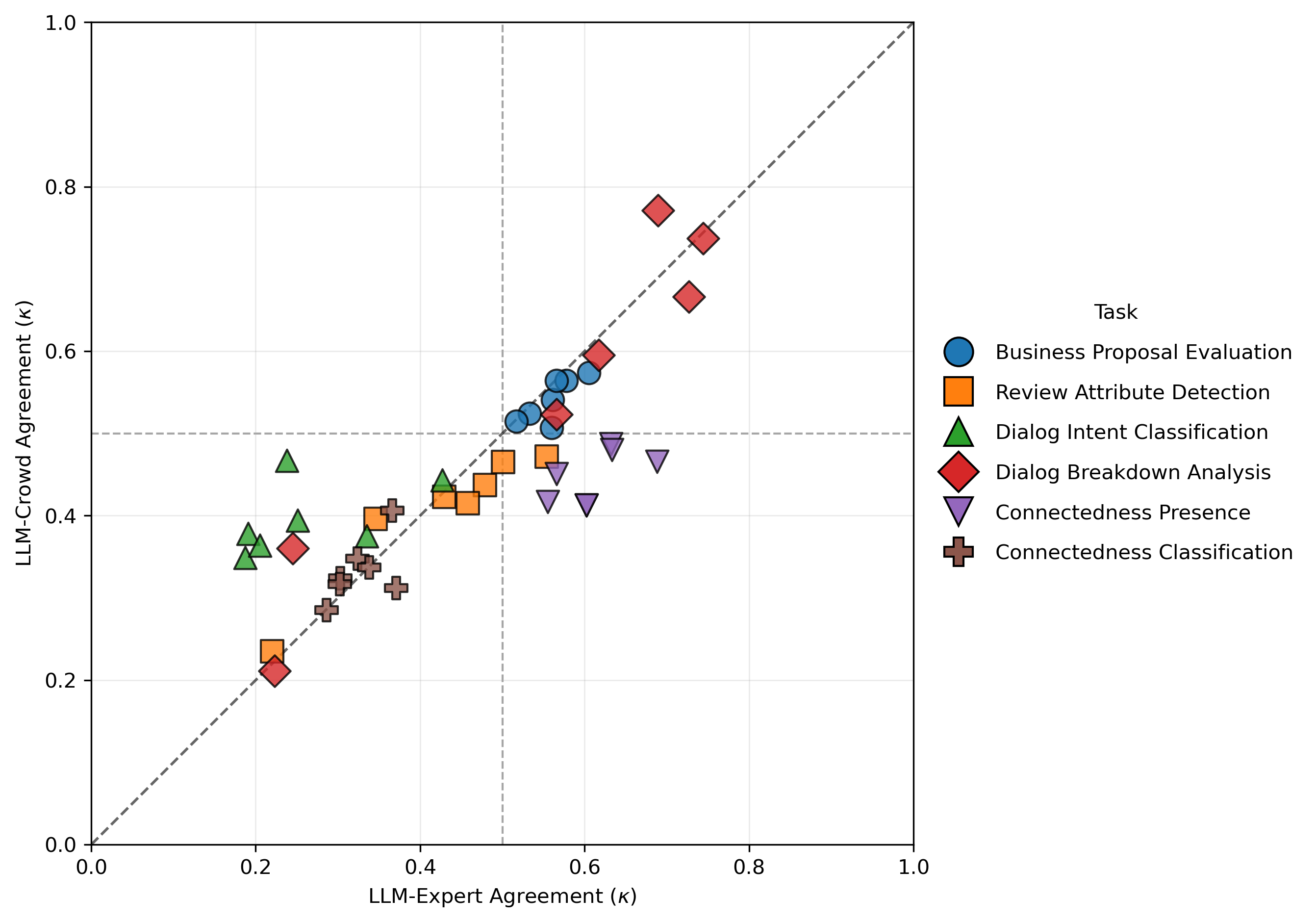}
     \begin{flushleft}
        \footnotesize \parbox{\linewidth}{
     \textit{Notes.} Each point represents a model-task pair with coordinates \((x=\kappa(\text{LLM},\text{expert}),\,y=\kappa(\text{LLM},\text{crowd}))\). Colors and markers denote tasks. The diagonal indicates equal agreement; points below it indicate stronger LLM alignment with experts, while points above it indicate stronger alignment with crowd workers. Gray dashed lines mark \(\kappa=0.5\) on each axis.
        }
    \end{flushleft}
       \label{fig:llm_kappa_with_crowd_vs_expert}
\end{figure}

We next compare how well LLM annotations align with expert versus crowd baselines. Figure~\ref{fig:llm_kappa_with_crowd_vs_expert} plots each model-task combination as a point with coordinates \((x=\kappa(\text{LLM},\text{expert}),\,y=\kappa(\text{LLM},\text{crowd}))\). The diagonal indicates equal agreement; points \emph{below} the diagonal indicate stronger alignment with experts, and points \emph{above} it indicate stronger alignment with crowds.
Two patterns emerge: first, alignment is not uniformly expert-leaning: the \emph{Dialog Intent Classification} points are mostly above the diagonal, whereas the \emph{Connectedness Presence} points cluster below it. Within our workflow, the dialog-intent pattern is consistent with positive co-labeling covariance \(J\) (Lemma~\ref{lemma:bias-aware}): LLMs and crowd workers tend to make the same mistakes, inflating reference agreement \(R\) without improving true agreement \(T\). Second, once both expert and crowd agreement enter the acceptable range (\(\kappa \geq 0.5\), marked by the dashed lines), almost all points are on or below the diagonal. In other words, when LLM-human agreement is high enough to be practically useful, it usually aligns more closely with experts than with crowd workers. Together, these patterns support the use of expert baselines for downstream prompt and model comparisons.

Furthermore, Appendix Figure~\ref{fig:sensitive_crowd_expert} shows how measured LLM–human agreement changes when expert-only labels are replaced with mixes that include crowd labels. The gap grows quickly for Dialog Intent Classification and Connectedness Presence, so conclusions shift under crowd baselines, while Business Proposal Evaluation and Review Attribute Detection are largely flat, so conclusions remain stable.
These patterns show how baseline choice can influence reference agreement through co-labeling covariance and motivate our use of high-IAA expert baselines.

A remaining concern is whether the configuration selected by SILICON could owe its observed $R$ advantage to co-labeling covariance $J$ rather than genuine accuracy $T$.
We address this by estimating $J(c)$ on items where all three expert annotators unanimously agree, so the majority-vote label is a near-perfect ground-truth proxy ($e\approx 0$).
On this high-confidence subset the LLM match rate estimates $T(c)$ cleanly, and we back out $J(c) = R(c) - a\,\hat T(c) - b$ per Lemma~\ref{lemma:bias-aware}.
As shown in Appendix~\ref{app:j_estimation}, across all six tasks the best-performing model exhibits $|J|\leq 0.01$ for four tasks, and the two multi-label exceptions (Review Attribute Detection, Connectedness Classification) have somewhat larger $|J|$.
This provides supportive evidence that the configuration rankings produced by SILICON are driven by true accuracy gains rather than shared error modes with the human reference.

\subsubsection{Reducing prompt-induced error.} \label{result:prompt}
Meta-prompt structure can introduce systematic measurement error even when annotation guidelines remain constant. We test whether strategic meta-prompt optimization reduces this source of error by examining three prompting strategies across multiple models while maintaining verbatim guideline integrity. Building on the methodological principle of fixing the annotation guidelines and optimizing only the meta-prompt, we explored three prompting strategies: base, persona, and chain-of-thought (CoT). We also examined their placement in either the system role (a dedicated instruction channel that sets the model's behavior before the conversation begins) or the user role (the standard message channel that carries the annotation request), resulting in six prompt configurations.
 We deploy these variants across four state-of-the-art LLMs in the business proposal evaluation task to isolate prompt-induced effects.

Table~\ref{tab:six_prompt_four_models} presents LLM-human agreement under different prompt configurations. The best-performing prompts across all models consistently involve system-role placement of the guidelines.
However, our regression-based statistical analysis does not provide sufficient evidence to reject the null hypothesis of performance equivalence between system and user role configurations. This provides mixed support for the impact of meta-prompt on measurement error: while we observe consistent trends favoring system role placement (which we advocate as a default strategy), the improvements do not reach statistical significance ($\Delta^{(P)}\approx 0$). Furthermore, the relative performance of the base, persona, and CoT prompt types shows considerable heterogeneity across models. For instance, when guidelines are placed in the system role, OA-CW1 achieves optimal performance with the base prompt, while AN-CW1 peaks with chain-of-thought prompting. This model-specific sensitivity suggests that meta-prompt optimization benefits vary by architecture.

Practically, the findings suggest that system role placement offers practical advantages for researchers operating under time or budget constraints. Furthermore, competitive performance can be obtained with minimal meta-prompt engineering. The prompt-induced error reduction may be smaller than anticipated, mainly because prescriptive guidelines limit the marginal gains from such meta-prompt optimization.

{\OneAndAHalfSpacedXI
\begin{table}[h!] \small
\centering
\caption{LLM-Human Agreement under Different Prompts (Business Proposal Evaluation Task)}
\label{tab:six_prompt_four_models}
\begin{threeparttable}
\begin{tabular}{@{\extracolsep{\fill}}>{\centering\arraybackslash}p{0.2\textwidth}
>{\centering\arraybackslash}p{0.11\textwidth}
>{\centering\arraybackslash}p{0.11\textwidth}
>{\centering\arraybackslash}p{0.11\textwidth}
>{\centering\arraybackslash}p{0.11\textwidth}
>{\centering\arraybackslash}p{0.11\textwidth}
>{\centering\arraybackslash}p{0.11\textwidth}@{}}
\toprule
Model Code          & \multicolumn{3}{c}{Guidelines in System Role} & \multicolumn{3}{c}{Guidelines in User Role} \\
\cmidrule(lr){2-4} \cmidrule(lr){5-7}
                        & Base & Persona & CoT & Base & Persona & CoT \\
\midrule
OA-CW1& 0.605& 0.582& \textbf{0.610}& 0.582& 0.592& 0.608
\\
GO-CW1 & \textbf{0.561}& 0.512& 0.487& 0.547& 0.544& 0.515
\\
AN-CW1 & 0.578& 0.503& \textbf{0.616}& 0.540& 0.577& 0.580
\\
ME-OW1  & 0.566& \textbf{0.589}& 0.527& 0.569& 0.555& 0.541\\

\bottomrule
\end{tabular}
\begin{tablenotes}
    \footnotesize
    \setlength{\baselineskip}{\normalbaselineskip}
    \item \textit{Notes.} The table presents Cohen's Kappa scores between LLM annotations and human reference labels under different prompt strategies. The three prompting strategies (base prompt, with personas, and chain-of-thought) and their placement (system role versus user role) generate six distinct prompts. Codes follow provider-openness-index naming, where CW denotes closed-weight and OW denotes open-weight.
\end{tablenotes}
\end{threeparttable}
\end{table}
}

\subsubsection{Reducing model-induced error.}
Even with guidelines, baselines, and prompts held constant, different LLM architectures introduce model-induced error through varying performance on identical annotation tasks. We test whether models produce significantly different annotation outcomes by evaluating seven state-of-the-art LLMs across nine representative management research tasks. The empirical results reveal two important insights.

First, cross-model differences warrant model testing and selection.
We find no universally superior model, as performance leadership varies substantially across annotation contexts (Table~\ref{tab:sum_model}). This variation provides strong support for the argument that for fixed $(P,G,H)$, model changes yield significant contrast ($\Delta^{(M)} \not= 0$).

Among the tested models, no single model dominates across contexts. AN-CW1 excels in tasks such as dialog intent classification ($\kappa=0.427$), connectedness presence ($\kappa=0.688$), and connectedness classification tasks ($\kappa=0.371$), whereas DS-OW1 performs best in dialog breakdown analysis ($\kappa=0.744$), criticism stance detection ($\kappa=0.580$), and sentiment analysis ($\kappa=0.935$). However, these two models show dramatic drops on other tasks. For instance, OA-CW1 achieves the highest performance in business proposal evaluation ($\kappa=0.605$) and language toxicity detection ($\kappa=0.471$), while OA-CW2 leads in review attribute detection ($\kappa=0.554$). These task leaders become the focal models in the open-weight baseline and upper-bound analyses below.
One interesting observation is that reasoning-focused models, such as OA-CW2 and DS-OW1, display greater performance variability across tasks.
For instance, OA-CW2 attains a high agreement level in review attribute detection, while its performance in dialog breakdown analysis drops sharply.
This volatility reflects optimization of the reasoning models for analytical cognition (conscious deliberation) rather than intuitive cognition (i.e., grounded, situational pattern recognition) \citep{patterson_intuitive_2017} that annotation tasks require, suggesting completion-optimized models may be more suitable for consistent annotation performance.
Overall, these findings establish the necessity of task-specific model evaluation and highlight two critical challenges: determining when low model confidence warrants additional votes from multiple models, and ensuring robust reproducibility as high-performing closed models face retirement cycles. We address these challenges in the following sections.

{\OneAndAHalfSpacedXI
\begin{table}[h!] \footnotesize
\centering
\caption{Cross-Model Performance Comparison: Human-LLM Agreement}
\label{tab:sum_model}
\begin{threeparttable}
\begin{tabular}{@{\extracolsep{\fill}}>{\centering\arraybackslash}p{0.26\textwidth}>{\centering\arraybackslash}p{0.08\textwidth}>{\centering\arraybackslash}p{0.09\textwidth}>{\centering\arraybackslash}p{0.09\textwidth}>{\centering\arraybackslash}p{0.10\textwidth}>{\centering\arraybackslash}p{0.09\linewidth}>{\centering\arraybackslash}p{0.09\linewidth}@{}>{\centering\arraybackslash}p{0.09\linewidth}}
\toprule
    Task& OA-CW1&OA-CW2&OA-OW1&GO-CW1&AN-CW1&ME-OW1&DS-OW1\\
\midrule
 Business Proposal Evaluation&\textbf{0.605}&0.560
&0.533
&0.561
&0.578
&0.566
 &0.517\\
 Review Attribute Detection& 0.479
& \textbf{0.554}
& 0.501
& 0.346
&0.429
& 0.220
 &0.458\\
 Dialog Intent Classification& 0.335
& 0.191
& 0.187
& 0.238
&\textbf{0.427}
& 0.205
 &0.251\\
  Dialog Breakdown Analysis&0.566
& 0.223
&0.617
&0.727
&0.689
&0.245
 &\textbf{0.744}\\
 Connectedness Presence& 0.632
& 0.555
& 0.602
& 0.566
& \textbf{0.688}
& 0.602
 &0.633\\
 Connectedness Classification& 0.366& 0.303& 0.286& 0.324& \textbf{0.371}& 0.338&0.302\\
Language Toxicity Detection&\textbf{0.471}& 0.307&0.389&0.407&0.390&0.366&0.340\\
Criticism Stance Detection& 0.540& 0.510&0.520& 0.550&0.570& 0.500&\textbf{0.580}\\
 Sentiment Analysis&0.918& 0.901&0.900&0.885&0.822&0.900&\textbf{0.935}\\
\bottomrule
\end{tabular}
\begin{tablenotes}
    \footnotesize
    \setlength{\baselineskip}{\normalbaselineskip}
    \item \textit{Notes.} The values represent Cohen's Kappa scores between LLM annotations and human reference labels. Codes follow provider-openness-index naming, where CW denotes closed-weight and OW denotes open-weight. Performance varies by task type, and no single model is uniformly strongest across contexts. Binary and multi-class tasks generally yield higher and more stable agreement, while multi-label tasks exhibit lower and more variable agreement.
\end{tablenotes}
\end{threeparttable}
\end{table}
}

Second, LLM suitability is task-dependent.
The empirical evidence from our comprehensive evaluation presents a subtle picture of LLM viability for text annotation tasks, instead of a binary verdict on LLM suitability. In other words, there is a relationship between task structure and annotation performance (measurement error).
For binary classification tasks, such as sentiment analysis, connectedness presence, and criticism stance detection, LLMs perform reliably well. Across almost all models, Cohen's Kappa scores against human annotations exceed 0.5, with the best-performing models in each task approaching or surpassing 0.6.
In multi-class classification tasks where each document receives a single label from mutually exclusive options, some models perform well, but variability across models is large. For business proposal evaluation, better-performing models reach $\kappa \approx 0.5$ to $0.6$, whereas for dialog breakdown analysis, performance spans roughly $\kappa \approx 0.2$ to $0.7$ across models. This dispersion indicates that current LLMs can address moderately complex categorization, yet reliability is highly task- and model-dependent, thereby necessitating validation on the focal task.
Finally, LLMs face significant challenges in multi-label classification tasks, where instances can belong to multiple categories simultaneously. Examples include dialog intent classification, review attribute detection, and connectedness classification. Performance in these tasks drops considerably. Dialog intent classification ($\kappa\approx 0.2$), review attribute detection ($\kappa\approx 0.4$), and connectedness classification ($\kappa\approx 0.3$) all demonstrate substantially lower agreement with expert annotations. This low performance suggests a substantial limitation in LLMs' ability to handle nuanced, overlapping, and multidimensional labeling schemes.

These results suggest that researchers should adopt a decision framework based on task structure when considering LLM implementation for annotation tasks. Binary and simple multi-class classification tasks represent ``low-hanging fruit'' where LLMs can reliably scale human annotation efforts. In contrast, complex multi-label classification tasks remain better suited to human annotation or may require more sophisticated annotation architectures that combine human expertise with LLM capabilities. Together, our findings caution against viewing LLMs as universal replacements for human annotation.\footnote{In addition, we applied five additional models: four more advanced ones (AN-CW2, GO-CW2, OA-CW3, and OA-CW4) and one less advanced model (ME-OW2). Models from the same series exhibit some improvement across versions (e.g., AN-CW2 relative to AN-CW1), though the performance gains are generally modest. The main conclusions remain robust across all models. Detailed results are reported in Table~\ref{tab:sum_model_add}.}

\subsubsection{Simulation.}\label{sec:simulation}
The preceding results demonstrate that SILICON reduces measurement error across various annotation tasks. A natural question follows: how does this reduction in measurement error affect the quality of downstream statistical inference? To quantify this, we conduct Monte Carlo simulations that examine the propagation of annotation error to regression-based hypothesis tests, a common downstream use case of annotated data in management research.

We consider the downstream inference task as a linear regression where the LLM-annotated binary classification variable serves as the explanatory variable, denoted as $y$. The true data-generating process is:
\begin{equation}
z_i = \beta y_i + u_i, \quad u_i \overset{\text{iid}}{\sim} N(0, \sigma^2),
\label{eq:dgp}
\end{equation}
where $z_i$ is a continuous outcome, $y_i \in \{0, 1\}$ is the unobserved ground truth label, $\beta$ is the true effect of interest, and $u_i$ is idiosyncratic noise. Since $y_i$ is unobserved, researchers instead estimate:
\begin{equation}
z_i = \gamma_0 + \gamma_1 \hat{y}_i + \epsilon_i,
\label{eq:ols}
\end{equation}
where $\hat{y}_i$ is the LLM-generated label from the selected configuration.

The simulation follows the sequential design of SILICON implementation: a human reference baseline is first established under fixed $(G,H)$, then multiple candidate LLM configurations, each representing a different $(P,M)$ combination, are evaluated against this reference, the best-performing configuration is selected based on observable $R$, and the selected configuration is deployed to generate labels for downstream regression analysis.

We map the conceptual components of a configuration $c = (G, H, P, M)$ to simulation parameters as follows: (1) human reference quality, determined jointly by guidelines $G$ and human baseline procedures $H$, is captured by the human error rate $e_h$ (the error rate $e$ from Assumption~\ref{assum:sln}), governing how often the reference label $r$ deviates from truth $y$.
(2) LLM annotation quality, determined by $P$ and $M$, is captured by the model error rate $e$ of a symmetric confusion matrix: $\Pr(\hat{y} \neq k \mid y = k) = e$, with errors distributed uniformly across the remaining $K-1$ classes. Lower $e$ corresponds to a more accurate prompt/model configuration.
(3) The tendency for shared error modes between the LLM and human, which reflects the co-labeling covariance ($J$), is captured by a coupling parameter $\rho$, the probability that the LLM copies the human reference label. We present an overview of the simulation procedure as follows, with the complete algorithm and parameter specifications provided in Appendix~\ref{app:simulation_procedure}.

{\OneAndAHalfSpacedXI
\begin{tcolorbox}[
    colback=white,
    colframe=black!75!white,
    arc=3mm,
    title={\textbf{Monte Carlo Simulation Procedure}},
    boxrule=1pt,
    left=10pt,
    right=10pt,
    top=8pt,
    bottom=8pt
]
Each replicate proceeds through six steps:
\begin{enumerate}
    \item \textbf{Generate ground truth.} Draw true labels $y_i$ for validation and test sets from balanced class probabilities.
    \item \textbf{Create human reference.} Generate reference labels $r_i$ by flipping each true label to a random alternative with probability $e_h$.
    \item \textbf{Generate LLM candidates.} For each of $M$ candidate configurations (denoted by $m$), produce predictions from the confusion matrix (parameterized by $e^{(m)}$), then with probability $\rho^{(m)}$, overwrite with the human reference to induce coupling.
    \item \textbf{Select best configuration.} Compute reference agreement $R^{(m)}$ for each candidate and select $m^* = \arg\max_m R^{(m)}$.
    \item \textbf{Deploy on test set.} Apply the selected configuration to the test set and compute true agreement $T^{(m^*)}$.
    \item \textbf{Run downstream inference.} Generate outcomes $z_i = \beta y_i + u_i$, regress on the selected labels via OLS, and record $\hat{\beta}$, CI coverage, and selection correctness.
\end{enumerate}
\end{tcolorbox}
}

We parameterize two regimes that differ in human reference quality and candidate model pools, while sharing the same coupling structure. Two factors contribute to a configuration's confusion matrix: (1) inherent model/prompt capability and (2) the guidelines effect (refined vs.\ one-shot). Because both regimes evaluate the same set of models, factor (1) is identical across regimes; only factor (2) differs. We therefore draw the candidate accuracy distributions with equal width and allow them to overlap, so that the best baseline configurations may match the worst improved ones. The coupling parameter $\rho$, which governs shared error modes between the LLM and the human reference, is also set identically across regimes: it reflects model-level tendencies that do not depend on guidelines quality. The only regime-specific parameters are (i) the human error rate $e_h$, capturing reference quality, and (ii) the mean of the accuracy distribution, capturing the guidelines-induced shift in average performance.

\begin{figure}[htbp]
    \centering
        \caption{Simulation Results: Selection and Inference}
    \includegraphics[width=\textwidth]{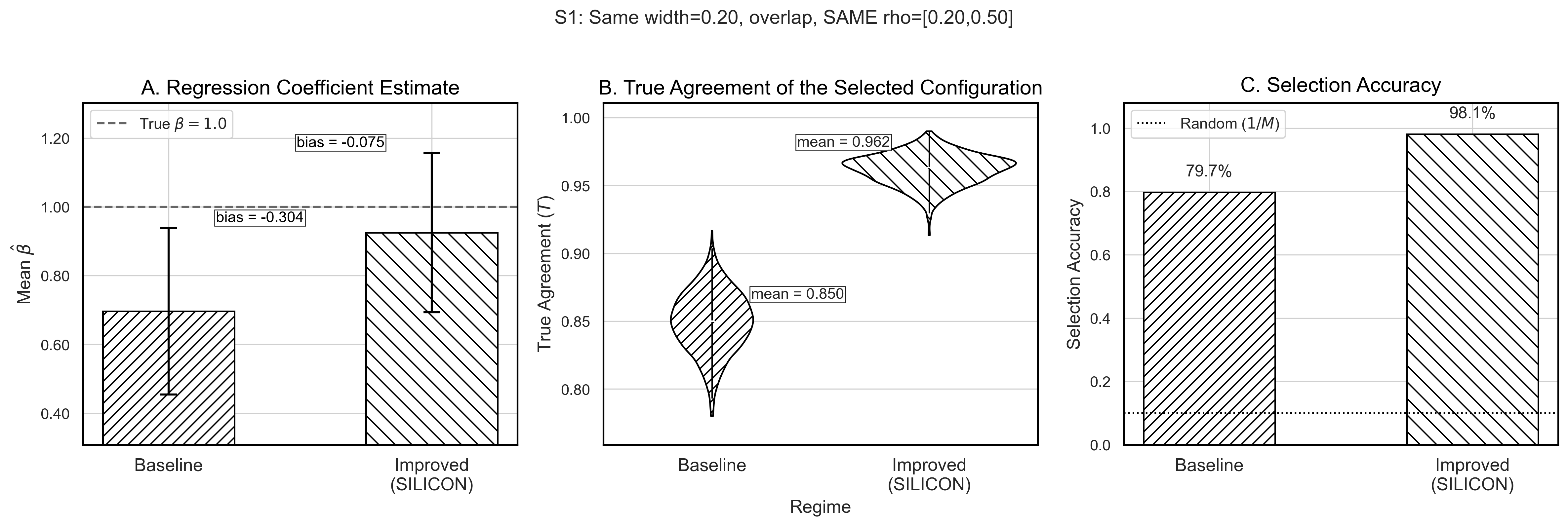}
    \begin{flushleft}
        \footnotesize \parbox{1.0\linewidth}{
        \textit{Notes.} Panel~A: Mean regression coefficient $\hat{\beta}$ with 95\% Monte Carlo CI; the dashed line indicates the true value $\beta = 1.0$. Panel~B: Distribution of the selected LLM configuration's true agreement ($T$) on the test set. Panel~C: Selection accuracy, defined as the proportion of replicates in which reference-agreement-based ($R$) selection identifies the configuration with the highest true agreement ($T$). The dotted line indicates the accuracy of random selection ($1/M$), where $M=10$ is the number of candidate LLM configurations compared against the human reference.
        Both regimes use the same coupling $\rho \in [0.20, 0.50]$ and accuracy distributions of equal width $0.20$, with overlap in $[0.75, 0.85]$. Baseline: $e_h = 0.15$, model accuracy $\in [0.65, 0.85]$. Improved (SILICON): $e_h = 0.02$, model accuracy $\in [0.75, 0.95]$.
        }
    \end{flushleft}
    \label{fig:simulation_fig1}
\end{figure}

Concretely, we compare two regimes across $1{,}000$ replicates: a \emph{baseline} regime representing conventional practice (human error rate $e_h = 0.15$, model accuracy drawn uniformly from $[0.65, 0.85]$, coupling $\rho \in [0.20, 0.50]$) and an \emph{improved (SILICON)} regime reflecting systematic error reduction ($e_h = 0.02$, model accuracy from $[0.75, 0.95]$, same coupling $\rho \in [0.20, 0.50]$). Both accuracy distributions have width $0.20$, with an overlap region $[0.75, 0.85]$. Figure~\ref{fig:simulation_fig1} presents the results. Panel~A shows that the baseline regime suffers severe attenuation bias: mean $\hat{\beta} = 0.696$ versus true $\beta = 1.0$ (bias $= -0.304$), with 95\% CI coverage collapsing to 31.7\%. The improved regime reduces bias to $-0.075$ (a 75\% reduction) and raises coverage to 89.6\%. Panel~B confirms that the improved regime consistently selects higher-quality configurations (mean $T = 0.962$ vs.\ $0.850$). Panel~C shows that selection accuracy improves dramatically from 79.7\% to 98.1\%---the cleaner reference ($e_h = 0.02$) makes observed agreement $R$ a far more reliable proxy for true agreement $T$, enabling near-perfect identification of the best candidate even when the accuracy pools overlap substantially.

We conduct extensive sensitivity analyses in Appendix~\ref{appendix:simulation_sensitivity} to verify robustness. These include scenarios with (i) nearly identical accuracy pools (overlap covering most of the distribution), (ii) asymmetric coupling where SILICON has lower $\rho$, and (iii) very low coupling for both regimes. Across all configurations, SILICON reduces bias by 36\%--75\% and at least doubles CI coverage relative to the baseline.

\subsection{Robust Reproducibility}\label{sec:robust_repro_results}

\subsubsection{Open-weight baseline test.} \label{sec:result_regression}

\begin{figure} [h]
    \centering
     \caption{Regression-based Performance Comparison Across Models}
    \includegraphics[width=1\linewidth]{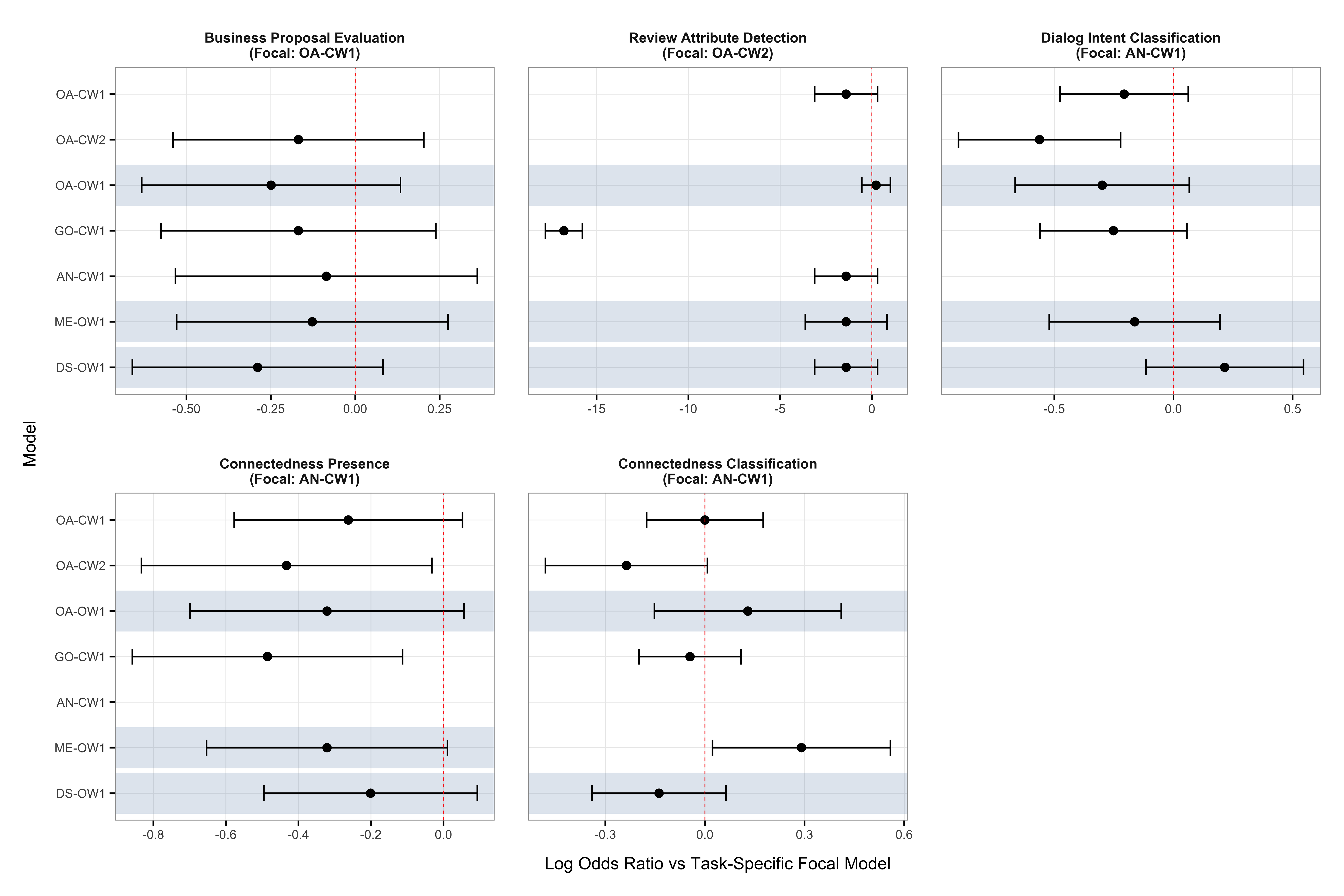}

   \begin{flushleft}
        \footnotesize \parbox{\linewidth}{
     \textit{Notes.} Regression-based comparison of model performance relative to the task-specific focal model (red dashed line, reference). The blue ribbons highlight all open-weight models whose 95\% confidence intervals include zero, indicating no statistically detectable difference from the focal model. Each point represents the estimated log-odds ratio of accuracy for a given model, with horizontal bars indicating 95\% confidence intervals. Points to the right of zero suggest higher relative performance compared to the focal model, while points to the left indicate lower relative performance. Dialog Breakdown Analysis is omitted because its best-performing model is already open-weight. Missing estimates occur when no exact matches exist between model annotations and the reference baseline, which is rare but possible for multi-label classification tasks. Because the dependent variable is item-level accuracy rather than Cohen's Kappa, model rankings in this figure can differ from the Kappa-based rankings used for focal model selection; see the accompanying footnote in the main text for a detailed explanation.
        }
    \end{flushleft}
    \label{fig:regression}
\end{figure}

Figure~\ref{fig:regression} shows regression-based comparison results, including coefficient estimates and 95\% confidence intervals for the five primary tasks whose best-performing model is closed. Each panel uses the best-performing model for that task as the omitted baseline. The dependent variable is whether a given label matches the expert ground truth. Because the best-performing model on each task serves as the omitted baseline, each coefficient represents how much more or less likely an alternative model is to produce a correct label, measured in log-odds, relative to that focal model on the same set of items.\footnote{The focal model is selected by Cohen's Kappa (weighted for multi-label tasks), whereas the regression evaluates item-level accuracy. These metrics serve complementary purposes and can rank models differently; see Appendix~\ref{app:fractional_logit} for a detailed discussion and robustness checks.} The blue ribbons highlight all open-weight models whose 95\% confidence intervals include zero and therefore do not statistically differ from the focal model. Dialog Breakdown Analysis is omitted from the figure because its best-performing model, DS-OW1, is already open-weight and therefore does not require an open-weight baseline test. Intervals that cross zero indicate that we cannot reject the null hypothesis of no performance difference relative to the focal model; intervals below zero indicate significantly worse performance, and above zero indicate significantly better performance.

We find that every task whose best-performing model is closed has at least one open-weight alternative with no statistically detectable difference from that focal model. Dialog Breakdown Analysis goes one step further: its best-performing model is DS-OW1, which is already open-weight, so robust reproducibility is satisfied by construction for that task. Together with the ensemble upper-bound analysis that follows, these results bracket the focal model's performance: the open-weight baseline guarantees reproducibility via a permanently accessible alternative, while the routed-ensemble results show how much improvement is empirically obtainable under stronger labeling support.

\subsubsection{Upper bound by ensemble performance.}
As described in Section~\ref{sec:multi-llm}, we construct an ensemble-based upper bound on the focal model's performance by augmenting the best-performing model on each task with the second- and third-best models from Table~\ref{tab:sum_model}. Specifically, we evaluate a routing rule in which items with $\mathrm{FSD}_i < \tau$ are sent to auxiliaries, while high-confidence items retain the focal model's label.\footnote{For $\tau=0$, performance equals the focal model alone; at $\tau=1$, all items are assigned via task-specific three-model majority vote.} Because the focal model is already the best-performing single model, full integration does not necessarily maximize performance: an interior threshold can outperform $\tau=1$ when auxiliary votes improve low-confidence items but would otherwise overwrite correct high-confidence focal labels. The upper bound for a task is therefore the highest observed point on its routing curve.

\begin{figure}[h!]
    \centering
    \caption{Ensemble Upper Bound: Performance of Multi-Model Annotation across Confidence Threshold}
       \label{fig:multi-llm-across-fsd}
    \begin{minipage}{1.0\textwidth}
        \centering
        \includegraphics[width=1\linewidth]{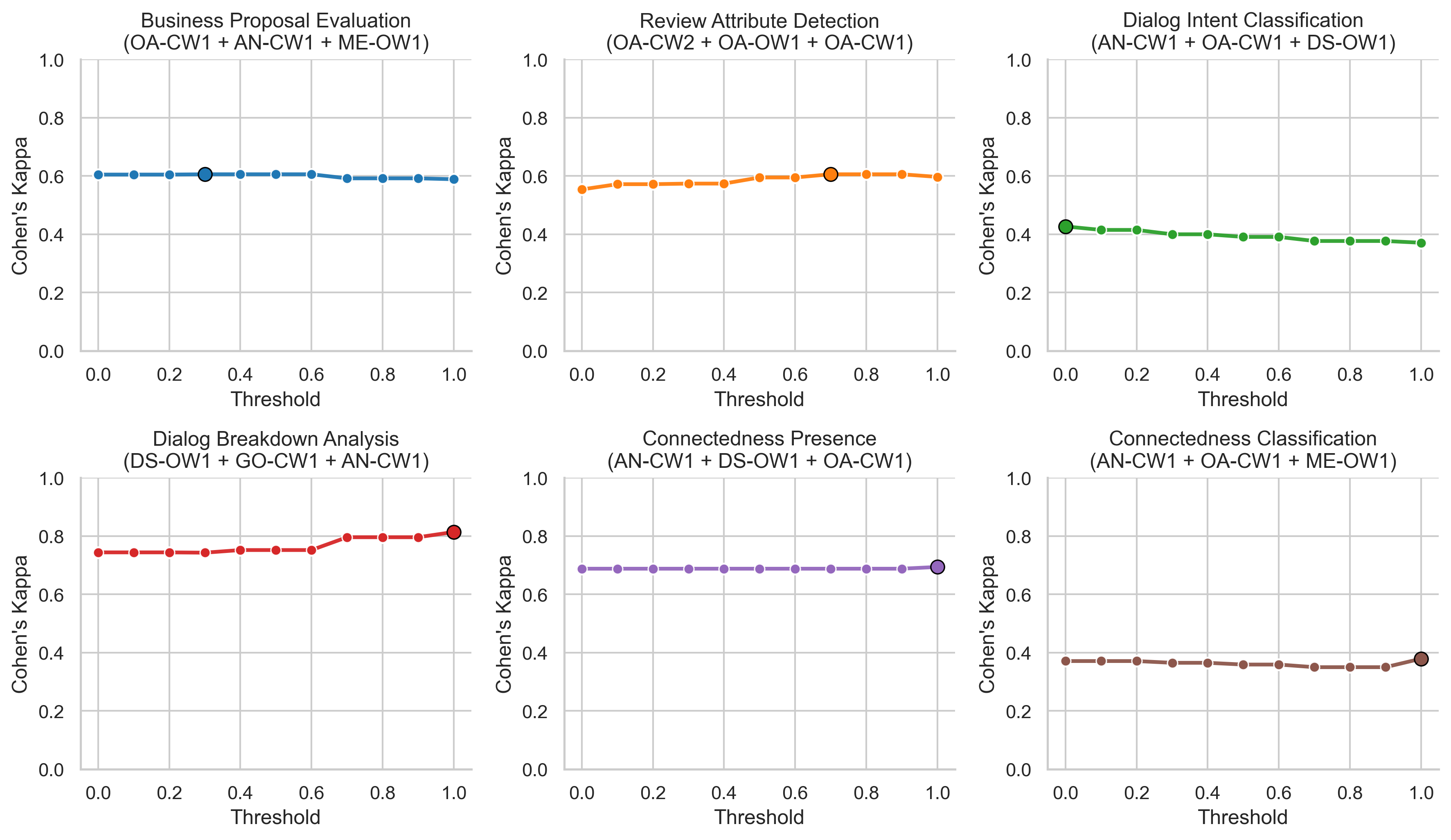}

    \end{minipage}

    \begin{flushleft}
        \footnotesize \parbox{1.0\linewidth}{
        \textit{Notes.} Cohen's Kappa vs. FSD routing threshold $\tau$ across six tasks when each task uses its own best-performing model as the focal model and the next two best models as auxiliaries. At $\tau=0$, only the focal model is used; at $\tau=1$, all items are assigned via task-specific three-model majority vote. The curves therefore quantify how routing changes performance across candidate ensemble configurations. The larger dot on each curve marks the threshold that yields the highest Cohen's Kappa for that task.
        }
    \end{flushleft}

\end{figure}

Figure~\ref{fig:multi-llm-across-fsd} plots Cohen's Kappa against human reference labels as a function of confidence threshold $\tau$ across six tasks.
The routing curves reveal that whether model aggregation actually improves performance is highly task-dependent. For Dialog Breakdown Analysis, performance of the model ensemble rises steadily through full integration, suggesting that GO-CW1 and AN-CW1 contribute complementary signals to the DS-OW1 focal model. However, Dialog Intent Classification illustrates the opposite extreme: it displays a downward-sloping routing curve, meaning auxiliary models actively degrade performance relative to the focal model. This is consistent with a setting where the auxiliary models share error modes with each other but not with the focal model, so that majority voting overrides correct focal labels with incorrect auxiliary predictions.

The curves also suggest that, in certain scenarios, additional labels should be introduced selectively to achieve the best outcome. The non-monotonic patterns in Business Proposal Evaluation and Review Attribute Detection confirm that the best ensemble configuration can occur at an interior routing threshold rather than at full integration, reinforcing that researchers should empirically calibrate the threshold rather than defaulting to majority vote on all items.

Finally, the routing curves inform the practical tradeoff between performance and cost. In tasks where aggregate performance increases sharply up to a certain threshold and then improves only marginally, the cost of routing additional items to multiple models becomes a relevant factor. Researchers must weigh whether the additional labeling cost is justified by the relatively small gain in overall performance beyond that threshold. When the curve is flat, as in Connectedness Presence, the additional cost is unwarranted. The decision to invest in multi-model labeling should therefore be guided by the routing curve on the validation sample before committing to multi-model deployment at scale.
To determine whether the best routed ensemble is statistically different from the focal model, researchers can apply the regression-based comparison by treating the selected routed ensemble as an additional treatment on the same items. This provides a formal test of whether the observed ensemble gain is statistically distinguishable from the focal model's performance.

\section{Discussion and Conclusion}

This paper addresses a challenge that is both urgent and underappreciated: the LLM used to annotate data today can be retired without warning, potentially before the research clears the rigorous academic review process. The speed at which frontier models are being retired makes robust reproducibility not just a future problem, but a present one.\footnote{For instance, OpenAI retired OA-CW1 (GPT-4o) from the user interface in 2025, unretired it, and finally re-retired it in 2026 (\href{https://openai.com/index/retiring-gpt-4o-and-older-models/}{link}, accessed March 25, 2026).} However, robust reproducibility is only meaningful if the performance signal is free from contamination by measurement error. We therefore show that the first step toward robust reproducibility is decomposing and controlling measurement error across the annotation pipeline, so that any discrepancy under a substitute model reflects a genuine model difference rather than uncontrolled noise from guidelines, baselines, or prompts. With measurement error controlled, we operationalize robust reproducibility as a lack of statistical difference, choose the best-performing model on each task as the focal model, and then establish an open-weight baseline together with an upper-bound comparison via ensemble-based multi-model labeling. Empirically, structured interventions at each stage substantially reduce error, and regression-based comparison validates that each closed focal model has at least one open-weight replacement with no statistically detectable performance difference. The routing analysis further shows that multi-model labeling is selectively useful because the best ensemble configuration can occur at an interior routing threshold rather than at full integration. In practice, researchers may choose a cheaper model instead of the best-performing one; if that chosen model is closed, they should still run the same open-weight baseline test before relying on it. While our focus is management research, the framework applies to any discipline preferring prescriptive annotation.

\subsection{Managerial Implications}\label{sec:managerial_implications}
Our findings have immediate relevance for organizations that rely on text classification for operational decisions. In practice, organizations run recurring text-classification workloads across diverse functions: customer service departments analyze support tickets and chat transcripts to identify emerging issues, human resources screen resumes and employee feedback for cultural fit and satisfaction indicators, marketing teams mine social media and product reviews for brand perception, and compliance units monitor internal communications for regulatory risks. SILICON provides these organizations with a systematic approach to deploy LLM-based annotation that maintains quality and long-term reproducibility.

The workflow's parsimonious design makes it particularly suitable for organizational adoption. Rather than requiring annotation of entire datasets, SILICON's validation approach uses representative samples of 100 to 200 documents to assess annotation quality, a scale that fits within typical pilot project budgets. For instance, a firm seeking to classify customer complaints can first apply SILICON to iteratively refine guidelines with domain experts from customer service, establish baselines using these experts rather than generic crowd workers, and test multiple LLMs on a sample before committing to full-scale deployment. Our results show that open-weight models achieve no statistically detectable difference in performance relative to closed models in the tasks we examined, offering organizations a path to avoid vendor lock-in while maintaining annotation quality. While LLMs are already more cost- and time-efficient than human annotation, SILICON extends these gains: ensemble-based multi-model labeling establishes performance bounds, and open-weight model substitution ensures long-term accessibility, all while ensuring robust reproducibility. By decomposing measurement error into interpretable sources, SILICON also provides actionable diagnostics: if annotation quality is poor, organizations can identify whether the issue stems from ambiguous guidelines, inadequate baselines, suboptimal prompts, or model limitations, enabling targeted improvements rather than costly trial-and-error.

\subsection{Limitations}\label{sec:limitations}
There are several limitations that warrant future research. First, our study adopts a prescriptive approach to address annotation tasks. The workflow assumes that high inter-annotator agreement is a desirable standard, which may not apply in research contexts that intentionally seek to capture variability in human perception or interpretation. In such settings, enforcing LLM alignment with a single ``ground truth'' may obscure meaningful heterogeneity. Furthermore, reasoning models may be especially desirable in such settings, compared to chat completion models. Future work could explore how LLMs can support, rather than suppress, annotation plurality, particularly when diversity of opinion is part of the theoretical construct being studied.
Second, while we emphasize expert-developed baselines, we do not account for the individual-level characteristics of annotators. Prior research suggests that LLMs can be tuned to emulate specific annotator styles or preferences \citep{hashemi_llm-rubric_2024}, opening the door to more personalized or demographic-aware annotation pipelines. Incorporating these perspectives remains an important direction for future methodological development.

	\putbib
	\end{bibunit}

\newpage
\begin{bibunit}
\begin{APPENDICES}

\begin{center}
            \large  \textbf{Online Appendices \\
          To Err Is Human; To Annotate, SILICON? Toward Robust Reproducibility in LLM Annotation}
        \end{center}

\setcounter{page}{1}
\setcounter{table}{0}
\renewcommand{\thetable}{A\arabic{table}}

\setcounter{figure}{0}
\renewcommand{\thefigure}{A\arabic{figure}}

\section{Methodological Details}
\subsection{Proof of Lemma \ref{lemma:bias-aware}} \label{appendix:proof}

Write $\pi_k=\Pr(y=k)$ and, for each configuration $c$, define
\[
p_{\ell\mid k}(c):=\Pr(\hat y_c=\ell\mid y=k),\qquad
q_{\ell\mid k}(c):=\Pr(r=\ell\mid y=k,c).
\]
Because human annotators label items under fixed $(G,H)$ without observing which prompt or model is being evaluated, the reference label satisfies $r\perp c\mid(x,G,H)$. Hence for any $k,\ell$, we have:
\[
\begin{aligned}
q_{\ell\mid k}(c)
&= \Pr(r=\ell\mid y=k,c)
= \mathbb E\!\left[\Pr(r=\ell\mid y=k,x,G,H,c)\mid y=k\right]\\
&= \mathbb E\!\left[\Pr(r=\ell\mid y=k,x,G,H)\mid y=k\right]
=: q_{\ell\mid k},
\end{aligned}
\]
which shows that the conditional reference probabilities $q_{\ell\mid k}$ (and thus the reference error rate $e=\Pr(r\neq y)$) are \emph{invariant across $c$}.
Now we decompose $R(c)$. For any $c$,
\[
R(c)
= \sum_{k=1}^K \pi_k \Pr(\hat y_c=r\mid y=k)
= \sum_{k=1}^K \pi_k \sum_{\ell=1}^K \Pr(\hat y_c=\ell,\,r=\ell\mid y=k).
\]
Add and subtract $p_{\ell\mid k}(c)\,q_{\ell\mid k}$ inside the inner sum:
\begin{equation} \label{eq:pr_yhat=cgivenk}
    \Pr(\hat y_c=r\mid y=k)
= \sum_{\ell=1}^K p_{\ell\mid k}(c)\,q_{\ell\mid k}
+ \sum_{\ell=1}^K\Big(\Pr(\hat y_c=\ell, r=\ell\mid y=k)-p_{\ell\mid k}(c)\,q_{\ell\mid k}\Big).
\end{equation}
By Assumption~\ref{assum:sln}, $q_{k\mid k}=1-e$ and $q_{\ell\mid k}=e/(K-1)$ for $\ell\neq k$. Therefore,
\begin{equation} \label{eq:sumpq}
    \sum_{\ell=1}^K p_{\ell\mid k}(c)\,q_{\ell\mid k}
=(1-e)\,p_{k\mid k}(c)+\frac{e}{K-1}\sum_{\ell\neq k} p_{\ell\mid k}(c)
=\Big((1-e)-\frac{e}{K-1}\Big)\, p_{k\mid k}(c) + \frac{e}{K-1}.
\end{equation}
Substituting Equation~\eqref{eq:sumpq} back into Equation~\eqref{eq:pr_yhat=cgivenk}, we have:
\[
\begin{aligned}
R(c)
&= \sum_{k=1}^K \pi_k \Big[ \Big((1-e)-\frac{e}{K-1}\Big)\, p_{k\mid k}(c) + \frac{e}{K-1} \Big]
+ \sum_{k=1}^K \pi_k \sum_{\ell=1}^K\Big(\Pr(\hat y_c=\ell, r=\ell\mid y=k)-p_{\ell\mid k}(c)\,q_{\ell\mid k}\Big) \\
&= a \sum_{k=1}^K \pi_k\, p_{k\mid k}(c) + b
+ \underbrace{\sum_{k=1}^K \pi_k \sum_{\ell=1}^K\Big(\Pr(\hat y_c=\ell, r=\ell\mid y=k)-p_{\ell\mid k}(c)\,q_{\ell\mid k}\Big)}_{=:~J(c)},
\end{aligned}
\]
with $a=(1-e)-\frac{e}{K-1}>0$ and $b=\frac{e}{K-1}$. Recognizing $\sum_k \pi_k\, p_{k\mid k}(c)=\Pr(\hat y_c=y)=T(c)$ completes the identity
\[
R(c)=a\,T(c)+b+J(c).
\]
Note that the term $J(c)$ captures conditional dependence between $\hat y_c$ and $r$ given $y$; if they were conditionally independent, $J(c)$ would be $0$ and the mapping would reduce to the linear form.
\vspace{2em}
\subsection{Measurement of Agreement Rate: Cohen's Kappa} \label{appendix:cohenkappa}

To measure agreement, one simple way is counting the raw number of matching annotations. However, the raw agreement fails to account for agreements that might occur by chance. This can lead to misleading conclusions, especially in cases where the likelihood of random agreement is high \citep{lee_common_2023}.
To overcome this limitation, we employed Cohen's Kappa ($\kappa$), which measures the level of agreement between two annotators while adjusting for chance agreement \citep{cohen_weighted_1968}. The formula for Cohen's Kappa is given by:
\begin{align}
    \kappa = \frac{P_{o} - P_e}{1 - P_e},
\end{align}
where \(P_o\) is the relative observed agreement among raters, and \(P_e\) is the hypothetical probability of agreement by chance.
Specifically, $\kappa=1$ indicates perfect agreement; $\kappa=0$ indicates agreement no better than chance; and $\kappa<0$ suggests less agreement than would be expected by chance. A $\kappa$ of 0.4 to 0.6 is commonly regarded as a threshold for sufficient inter-annotator agreement \citep{landis_measurement_1977}.

In our multi-label classification task, annotators could assign multiple distinct labels (referred to as ``units") to a single item (e.g., a social media post). We define the overall label(s) assigned to an item as a ``set". For instance, in the toxicity detection task, there are in total three units,  ``fearspeech", ``hatespeech", and ``normal", and a set could be ``fearspeech, hatespeech".
It is apparent that the agreement level between ``fearspeech, hatespeech" and ``fearspeech'' should be higher than that between ``hatespeech" and ``fearspeech''.
To account for this, we use Weighted Cohen's Kappa \citep{cohen_weighted_1968}, which is given by
\begin{align}
    \kappa = 1 - \frac{\sum_{i=1}^{k} \sum_{j=1}^{k} w_{ij} x_{ij}}{\sum_{i=1}^{k} \sum_{j=1}^{k} w_{ij} m_{ij}},
\end{align}
where \( w_{ij} \) is the weight matrix, \( x_{ij} \) is the observed matrix, and \( m_{ij} \) is the expected matrix.

We follow \cite{passonneau_measuring_2006} to derive the weight matrix $W$ for our calculations. The weight $w$ ranges from 0 (identical sets) to 1 (disjoint sets).
The weight $w$ between two sets $P$ and $Q$ is defined as:
\begin{equation}
    w=1-J\cdot M,
\end{equation}
where $J$ is the Jaccard metric \citep{jaccard_nouvelles_1908} and $M$ represents monotonicity.
Specifically, $J$ measures the size difference between two sets, independently of their structural relationship. It is calculated as the ratio of the cardinality of the intersection to the cardinality of the union of the two sets.  $J$ ranges from 0 (disjoint sets) to 1 (identical sets).
The $M$ term captures the structural relationship between sets:  if two sets $Q$ and $P$ are identical, $M$ is 1; if one set is a subset of the other, $M$ is 2/3; if the intersection and the two set differences are all non-null, then $M$ is 1/3; if the sets are disjoint, $M$ is 0.
Altogether, the weight $w$ reflects both the size difference and the structural relationship between the two sets.
For instance, in the language toxicity detection task, the weight between sets ``fearspeech" and ``hatespeech" is 1, and the weight between ``fearspeech'' and ``fearspeech, hatespeech'' is 2/3.

\section{Regression-based Model Comparison}
\setcounter{table}{0}
\renewcommand{\thetable}{B\arabic{table}}

\setcounter{figure}{0}
\renewcommand{\thefigure}{B\arabic{figure}}

\subsection{A Regression-based Approach to Compare LLM Annotation Performance}\label{appendix:regression}

Consider the scenario where we use the same prompt across multiple models applied to the same text sample.\footnote{This approach in essence is about comparing LLM annotation performance across different treatments. Treatments could be different models or different prompts, and the same logic outlined here applies to both scenarios.}
Our goal is to statistically compare model performances and identify models with equivalent results. Specifically, we aim to determine whether a model's performance metric differs significantly from others or if certain models yield statistically indistinguishable performance metrics.

In this setting, there are two sources of uncertainty:
\begin{enumerate}
    \item \textit{Sampling uncertainty:} This arises from inferring population parameters (e.g., differences in LLM performance across models) based on the human baseline sample, which is randomly drawn from the human baseline.
    \item \textit{Stochastic output uncertainty:} This stems from the non-deterministic nature of LLM outputs. The literature (e.g., \citealt{pangakis_automated_2023}) documents this variability, and common approaches to account for this uncertainty include running the model multiple times and calculating the consistency of its annotation results.
\end{enumerate}

Our focus is on addressing the first source of uncertainty. Instead of computing overall performance metrics for each model and comparing them with pairwise tests or bootstraps, we propose a logistic regression-based approach.

Suppose we use $M$ models to annotate a sample with $I$ items (e.g., a sample consisting of $I$ business proposals). We treat each model as a distinct ``treatment'' and aim to assess the ``treatment effect'' of one model compared to a baseline model.
Denoting the model by $m$ (1, 2, ..., $M$) and the item by $i$ (1, 2, ..., $I$), the unit of analysis is the item-model pair, $im$.
The dependent Variable ($\mathit{Matched}_{im}$) is an indicator variable equal to 1 if model $m$'s label for item $i$ matches the human reference label, and 0 otherwise.\footnote{This binary “matched” measure is straightforward and easily interpretable at the item level. When aggregated across items, it provides the model’s accuracy. However, a slight discrepancy exists since in our case studies, we use Cohen's Kappa to account for chance agreement. This adjustment is made because Cohen's Kappa is a chance-corrected measure calculated over all items, which makes it less intuitive at the per-item level. Nonetheless, applying a Kappa-like transformation to derive the dependent variable would still align with the same regression logic.}

We then perform a logistic regression of the dependent variable  $\mathit{Matched}_{im}$  on all treatment dummies, excluding one due to perfect collinearity. The regression equation is:
\begin{equation}
  \text{logit}(P(\mathit{Matched}_{im}=1)) =\alpha_0+\alpha_1 \mathds{1}(m=1)+\alpha_2 \mathds{1}(m=2)+...+\alpha_{M-1}\mathds{1}(m=M-1).
\end{equation}
Here, the $M$-th treatment is omitted as the baseline. The coefficients are interpreted in terms of log-odds:
\begin{equation}
\begin{split}
\alpha_0 =& \log\left(\frac{P(Matched_{im}=1|m=M)}{1 - P(Matched_{im}=1|m=M)}\right), \\
\alpha_1 =& \log\left(\frac{P(Matched_{im}=1|m=1)}{1 - P(Matched_{im}=1|m=1)}\right) - \alpha_0, \\
\alpha_2 =& \log\left(\frac{P(Matched_{im}=1|m=2)}{1 - P(Matched_{im}=1|m=2)}\right) - \alpha_0, \\
\dots\\
\alpha_{M-1} =& \log\left(\frac{P(Matched_{im}=1|m=M-1)}{1 - P(Matched_{im}=1|m=M-1)}\right) - \alpha_0.
\end{split}
\end{equation}
Crucially, we cluster standard errors at the item level to account for the fact that each item $i$ is ``tested'' multiple times—once per model.
By design, the regression estimates represent differences in log-odds of matching the human reference label between each model and the baseline model. Hypotheses can be tested within this logistic framework, such as (1) joint test: $H_a$: $\alpha_1 = \alpha_2 = \dots = \alpha_{M-1} =0$, using a likelihood ratio test, and (2) individual tests: $H_{b1}$: $\alpha_1 = 0$, $H_{b2}$: $\alpha_2 = 0$, ..., $H_{b(M-1)}$: $\alpha_{M-1} = 0$.

Furthermore, under this framework, the second source of uncertainty (stochasticity in LLM outputs) can also be addressed with additional assumptions. Specifically, we consider that LLMs aim to produce an ``intended output''\footnote{One could argue that the intended output is the one with the highest probability of being generated by the LLM model when the temperature is set to 0. However, we take the stance that the intended output may not necessarily align with the output generated under this condition. This difference is not the focus of the study, and we leave its discussion for future work.} but sometimes generate a different ``actual output'' due to non-determinism.
From a regression perspective, this discrepancy can be treated as a form of measurement error. Assuming the measurement error is independently and identically distributed (i.i.d.), the regression estimates remain unbiased when the sample size is sufficiently large.

\subsection{Robustness: Fractional Logistic Regression with Weighted Agreement as Outcome Variable}\label{app:fractional_logit}

The focal model is selected by Cohen's Kappa (weighted for multi-label tasks), whereas the regression evaluates item-level accuracy. These metrics serve complementary purposes and can rank models differently for two reasons. First, for multi-label tasks, Kappa uses MASI-based partial-agreement weights, whereas the regression's binary dependent variable requires an exact match. Second, Kappa corrects for chance agreement based on marginal label distributions, whereas the regression credits every correct label equally regardless of its base rate. As a result, a model can appear to outperform the focal model in the regression without contradicting the Kappa-based selection: the model may produce more exact (or partial) matches on the evaluated items, yet receive a lower Kappa because its predicted label distribution inflates expected agreement by chance. This pattern is most pronounced for multi-label tasks such as Connectedness Classification, where ME-OW1 achieves the highest item-level accuracy but ranks third in Kappa. We view this divergence as informative rather than problematic: the regression directly tests whether a substitute model reproduces correct labels at the same rate as the focal model, which is the practical reproducibility question, while Kappa provides a chance-corrected summary that is more appropriate for overall model selection.

As a robustness check, we re-estimated all model comparison regressions using a weighted agreement dependent variable and fractional logistic regression. For each item, the dependent variable is a MASI-based weighted agreement score (continuous between 0 and 1), where exact matches receive 1.0, partial matches receive intermediate credit based on Jaccard similarity and set monotonicity \citep{passonneau_measuring_2006}, and complete disagreements receive 0.0. We estimate the model using quasi-maximum likelihood with a logit link \citep{papke_econometric_1996}, which is the appropriate estimator for fractional response variables. Standard errors remain clustered at the item level.

Figure~\ref{fig:fractional_logit} shows the results. For single-label tasks (Business Proposal Evaluation, Connectedness Presence), the binary and weighted dependent variables are identical, producing identical estimates. For multi-label tasks, three patterns emerge. First, no model coefficient changes sign: models that underperform (outperform) the focal model under binary accuracy continue to do so under weighted accuracy. Second, the weighted DV resolves quasi-complete separation in Review Attribute Detection, where binary exact-match rates are near zero for most models. Third, models with high partial-agreement rates---particularly ME-OW1 in Connectedness Classification---show amplified positive coefficients (from $+0.29$ to $+0.60$), reflecting their tendency to produce partially correct multi-label predictions. These results confirm that the qualitative conclusions of the open-weight baseline comparison are robust to the choice of agreement metric.

\begin{figure}[h]
    \centering
    \caption{Regression-based Performance Comparison Across Models (Fractional Logit, Weighted Agreement DV)}
    \includegraphics[width=1\linewidth]{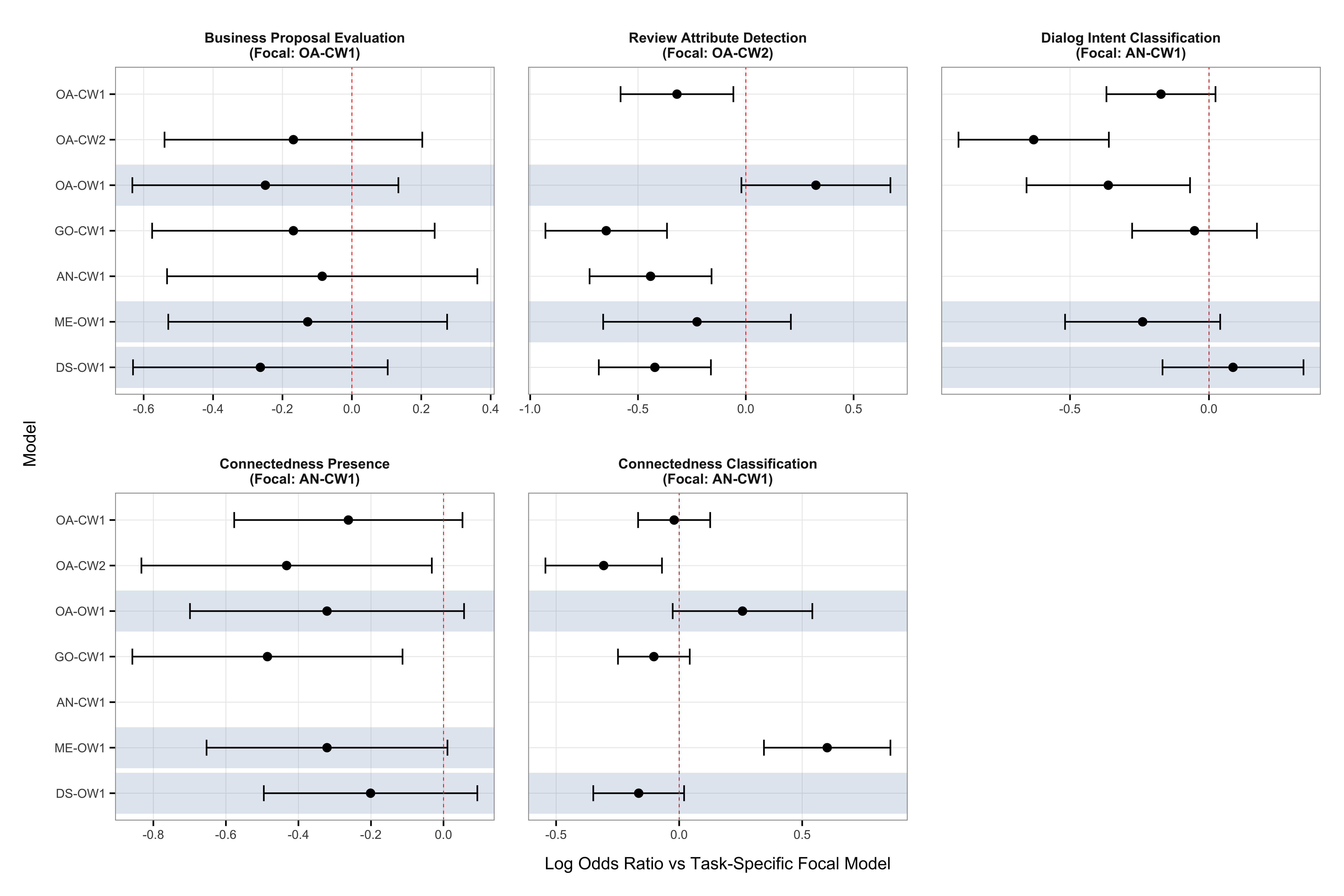}
    \begin{flushleft}
        \footnotesize \parbox{\linewidth}{
        \textit{Notes.} Same specification as Figure~\ref{fig:regression}, except the dependent variable is the MASI-based weighted agreement score (continuous 0--1) and estimation uses fractional logistic regression (quasibinomial family with logit link). Blue ribbons indicate open-weight models whose 95\% confidence intervals include zero.
        }
    \end{flushleft}
    \label{fig:fractional_logit}
\end{figure}

\section{Ensemble-based Performance Ceiling Details}
\setcounter{table}{0}
\renewcommand{\thetable}{C\arabic{table}}

\setcounter{figure}{0}
\renewcommand{\thefigure}{C\arabic{figure}}

\subsection{Computing \textit{FSD} and Two Practical Approaches}\label{app:fsd}
FSD evaluates LLM confidence (or consistency) by measuring the discrepancy in agreement between the most common and second most common answers from multiple sampled outputs. A larger FSD indicates higher confidence, indicating that the model’s responses are more concentrated on a single answer over multiple runs.
We summarize two ways to obtain the required distribution in practice.

\begin{enumerate}\setlength{\itemsep}{2pt}
\item \textbf{Log-probability approach (when available).} If token log-probabilities are exposed (common for open-weight models and some closed APIs), infer the top two option probabilities from the first disambiguating token of each option and compute FSD from those. There are some limitations to this approach:  (i) some APIs do not expose log-probs; (ii) with very large or combinatorial label spaces (e.g., multi-label), mapping options to first tokens becomes cumbersome.
\item \textbf{Sampling approach (model-agnostic).} Sample multiple independent outputs from the focal model for each item, form the empirical distribution over options, and compute FSD directly from observed frequencies. This resolves both limitations above at the expense of additional cost.
\end{enumerate}

\subsection{Confidence-Based Routing: Implementation Details}\label{app:fsd_routing}

The ensemble upper bound described in Sections~\ref{sec:multi-llm} and~\ref{sec:robust_repro_results} is operationalized by sweeping the routing threshold across candidate values and evaluating the resulting routed ensemble. This subsection provides the implementation details for that routing procedure.

\textbf{FSD computation.} For each item, we generate five independent responses from the focal model using temperature = 1, then calculate the First-Second Distance as the difference between the frequency of the most common response and the frequency of the second most common response. A larger FSD indicates higher confidence, suggesting the model's responses concentrate on a single answer across multiple runs.

 \textbf{Threshold calibration.} Items with $\mathrm{FSD}_i < \tau$ are routed to auxiliary models for majority voting, while high-confidence items retain focal-model labels. We recommend empirical calibration on the expert-annotated validation sample: evaluate Cohen's Kappa over a grid of threshold values and identify the threshold that yields the highest observed performance for the task-specific ensemble configuration. Because the focal model is already the strongest single model, the best threshold may be an interior value rather than full integration.

 \textbf{Empirical findings.} Across six primary tasks, routing traces task-dependent paths over candidate ensemble configurations (see Figure~\ref{fig:multi-llm-across-fsd}). Some tasks show clear improvements at interior thresholds, while others remain flat or decline as more items are routed. Full integration is therefore one candidate configuration, not the definition of the upper bound. If researchers wish to assess whether the selected routed ensemble statistically differs from the focal model, they can apply the regression-based comparison described in Appendix~\ref{appendix:regression}.

\section{Case Study and Workflow Details}
\setcounter{table}{0}
\renewcommand{\thetable}{D\arabic{table}}

\setcounter{figure}{0}
\renewcommand{\thefigure}{D\arabic{figure}}

\subsection{Case Study Summary}
\label{app:sum_cases}

We present the research contexts, original annotation approach, our implementation, and key findings of the seven case studies in Table~\ref{tab:sum_cases}.
{\OneAndAHalfSpacedXI
\begin{table} \footnotesize
\centering
\caption{Summary of Case Studies}
\label{tab:sum_cases}
	\begin{minipage}{\columnwidth}
		\begin{center}
			 \begin{tabular}{>{\raggedright\arraybackslash}p{0.1\linewidth}>{\raggedright\arraybackslash}p{0.18\linewidth}>{\raggedright\arraybackslash}p{0.20\linewidth}>{\raggedright\arraybackslash}p{0.23\linewidth}>{\raggedright\arraybackslash}p{0.2\linewidth}}
    \toprule
     \textbf{Task}&
    \textbf{Research Context} & \textbf{Original Text Annotation Approach} & \textbf{SILICON Implementation} & \textbf{Key Findings} \\ \midrule

     Business Proposal Evaluation&Understanding DAO governance through systematic analysis of business proposal content and decision delegation patterns & Manual classification by RAs into six predefined categories & Developed expert-validated guidelines with 3 RAs; established baseline using 109 proposals; tested multiple LLM configurations & High-performing LLMs achieved moderate to high agreement with expert baseline\\ \hline

     Review Attribute Detection&Quantifying consumers’ purchasing decisions through systematic analysis of product review content & Training a deep learning model using 5,000 reviews annotated by Amazon Mechanical Turk workers & Developed expert-validated guidelines with 3 RAs; established baseline using 180 reviews; tested multiple LLM prompt and model configurations & 1. Moderate agreement achieved with expert baseline across selected models \newline 2. Significant performance variation observed between different LLMs for this complex multi-label task \\ \hline

     Dialog Intent Classification and Breakdown Analysis&Evaluating human-computer interactions in customer support through systematic dialog analysis & None. This is a novel evaluation of human-computer interactions in customer support through systematic dialog analysis & Developed expert-validated guidelines with 3 RAs; established baseline using 195 conversational turns; tested multiple LLM prompt and model configurations & 1. Intent classification showed consistently low agreement between LLMs and expert baseline \newline 2. Breakdown analysis demonstrated moderate to high agreement levels across tested models \\ \hline

     Affective Content Evaluation&Understanding how a sense of connectedness affects consumer willingness to pay and product sales performance & Ongoing work involving manual annotations & Developed expert-validated guidelines with three research assistants; established a baseline using 120 product descriptions; tested multiple LLM prompting and model configurations & 1. Moderate to high agreement with expert baseline achieved for connectedness presence detection \newline 2. Low agreement observed for connectedness classification across detailed categories \\ \hline

     Language Toxicity Detection&Demonstrating fear speech’s prevalence, influence, and subtlety compared to hate speech on social media platforms & Manual classification into one of four types using a combination of experts and Amazon Mechanical Turk workers & Used the annotation guidelines and a random sample of the human baseline from the original paper, which involves 160 posts; explored multiple prompts and models & Agreement level between LLMs and crowd worker baseline is moderate to low \\ \hline

     Criticism Stance Detection&Assessing online attention and criticism toward retracted scientific papers over time & Manual classification of Twitter posts to identify expressions of criticism toward specific scientific papers, using trained annotators & Used the annotation guidelines and a random sample of the human baseline from the original paper consisting of 200 posts; explored multiple prompts and models & Agreement level between LLMs and crowd worker baseline is moderate \\ \hline

     Sentiment Analysis&Comparing human and computational attention in text classification tasks, focusing on sentiment analysis & Manual labeling of sentiment value and human attention maps of Yelp customer reviews using Amazon Mechanical Turk workers & Used the annotation guidelines and a random sample of the human baseline from the original paper consisting of 120 reviews; explored multiple models & Agreement level between LLMs and crowd worker baseline is very high across all models tested \\ \bottomrule

    \end{tabular}
		\end{center}
	\end{minipage}
\end{table}
}

\newpage
\subsection{SILICON Workflow Details} \label{app:guideline}

Figure~\ref{fig:workflow} presents the SILICON workflow in detail. The workflow organizes how we (i) develop reliable human annotation guidelines, (ii) establish human baselines, and (iii) subsequently optimize meta-prompts and evaluate multiple LLMs.

\begin{figure}[h!]
    \centering
        \caption{SILICON: A Systematic Workflow for LLM-based Text Annotation}
    \includegraphics[width=.95\linewidth]{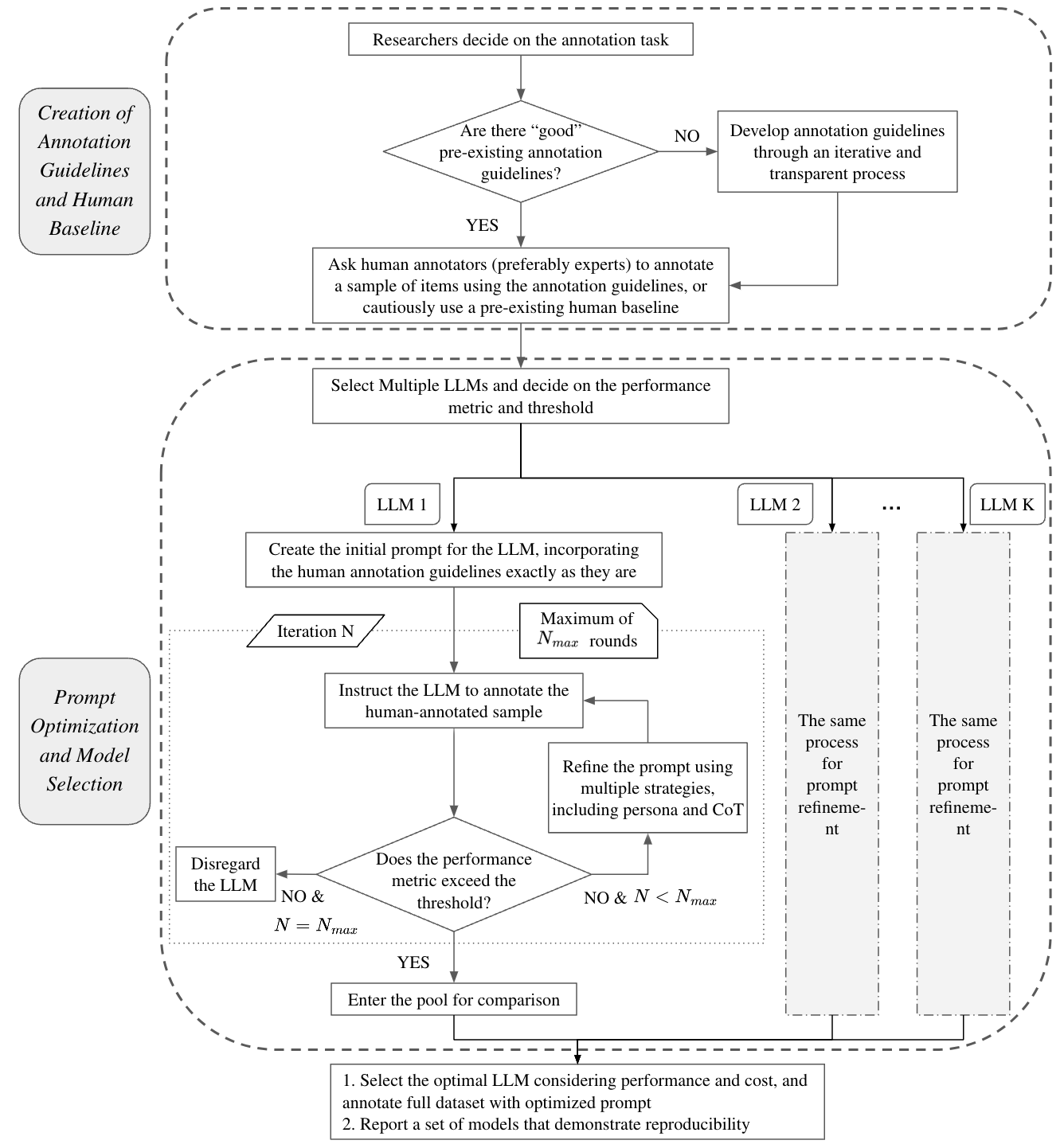}

    \label{fig:workflow}
\end{figure}

Figure~\ref{fig:leap_workflow} shows the detailed workflow for developing the annotation guidelines that we adopted. This workflow was originally proposed by \cite{lee_common_2023} and we have integrated it into the SILICON workflow. In the figure, the term ``researcher'' denotes expert annotators, which may include research assistants or the researchers themselves.

\begin{figure}[htb]
    \centering
        \caption{A Detailed Workflow of Developing Annotation Guidelines from \cite{lee_common_2023}}
    \includegraphics[width=0.85\linewidth]{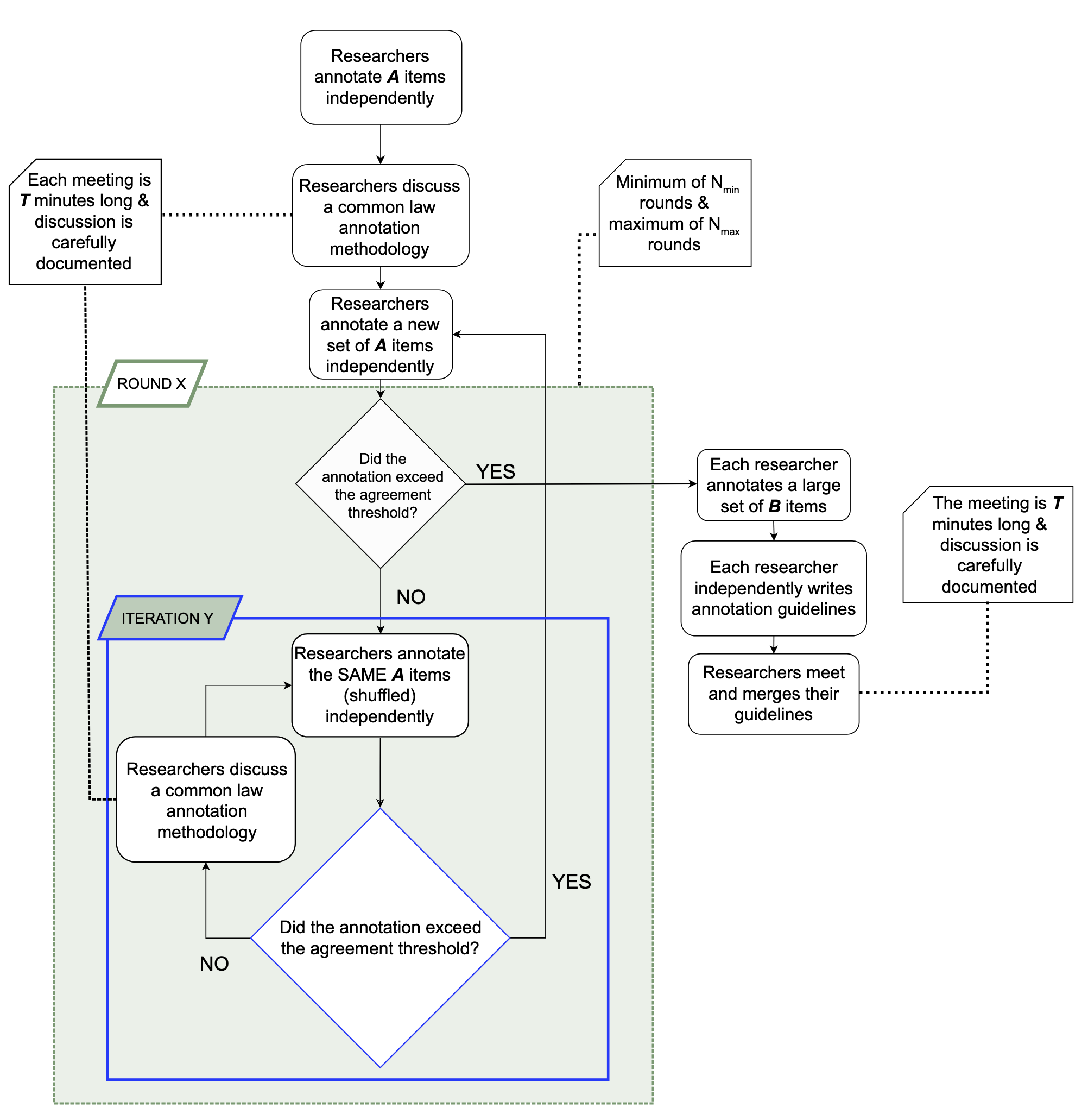}

    \label{fig:leap_workflow}
\end{figure}

We emphasize two practical takeaways from applying this workflow across tasks: (1) pre-existing guidelines often fail to produce reliable human baselines due to inconsistent interpretations, and (2) achieving high IAA typically requires multiple rounds of revision. Figure~\ref{fig:ra_iteration} shows the evolution of agreement levels across multiple rounds. Most tasks start with very poor agreement levels in the first iteration. Over a few iterations, we observe substantial improvements, which highlights the importance of iteratively refining annotation guidelines to resolve ambiguities and ensure consistent interpretation, especially for complex tasks.

\begin{figure} [h]
    \centering
    \caption{Iterative Processes of Annotation Guideline Development}
    \includegraphics[width=0.85\linewidth]{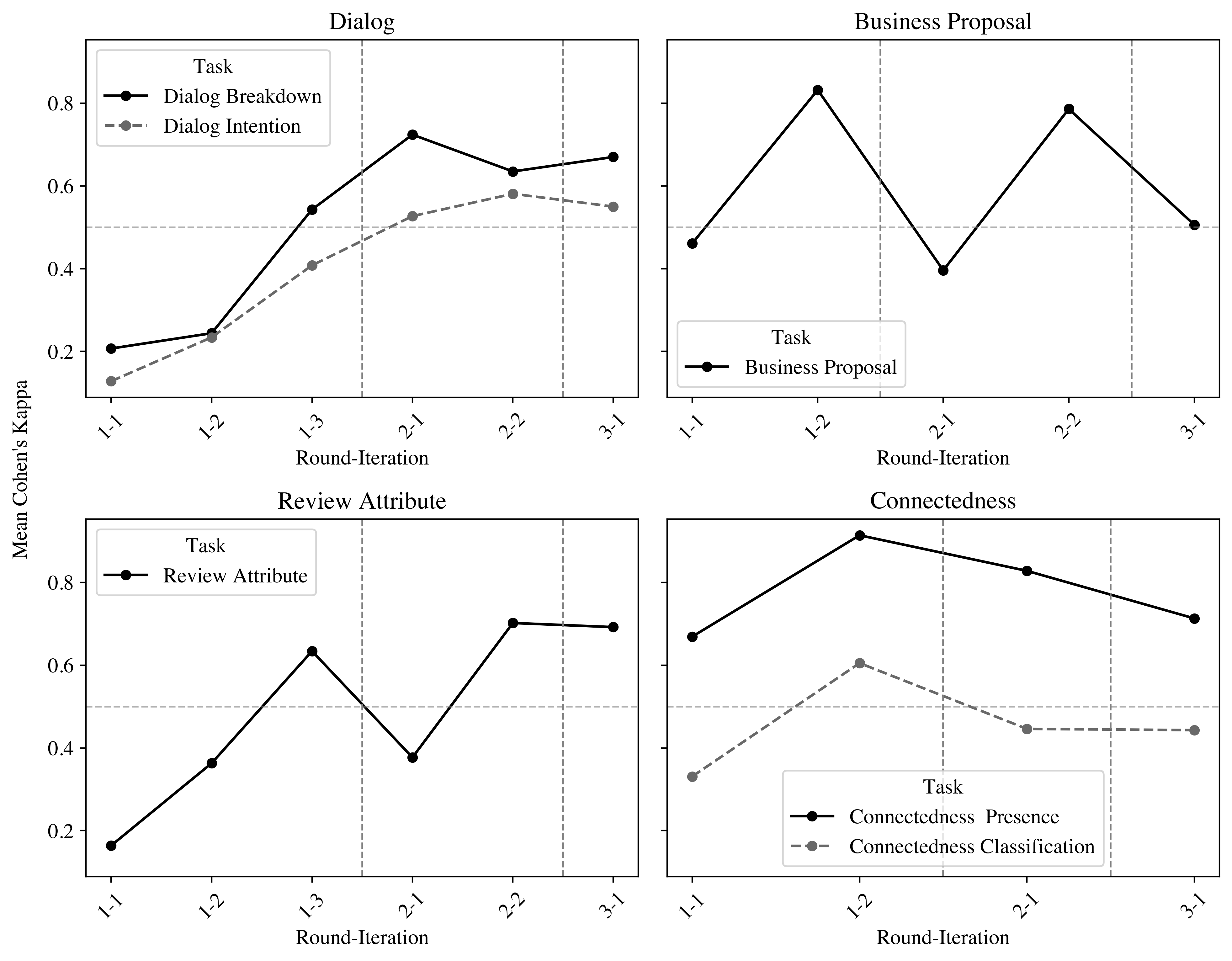}

    \begin{flushleft}
        \footnotesize \parbox{\linewidth}{
     \textit{Notes.} The y-axis shows the IAA among three annotation experts. Achieving a high level of IAA (i.e., $\kappa$)  often requires multiple iterations of refinement of annotation guidelines.
        }
    \end{flushleft}

    \label{fig:ra_iteration}
\end{figure}

To illustrate the process in practice, we provide an example. Specifically, we detail the creation of guidelines for the dialog analysis task, as presented in Table~\ref{tab:leap_dialog}.
We recruit three undergraduate RAs to undertake this iterative process.
Initially, they receive an overview of the task, including the contextual background and labels for the two classification tasks. The RAs independently label a small sample and then discuss their results via Zoom, emphasizing points of disagreement and refining label definitions (Iteration 1).
Following this initial meeting, the RAs annotate the same shuffled sample twice more, meeting after each annotation to discuss their results (Iterations 2 and 3). After reaching an agreement rate threshold in their third iteration, they move to a new sample, repeating the annotation and meeting process twice (Iterations 4 and 5). In their next sample, they achieve the agreement threshold on their first attempt (Iteration 6), concluding the iteration process.
Finally, the RAs meet to consolidate their independent annotation guidelines and annotate a larger sample to establish a human baseline.  Throughout the process, their mean $\kappa$ for the intent classification task improves from 0.128 to 0.61, and their agreement for breakdown analysis increases from 0.207 to 0.665.
The annotation guidelines and human annotation baseline are further used in the next phase of SILICON: optimizing prompts for LLMs and testing multiple LLMs' performance.

{\OneAndAHalfSpacedXI
\begin{table}[H] \footnotesize
    \centering
\caption{Development of Annotation Guidelines for Dialog Analysis Task}
\label{tab:leap_dialog}
    \begin{tabular}{c >{\centering\arraybackslash}p{0.24\linewidth}c ccc}
    \toprule
         Round&  Iteration&Number of& Number of& Intent Analysis&Breakdown Analysis\\
 & & conversations& utterances& Mean $\kappa$& Mean $\kappa$\\
    \midrule
         1&  1&5& 81& 0.128&0.207\\
 1& 2& 5&  81& 0.234&0.244\\
         1&  3&5& 81& 0.408&0.543\\
         2&  1&6& 101& 0.527&0.724\\
 2& 2& 6& 101& 0.581& 0.635\\
 3& 1& 6& 100& 0.550& 0.670\\
 --& Large sample (the human baseline)& 18&  195& 0.61&0.665	\\
    \bottomrule
    \end{tabular}

\end{table}
}

The processes for business proposal evaluation, review attribute detection, and affective content evaluation follow the same approach, wherein we recruit three RAs for each task to go through the iterative process. The details are summarized in Tables \ref{tab:leap_dao} through \ref{tab:leap_connected}.

{\OneAndAHalfSpacedXI
\begin{table}[H] \footnotesize
    \centering
\caption{Development of Annotation Guidelines for Business Proposal Evaluation}
\label{tab:leap_dao}
    \begin{tabular}{c >{\centering\arraybackslash}p{0.35\linewidth}c c} \toprule
         Round&  Iteration&Number of proposals& Mean $\kappa$\\
         \midrule
         1&  1&50& 0.461\\
 1& 2& 50& 0.832\\
         2&  1&49& 0.396\\
         2&  2&49& 0.786\\
 3& 1& 50& 0.506\\
 --& Large sample (the human baseline)& 109& 0.529\\
      \bottomrule
    \end{tabular}

\end{table}
}

{\OneAndAHalfSpacedXI
\begin{table}[H] \footnotesize
    \centering
\caption{Development of Annotation Guidelines for Review Attribute Detection Task}
\label{tab:leap_review}
    \begin{tabular}{c >{\centering\arraybackslash}p{0.35\linewidth}c c}
    \toprule
         Round&  Iteration&Number of Reviews& Mean $\kappa$\\
         \midrule
         1&  1&67& 0.164\\
 1& 2& 67& 0.363\\
         1&  3&67& 0.634\\
         2&  1&65& 0.377\\
 2& 2& 65& 0.702\\
 3& 1& 65& 0.692\\
 --& Large sample (the human baseline)& 180& 0.567\\
      \bottomrule
    \end{tabular}

\end{table}
}
{\OneAndAHalfSpacedXI
\begin{table}[H] \footnotesize
    \centering
\caption{Development of Annotation Guidelines for Affective Content Evaluation Task}
\label{tab:leap_connected}
    \begin{tabular}{c >{\centering\arraybackslash}p{0.24\linewidth}c c c}
    \toprule
         Round&  Iteration&Number of& Connectedness Presence&Connectedness Classification\\
 & & product pages& Mean $\kappa$& Mean $\kappa$\\
         \midrule
         1&  1&50
& 0.618
&--\\
 1& 2& 50
& 0.669
&0.331
\\
 2& 1& 50
& 0.914
&0.605
\\
 2& 2& 50
& 0.828
&0.446
\\
 3& 1& 50
& 0.713
&0.443
\\
 --& Large sample (the human baseline)& 120& 0.822&0.518\\
    \bottomrule
    \end{tabular}

\end{table}
}

\section{Co-labeling Covariance Validation}
\setcounter{table}{0}
\renewcommand{\thetable}{E\arabic{table}}

\setcounter{figure}{0}
\renewcommand{\thefigure}{E\arabic{figure}}

\subsection{Sensitivity of Baseline Choice}

To quantify how baseline choice affects identification, we progressively mix crowd labels into the expert baseline and track \(\lvert \kappa(\text{LLM},\text{expert})-\kappa(\text{LLM},\text{mixed}) \rvert\) as the crowd proportion \(\alpha\) increases; results are reported in Figure~\ref{fig:sensitive_crowd_expert}.  Because crowd workers and LLMs may share systematic error modes on ambiguous items, introducing crowd labels can inflate the co-labeling covariance $J(c)$ identified in Lemma~\ref{lemma:bias-aware}; the rate of drift in $\kappa$ as $\alpha$ grows therefore provides a direct diagnostic for $J$'s empirical bite under each baseline regime.  Sensitivity is strongly task-dependent.  \emph{Business Proposal Evaluation} and \emph{Review Attribute Detection} are largely flat ($\le 0.08$), indicating that $J$ is empirically negligible for these tasks regardless of baseline choice.  \emph{Dialog Intent Classification} exhibits a near-linear, steep rise up to around $0.20$ at $\alpha=1$, and \emph{Connectedness Presence} also shows pronounced growth for $\alpha \ge 0.6$---evidence that crowd workers and LLMs share substantial error modes on these tasks, exactly the pattern that inflates $J$.  Non-monotonicity in \emph{Connectedness Classification} points to heterogeneous, non-systematic crowd deviations.  This task-level heterogeneity validates our use of high-IAA expert baselines to minimize $J$ across the board, while also flagging Dialog Intent Classification as the task where residual $J$ under expert baselines may be largest.

 \begin{figure} [h]
    \centering
    \caption{Sensitivity of LLM–Expert Agreement to Crowd–Expert Label Mixing}
    \includegraphics[width=1\linewidth]{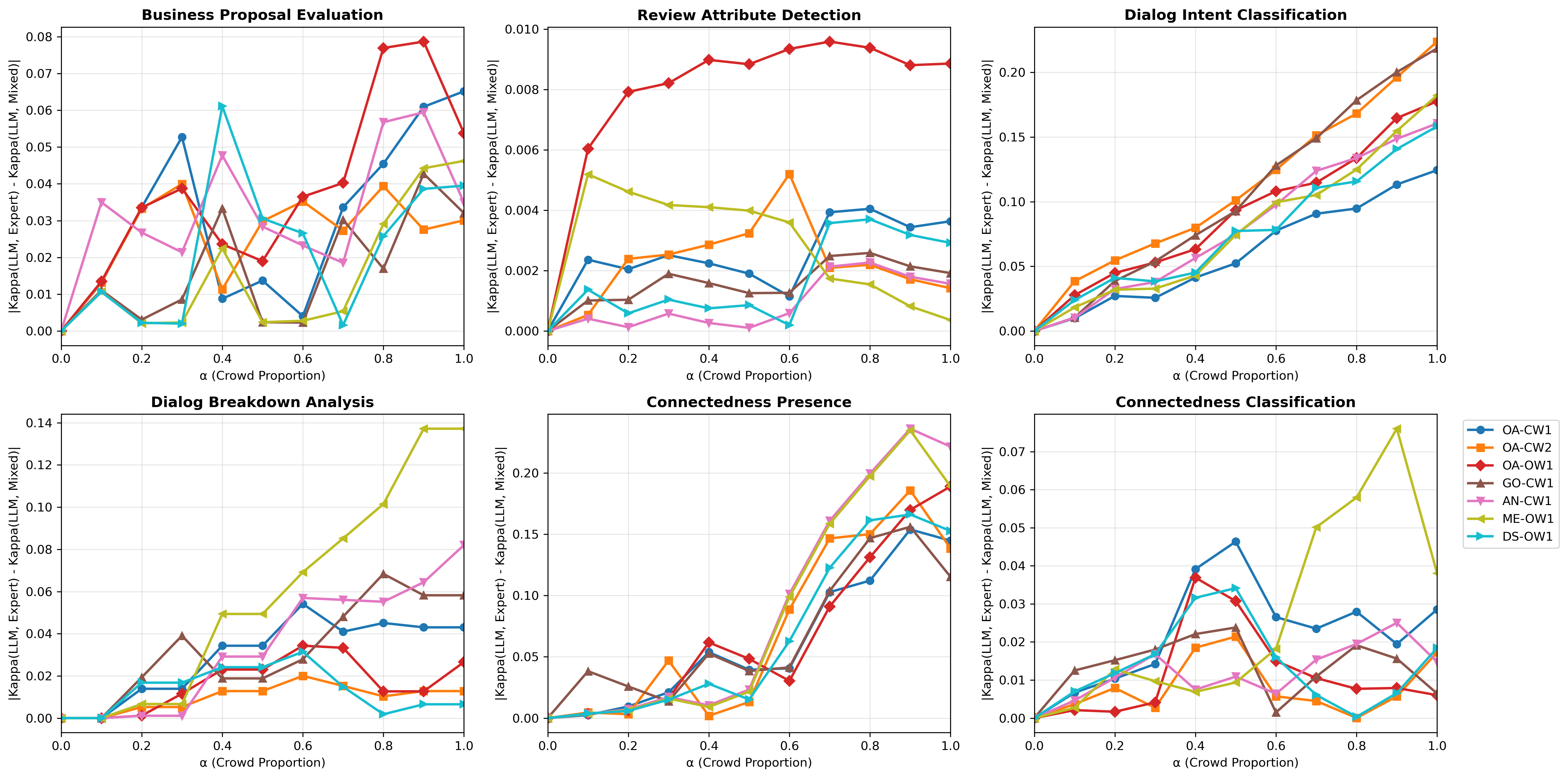}

    \begin{flushleft}
        \footnotesize \parbox{\linewidth}{
     \textit{Notes.} For each task and model, the y-axis reports the absolute change in Cohen's $\kappa$ when moving from expert labels to a mixed label set, $|\,\kappa(\text{LLM}, \text{expert}) - \kappa(\text{LLM}, (1-\alpha)\cdot \text{expert} + \alpha\cdot \text{crowd})\,|$, as the crowd proportion $\alpha$ increases. The slope and shape of each curve, i.e., how quickly the gap grows, reveal robustness to crowd labels (where flatter curves indicate robustness and steep curves indicate $\kappa$ is sensitive to expert-only annotations).
        }
    \end{flushleft}
    \label{fig:sensitive_crowd_expert}
    \vspace{-.75em}
\end{figure}

\subsection{Direct Estimation of Co-labeling Covariance}\label{app:j_estimation}

Lemma~\ref{lemma:bias-aware} decomposes reference agreement as $R(c) = a\,T(c) + b + J(c)$, where $J(c)$ captures shared error modes between the LLM and human reference.  Because $T(c)$ and $J(c)$ are not separately observed from $R(c)$ alone, one might worry that configurations selected by SILICON owe their high $R$ to inflated $J$ rather than genuine accuracy $T$.  We bound $J$ empirically by exploiting items on which all three expert annotators unanimously agree.  On these items the majority-vote reference is a near-perfect proxy for ground truth ($e\approx 0$), so the LLM's match rate against the unanimous label directly estimates true accuracy $\hat T(c)$.  We then back out
\[
  J(c) \;=\; R(c) \;-\; a\,\hat T(c) \;-\; b,
\]
where $R(c)$ is the full-sample agreement and $a,b$ are the task-level constants from Lemma~\ref{lemma:bias-aware}, with $e$ estimated as the average individual-annotator disagreement rate against the majority vote.

Table~\ref{tab:j_estimation} reports $\hat T$, $R$, and the resulting $J$ for each task--model pair.  For the best-performing model on each task, $|J|$ is small: at most 0.01 on four of six tasks (Business Proposal, Dialog Breakdown, Connectedness Presence, Dialog Intent) and modestly larger on the two multi-label tasks where the unanimous subset is smaller.
Because $\hat T$ is estimated on the unanimous (and typically easier) subset, it may slightly overestimate full-sample true accuracy, biasing $J$ downward.  The reported $|J|$ values are therefore conservative upper bounds.
Together with the crowd-mixing sensitivity analysis in Figure~\ref{fig:sensitive_crowd_expert}, which shows that $J$'s empirical bite is negligible for tasks with flat drift curves and is mitigated by expert baselines for tasks with steep curves, these results support the conclusion that SILICON's configuration rankings reflect genuine accuracy gains rather than co-labeling artifacts.

{\OneAndAHalfSpacedXI
\begin{table}[h!]\footnotesize
\centering
\caption{Estimated Co-labeling Covariance $J(c)$ on Unanimous-Agreement Subsets}
\label{tab:j_estimation}
\begin{threeparttable}
\begin{tabular}{@{}llccccccc@{}}
\toprule
Task & Model & $N$ & $N_{\text{unan}}$ (\%) & $\hat{T}$ & $R$ & $e$ & $K$ & $J$ \\
\midrule
Business Proposal Evaluation & OA-CW1 & 109 & 50 (45.9\%) & 0.860 & 0.697 & 0.180 & 6 & $-$0.013 \\
                   & OA-CW2 &     &             & 0.860 & 0.661 &       &   & $-$0.049 \\
                   & OA-OW1 &     &             & 0.760 & 0.642 &       &   & \phantom{$-$}0.011 \\
                   & GO-CW1 &     &             & 0.840 & 0.661 &       &   & $-$0.034 \\
                   & AN-CW1 &     &             & 0.880 & 0.679 &       &   & $-$0.047 \\
                   & ME-OW1 &     &             & 0.820 & 0.670 &       &   & $-$0.009 \\
                   & DS-OW1 &     &             & 0.820 & 0.639 &       &   & $-$0.040 \\
\midrule
Review Attribute Detection  & OA-CW1 & 180 & 64 (35.6\%) & 0.516 & 0.383 & 0.259 & 34 & $-$0.003 \\
                   & OA-CW2 &     &             & 0.672 & 0.494 &       &    & $-$0.006 \\
                   & OA-OW1 &     &             & 0.594 & 0.439 &       &    & $-$0.004 \\
                   & GO-CW1 &     &             & 0.266 & 0.217 &       &    & \phantom{$-$}0.014 \\
                   & AN-CW1 &     &             & 0.328 & 0.322 &       &    & \phantom{$-$}0.074 \\
                   & ME-OW1 &     &             & 0.094 & 0.094 &       &    & \phantom{$-$}0.018 \\
                   & DS-OW1 &     &             & 0.391 & 0.378 &       &    & \phantom{$-$}0.084 \\
\midrule
Dialog Intent Classification     & OA-CW1 & 195 & 84 (43.1\%) & 0.500 & 0.314 & 0.234 & 18 & $-$0.076 \\
                   & OA-CW2 &     &             & 0.226 & 0.100 &       &    & $-$0.085 \\
                   & OA-OW1 &     &             & 0.250 & 0.131 &       &    & $-$0.071 \\
                   & GO-CW1 &     &             & 0.286 & 0.178 &       &    & $-$0.051 \\
                   & AN-CW1 &     &             & 0.679 & 0.450 &       &    & $-$0.074 \\
                   & ME-OW1 &     &             & 0.250 & 0.147 &       &    & $-$0.055 \\
                   & DS-OW1 &     &             & 0.381 & 0.199 &       &    & $-$0.101 \\
\midrule
Dialog Breakdown Analysis   & OA-CW1 & 195 & 137 (70.3\%) & 0.816 & 0.753 & 0.099 & 3 & \phantom{$-$}0.008 \\
                   & OA-CW2 &     &              & 0.540 & 0.559 &       &   & \phantom{$-$}0.050 \\
                   & OA-OW1 &     &              & 0.893 & 0.821 &       &   & \phantom{$-$}0.011 \\
                   & GO-CW1 &     &              & 0.992 & 0.896 &       &   & \phantom{$-$}0.002 \\
                   & AN-CW1 &     &              & 0.905 & 0.821 &       &   & \phantom{$-$}0.000 \\
                   & ME-OW1 &     &              & 0.661 & 0.636 &       &   & \phantom{$-$}0.024 \\
                   & DS-OW1 &     &              & 0.941 & 0.856 &       &   & \phantom{$-$}0.005 \\
\midrule
Connected Presence  & OA-CW1 & 120 & 107 (89.2\%) & 0.878 & 0.833 & 0.047 & 2 & $-$0.009 \\
                   & OA-CW2 &     &              & 0.841 & 0.808 &       &   & $-$0.001 \\
                   & OA-OW1 &     &              & 0.878 & 0.825 &       &   & $-$0.018 \\
                   & GO-CW1 &     &              & 0.860 & 0.800 &       &   & $-$0.026 \\
                   & AN-CW1 &     &              & 0.916 & 0.867 &       &   & $-$0.010 \\
                   & ME-OW1 &     &              & 0.878 & 0.825 &       &   & $-$0.018 \\
                   & DS-OW1 &     &              & 0.878 & 0.842 &       &   & $-$0.001 \\
\midrule
Connected Classification& OA-CW1 & 120 & 43 (35.8\%)  & 0.721 & 0.350 & 0.344 & 24 & $-$0.127 \\
                   & OA-CW2 &     &              & 0.605 & 0.308 &       &    & $-$0.094 \\
                   & OA-OW1 &     &              & 0.605 & 0.292 &       &    & $-$0.111 \\
                   & GO-CW1 &     &              & 0.698 & 0.333 &       &    & $-$0.129 \\
                   & AN-CW1 &     &              & 0.721 & 0.358 &       &    & $-$0.118 \\
                   & ME-OW1 &     &              & 0.698 & 0.325 &       &    & $-$0.137 \\
                   & DS-OW1 &     &              & 0.605 & 0.308 &       &    & $-$0.094 \\
\bottomrule
\end{tabular}
\begin{tablenotes}
    \footnotesize
    \setlength{\baselineskip}{\normalbaselineskip}
    \item \textit{Notes.} $N_{\text{unan}}$ is the number of items on which all three expert annotators unanimously agree.  $\hat{T}$ is the LLM's match rate against the unanimous label (proxy for true accuracy).  $R$ is the full-sample agreement with the expert majority vote.  $e$ is the average annotator error rate against the majority vote; $K$ is the number of distinct label classes.  $J = R - a\hat{T} - b$, where $a = (1-e) - e/(K-1)$ and $b = e/(K-1)$.  Because $\hat{T}$ is estimated on the (typically easier) unanimous subset, it may overestimate full-sample $T$, making the reported $|J|$ values conservative upper bounds.
\end{tablenotes}
\end{threeparttable}
\end{table}
}

\clearpage
\section{Additional Model Results} \label{app:addition_result}
\setcounter{table}{0}
\renewcommand{\thetable}{F\arabic{table}}

\setcounter{figure}{0}
\renewcommand{\thefigure}{F\arabic{figure}}

In addition to the seven models consistently evaluated in the main analysis, we also assess five additional models: four more advanced ones (AN-CW2, GO-CW2, OA-CW3, and OA-CW4) and one less advanced model (ME-OW2). Their performance is reported in Table~\ref{tab:sum_model_add}.
Overall, we observe only modest performance improvements within model families across versions. In terms of the best-performing model for each task, we find that the additional models emerge as top performers in several cases: AN-CW2 performs best on dialog breakdown analysis and connectedness classification; OA-CW4 performs best in detecting language toxicity; and GO-CW2 achieves the highest agreement level in criticism stance detection.
The improvement in criticism stance detection by GO-CW2 is substantial, with $\kappa$ value increasing from around 0.5 (all other models) to 0.76. We suspect this gain may be attributed to memorization rather than genuine advances in inference capability, as the underlying dataset \citep{peng_dynamics_2022} is publicly available and may have been included in the model's training data.
{\OneAndAHalfSpacedXI
\begin{table}[h!] \footnotesize
\centering
\caption{Cross-Model Performance Comparison (with Additional Models)}
\label{tab:sum_model_add}
\begin{threeparttable}
\begin{tabular}{@{\extracolsep{\fill}}>{\centering\arraybackslash}p{0.12\textwidth}>{\centering\arraybackslash}p{0.05\textwidth}>{\centering\arraybackslash}p{0.05\textwidth}>{\centering\arraybackslash}p{0.05\linewidth}>{\centering\arraybackslash}p{0.06\textwidth}>{\centering\arraybackslash}p{0.06\textwidth}>{\centering\arraybackslash}p{0.06\textwidth}>{\centering\arraybackslash}p{0.06\textwidth}>{\centering\arraybackslash}p{0.06\textwidth}>{\centering\arraybackslash}p{0.06\textwidth}>{\centering\arraybackslash}p{0.06\textwidth}>{\centering\arraybackslash}p{0.085\textwidth}@{}>{\centering\arraybackslash}p{0.06\linewidth}}
\toprule
    Task& OA-CW1&OA-CW2&OA-CW3&OA-CW4&OA-OW1&GO-CW1&GO-CW2&AN-CW1&AN-CW2&ME-OW1&ME-OW2&DS-OW1\\
\midrule
 Business Proposal Evaluation&\textbf{0.605}
 &0.560 &0.550
&0.592
 &0.533 &0.561& 0.562&0.578
 &0.566& 0.566
&0.485
 &0.517\\
 Review Attribute Detection& 0.479
 &\textbf{0.554} &0.422
& 0.466
 &0.501 &0.346& 0.382&0.429
 &0.388& 0.220
&0.152
 &0.458\\
 Dialog Intent Classification& 0.335
 &0.191 &0.372
& 0.399
 &0.187 &0.238& 0.375&\textbf{0.427}
 &0.364& 0.205
&0.193
 &0.251\\
  Dialog Breakdown Analysis&0.566
 &0.223 &0.717
&0.717
 &0.617 &0.727&0.685&0.689
 &\textbf{0.796}&0.245
&0.252
 &0.744\\
 Connectedness Presence& 0.632
 &0.555 &0.678
& 0.625
 &0.602 &0.566& 0.566& \textbf{0.688}
 &0.681& 0.602
&0.232
 &0.633\\
 Connectedness Classification& 0.366 &0.303 &0.354
& 0.355 &0.286 &0.324& 0.303& 0.371 &\textbf{0.372}& 0.338&0.177 &0.302\\
Language Toxicity Detection&0.471&0.307 &0.497
&\textbf{0.516} &0.389 &0.407&0.463&0.390  &0.436& 0.366&0.479 &0.340\\
Criticism Stance Detection& 0.540 &0.510 &0.540
& 0.470 &0.520 &0.550& \textbf{0.760}&0.570 &0.540& 0.500&0.500 &0.580\\
 Sentiment Analysis&0.918 &0.901 &0.905&0.918 &0.900 &0.885&0.902&0.822  &0.847& 0.900&0.901 &\textbf{0.935}\\
\bottomrule
\end{tabular}
\begin{tablenotes}
    \footnotesize
    \setlength{\baselineskip}{\normalbaselineskip}
    \item \textit{Notes.} The values represent Cohen's Kappa scores between LLM annotations and human reference labels. Codes follow provider-openness-index naming, where CW denotes closed-weight and OW denotes open-weight.
\end{tablenotes}
\end{threeparttable}
\end{table}
}

\section{Simulation Details}
\label{appendix:simulation}
\setcounter{table}{0}
\renewcommand{\thetable}{G\arabic{table}}

\setcounter{figure}{0}
\renewcommand{\thefigure}{G\arabic{figure}}

\subsection{Simulation Procedure}\label{app:simulation_procedure}

This appendix provides complete procedural details for the Monte Carlo simulation study, including the mapping between conceptual constructs in the SILICON framework and simulation parameters, the data-generating process, and the formal algorithm for each replicate.

The simulation is designed to mirror the workflow a researcher would realistically follow when deploying LLM annotation for downstream inference. In practice, researchers first establish a human reference baseline by having annotators label a validation sample according to annotation guidelines. They then evaluate multiple candidate LLM configurations---corresponding to different prompt formulations, model choices, or parameter settings---by measuring agreement with this human reference. Because ground truth is unobservable, researchers must select among candidates based on reference agreement $R$ rather than true agreement $T$. Finally, they deploy the selected configuration to annotate data for use in statistical analysis. The simulation formalizes each step of this pipeline to quantify how measurement error propagates through to downstream estimators.

The conceptual configuration $c = (G, H, P, M)$ from the main text maps to simulation parameters as follows. Guidelines $G$ and human baseline procedures $H$ jointly determine human reference quality, which we operationalize through the human error rate $e_h \in [0, 1)$ (the error rate $e$ from Assumption~\ref{assum:sln}): for each item with true label $y$, the human reference equals $r = y$ with probability $1 - e_h$, and with probability $e_h$ it takes a uniformly random value from the remaining $K - 1$ classes. Lower $e_h$ corresponds to higher-quality guidelines and expert annotators who make fewer mistakes. Prompt $P$ and model $M$ jointly determine LLM annotation quality, which we operationalize through two parameters: the model error rate $e$ of the confusion matrix, specifying $\Pr(\hat{y} \neq k \mid y = k) = e$ for each class (with errors distributed uniformly across the remaining $K-1$ classes), and the coupling parameter $\rho \in [0, 1]$, specifying the probability that the LLM copies the human reference label rather than drawing independently from the confusion matrix. Lower $e$ corresponds to better prompt/model configurations that more accurately recover ground truth; higher $\rho$ corresponds to greater co-labeling covariance $J(c)$, which arises when the LLM and human share error modes due to ambiguous guidelines or similar decision boundaries. The simulation does not distinguish between prompt-induced and model-induced effects because both operate through the same mechanism: they determine the joint distribution of LLM predictions given truth and human reference.

The data-generating process operates on a binary classification setting with $K = 2$ classes and balanced class probabilities $\pi = (0.5, 0.5)$. Ground truth labels are drawn as $y_i \sim \text{Categorical}(\pi)$. The outcome variable for downstream inference is generated as $z_i = \beta y_i + u_i$ with true effect $\beta = 1.0$ and noise $u_i \overset{\text{iid}}{\sim} N(0, \sigma^2)$ where $\sigma = 1.0$. Human reference labels are generated under Assumption~\ref{assum:sln}:
\[
\Pr(r_i = \ell \mid y_i = k) =
\begin{cases}
1 - e_h & \text{if } \ell = k, \\
e_h/(K - 1) & \text{if } \ell \neq k.
\end{cases}
\]
Each candidate LLM configuration $m$ is characterized by a symmetric confusion matrix with error rate $e^{(m)}$ and a coupling parameter $\rho^{(m)}$. The confusion matrix is:
\[
C^{(m)}_{k\ell} =
\begin{cases}
1 - e^{(m)} & \text{if } \ell = k, \\
e^{(m)}/(K - 1) & \text{if } \ell \neq k.
\end{cases}
\]
LLM predictions are generated in two stages. First, an independent baseline prediction is drawn: $\Pr(\tilde{y}_i = \ell \mid y_i = k) = C^{(m)}_{k\ell}$. Second, coupling is introduced: with probability $\rho^{(m)}$, the LLM copies the human reference $r_i$; otherwise it retains the independent prediction $\tilde{y}_i$. This mechanism generates positive co-labeling covariance $J(c)$ in Lemma~\ref{lemma:bias-aware}.

A candidate pool consists of $M = 10$ configurations. Error rates $\{e^{(m)}\}_{m=1}^M$ are evenly spaced across a specified range, and coupling parameters $\{\rho^{(m)}\}_{m=1}^M$ are evenly spaced across another range but then randomly permuted so that $e$ and $\rho$ are statistically independent within the pool. This ensures that selection based on $R$ cannot trivially rank configurations by error rate alone; instead, the selection noise induced by coupling creates realistic uncertainty about which configuration truly maximizes $T$.

A key design choice is that the baseline and improved accuracy distributions share the same width and overlap. Two factors contribute to a configuration's confusion matrix (accuracy): (1) inherent model/prompt capability and (2) the effect of annotation guidelines (refined vs.\ one-shot). Because both regimes evaluate the same set of models, factor (1)---which drives the width of the accuracy distribution---is identical across regimes. Only factor (2) differs, shifting the mean accuracy upward for the improved regime. We therefore parameterize the baseline accuracy range as $[0.65, 0.85]$ and the improved range as $[0.75, 0.95]$, both with width $0.20$ and overlapping in $[0.75, 0.85]$. This overlap means that the best baseline configurations are comparable to the worst improved ones, preventing the comparison from being mechanically driven by non-overlapping accuracy pools. For the same reason, the coupling parameter $\rho \in [0.20, 0.50]$ is identical across regimes: $\rho$ reflects model-level tendencies to mimic reference labels, which depends on the model architecture and prompt structure rather than guidelines quality. The only parameters that differ across regimes are the human error rate $e_h$ (capturing reference quality) and the center of the accuracy distribution (capturing the guidelines-induced performance shift).

Algorithm~\ref{alg:simulation} presents the complete procedure for one Monte Carlo replicate.
The complete simulation runs $1{,}000$ replicates for each regime and aggregates results to compute mean $\hat{\beta}$, bias $(\hat{\beta} - \beta)$, 95\% CI coverage, mean test-set $T$, and selection accuracy (fraction of replicates where $m^* = m^\dagger$).
Table~\ref{tab:simulation_params} summarizes all parameter values.

\begin{algorithm}[t]
\caption{Selection-and-Inference Replicate}
\label{alg:simulation}
\DontPrintSemicolon

\KwIn{Human error rate $e_h$, candidate pool $\{(e^{(m)}, \rho^{(m)})\}_{m=1}^M$, validation size $N_{\text{val}}$, test size $N_{\text{test}}$, true effect $\beta$, noise $\sigma$}
\KwOut{Coefficient estimate $\hat{\beta}$, test-set $T$, selection correctness indicator}

\textit{Step 1: Generate validation data}\;
Draw $y_i^{(\text{val})} \sim \text{Categorical}(\pi)$ for $i = 1,\ldots,N_{\text{val}}$\;
Generate human reference $r_i^{(\text{val})}$ with error rate $e_h$\;

\textit{Step 2: Score each candidate on validation set}\;
\For{$m \gets 1$ \KwTo $M$}{
  \For{$i \gets 1$ \KwTo $N_{\text{val}}$}{
    Draw $\tilde{y}_i \sim C^{(m)}_{y_i^{(\text{val})}, \cdot}$\;
    Set $\hat{y}^{(m)}_i \gets r_i^{(\text{val})}$ with prob $\rho^{(m)}$, else $\hat{y}^{(m)}_i \gets \tilde{y}_i$\;
  }
  $R^{(m)} \gets N_{\text{val}}^{-1} \sum_{i=1}^{N_{\text{val}}} \mathbf{1}[\hat{y}^{(m)}_i = r_i^{(\text{val})}]$\;
  $T^{(m)} \gets N_{\text{val}}^{-1} \sum_{i=1}^{N_{\text{val}}} \mathbf{1}[\hat{y}^{(m)}_i = y_i^{(\text{val})}]$\;
}

\textit{Step 3: Select best configuration by observable $R$}\;
$m^* \gets \arg\max_m R^{(m)}$ \tcp*[r]{Researcher's choice (observable)}\;
$m^\dagger \gets \arg\max_m T^{(m)}$ \tcp*[r]{Oracle best (unobservable)}\;
Record selection correctness: $\mathbf{1}[m^* = m^\dagger]$\;

\textit{Step 4: Generate test data}\;
Draw $y_i^{(\text{test})} \sim \text{Categorical}(\pi)$ for $i = 1,\ldots,N_{\text{test}}$\;
$u_i \sim N(0,\sigma^2)$ and $z_i \gets \beta \cdot y_i^{(\text{test})} + u_i$\;
Generate human reference $r_i^{(\text{test})}$ with error rate $e_h$\;

\textit{Step 5: Deploy selected configuration on test set}\;
Generate $\hat{y}^{(m^*)}_i$ for $i = 1,\ldots,N_{\text{test}}$ using confusion matrix $C^{(m^*)}$ and coupling $\rho^{(m^*)}$\;
$T^{(m^*)} \gets N_{\text{test}}^{-1} \sum_{i=1}^{N_{\text{test}}} \mathbf{1}[\hat{y}^{(m^*)}_i = y_i^{(\text{test})}]$\;

\textit{Step 6: Run downstream inference}\;
Fit OLS: $z = \gamma_0 + \gamma_1 \hat{y}^{(m^*)} + \epsilon$\;
Record $\hat{\beta} \gets \hat{\gamma}_1$, standard error, 95\% CI, and whether CI covers $\beta$\;

\end{algorithm}

\begin{table}[htbp]
\centering
\caption{Simulation Parameter Values}
\label{tab:simulation_params}
\begin{tabular}{lll}
\toprule
Parameter & Symbol & Value \\
\midrule
Number of classes & $K$ & 2 \\
Class probabilities & $\pi$ & $(0.5, 0.5)$ \\
Validation set size & $N_{\text{val}}$ & 1,000 \\
Test set size & $N_{\text{test}}$ & 300 \\
Number of candidates & $M$ & 10 \\
Monte Carlo replicates & $N_{\text{rep}}$ & 1,000 \\
True effect size & $\beta$ & 1.0 \\
Outcome noise & $\sigma$ & 1.0 \\
\midrule
\multicolumn{3}{l}{\textit{Baseline Regime}} \\
Human error rate & $e_h$ & 0.15 \\
Candidate accuracy range & $[1-e_{\max}, 1-e_{\min}]$ & $[0.65, 0.85]$ \\
Candidate coupling range & $[\rho_{\min}, \rho_{\max}]$ & $[0.20, 0.50]$ \\
\midrule
\multicolumn{3}{l}{\textit{Improved (SILICON) Regime}} \\
Human error rate & $e_h$ & 0.02 \\
Candidate accuracy range & $[1-e_{\max}, 1-e_{\min}]$ & $[0.75, 0.95]$ \\
Candidate coupling range & $[\rho_{\min}, \rho_{\max}]$ & $[0.20, 0.50]$ \\
\bottomrule
\end{tabular}

\vspace{0.5em}
\footnotesize \textit{Notes.} Both regimes share the same coupling range $\rho \in [0.20, 0.50]$ and candidate accuracy distributions of equal width (0.20), reflecting the assumption that both regimes evaluate the same underlying models. The accuracy distributions overlap in $[0.75, 0.85]$: the best baseline configurations may match the worst improved configurations. The only regime-specific parameters are $e_h$ (reference quality) and the mean of the accuracy distribution (guidelines-induced shift). The human error rates $e_h = 0.15$ (baseline) and $e_h = 0.02$ (improved) are calibrated from the empirical annotation results.
\end{table}

\subsection{Sensitivity Analysis}
\label{appendix:simulation_sensitivity}

The main simulation uses overlapping accuracy distributions with identical coupling ($\rho$) across regimes. We conduct three sensitivity analyses to verify that the results are not driven by specific parameter choices.

\paragraph{Large accuracy overlap (Figure~\ref{fig:sensitivity_s3}).}
We narrow the gap between accuracy distributions by shifting the improved range from $[0.75, 0.95]$ to $[0.70, 0.90]$, so that the overlap region $[0.70, 0.85]$ covers most of both distributions. The coupling parameter remains identical at $\rho \in [0.20, 0.50]$. Even under this near-complete overlap---where the improved pool is only marginally better on average---SILICON still reduces bias by 57\% (from $-0.304$ to $-0.132$) and more than doubles CI coverage (from 31.7\% to 80.5\%). Selection accuracy remains high at 96.7\%. This confirms that the improvement is not driven by a large gap between accuracy pools; rather, the cleaner reference ($e_h = 0.02$) reliably identifies the best candidate regardless of how closely the pools overlap.

\begin{figure}[htbp]
    \centering
    \caption{Sensitivity: Large Accuracy Overlap, Same Coupling}
    \includegraphics[width=\textwidth]{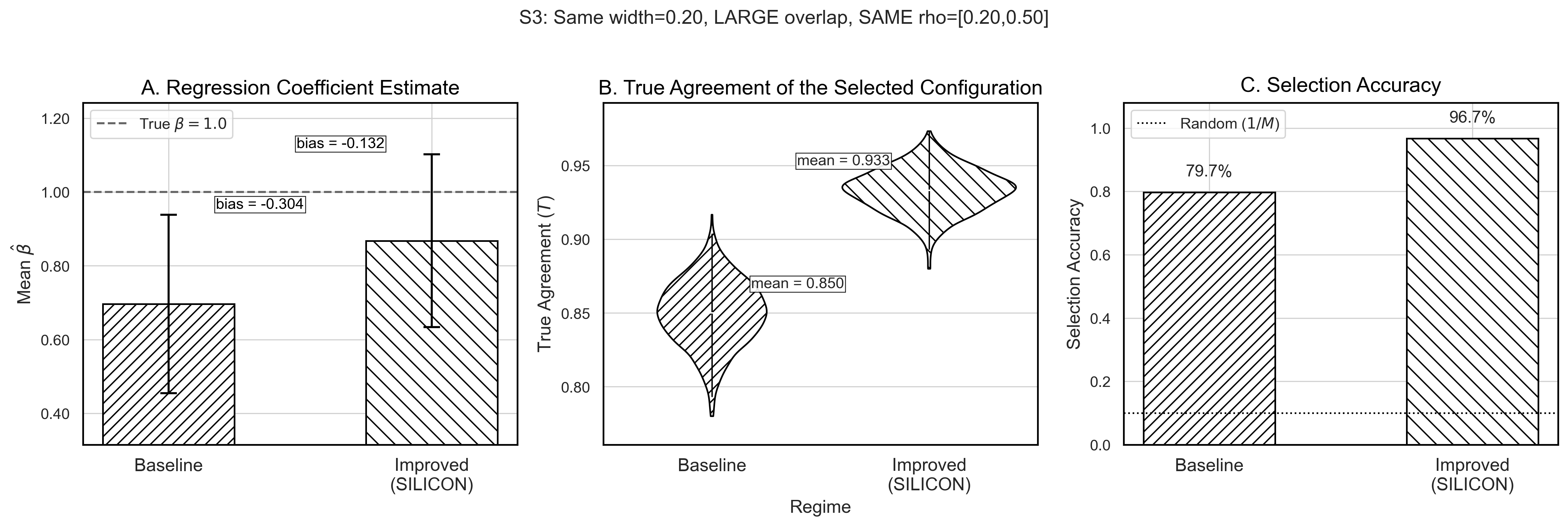}
    \begin{flushleft}
        \footnotesize \parbox{1.0\linewidth}{
        \textit{Notes.} Same format as Figure~\ref{fig:simulation_fig1}. Both regimes use $\rho \in [0.20, 0.50]$. Baseline: $e_h = 0.15$, accuracy $\in [0.65, 0.85]$. Improved: $e_h = 0.02$, accuracy $\in [0.70, 0.90]$. Overlap covers $[0.70, 0.85]$.
        }
    \end{flushleft}
    \label{fig:sensitivity_s3}
\end{figure}

\paragraph{Asymmetric coupling (Figure~\ref{fig:sensitivity_s2}).}
We test whether the results change if SILICON has lower coupling than the baseline, which might occur if refined guidelines reduce the tendency of LLMs to mimic reference labels. Using the moderate-overlap accuracy distributions ($[0.65, 0.85]$ vs.\ $[0.75, 0.95]$), we set $\rho \in [0.20, 0.50]$ for the baseline but $\rho \in [0.02, 0.05]$ for SILICON. Counterintuitively, giving SILICON lower coupling yields slightly \emph{worse} performance (bias $= -0.100$ vs.\ $-0.075$ in the main specification). This occurs because high coupling with an accurate reference ($e_h = 0.02$) effectively boosts the LLM's true agreement---the model ``copies'' the correct reference label more often. With low coupling, the LLM relies entirely on its intrinsic accuracy. Thus, the main simulation's assumption of equal $\rho$ is, if anything, conservative: it does not artificially favor SILICON through asymmetric coupling.

\begin{figure}[htbp]
    \centering
    \caption{Sensitivity: Moderate Overlap, SILICON Lower Coupling}
    \includegraphics[width=\textwidth]{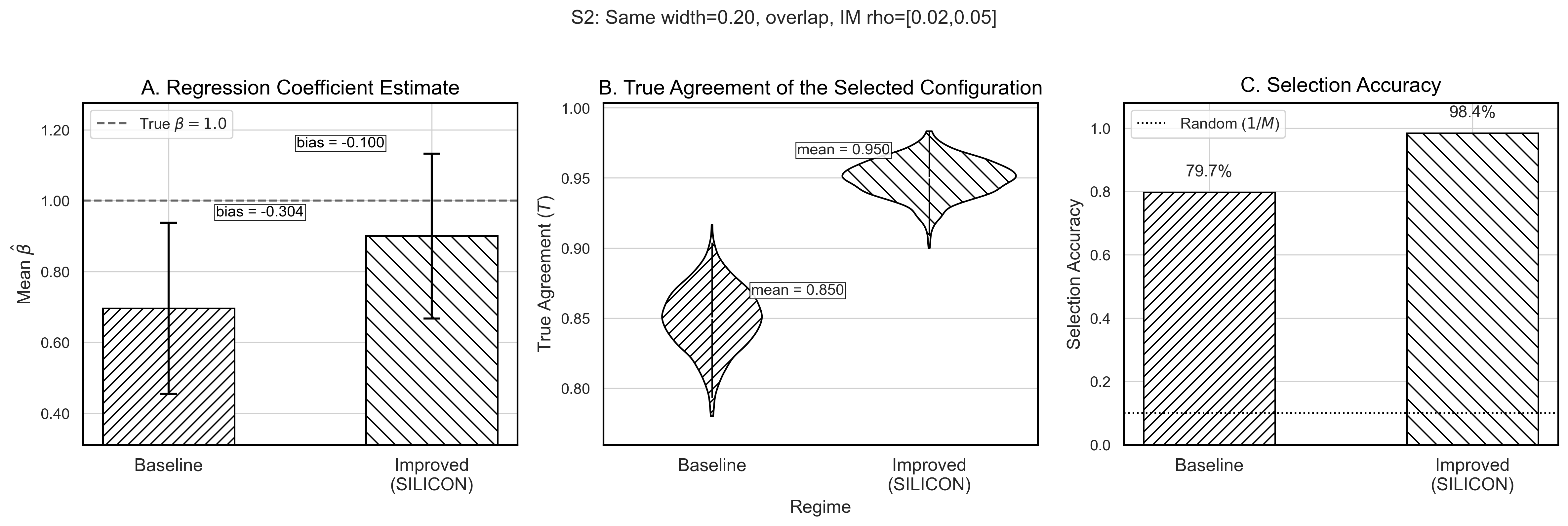}
    \begin{flushleft}
        \footnotesize \parbox{1.0\linewidth}{
        \textit{Notes.} Same format as Figure~\ref{fig:simulation_fig1}. Baseline: $e_h = 0.15$, accuracy $\in [0.65, 0.85]$, $\rho \in [0.20, 0.50]$. Improved: $e_h = 0.02$, accuracy $\in [0.75, 0.95]$, $\rho \in [0.02, 0.05]$.
        }
    \end{flushleft}
    \label{fig:sensitivity_s2}
\end{figure}

\paragraph{Very low coupling for both regimes (Figure~\ref{fig:sensitivity_r3}).}
We set $\rho \in [0.02, 0.05]$ for both regimes, simulating a scenario in which LLMs produce predictions nearly independently of the human reference. Under this minimal-coupling scenario, SILICON still reduces bias by 68\% (from $-0.310$ to $-0.100$) and nearly triples CI coverage (from 30.0\% to 85.4\%). Selection accuracy for SILICON remains near-perfect at 98.4\%. These results confirm that the improvement does not depend on coupling: even when the LLM and human reference are nearly independent, the lower $e_h$ in the improved regime makes $R$ a reliable proxy for $T$, enabling accurate selection and reduced downstream bias.

\begin{figure}[htbp]
    \centering
    \caption{Sensitivity: Very Low Coupling for Both Regimes}
    \includegraphics[width=\textwidth]{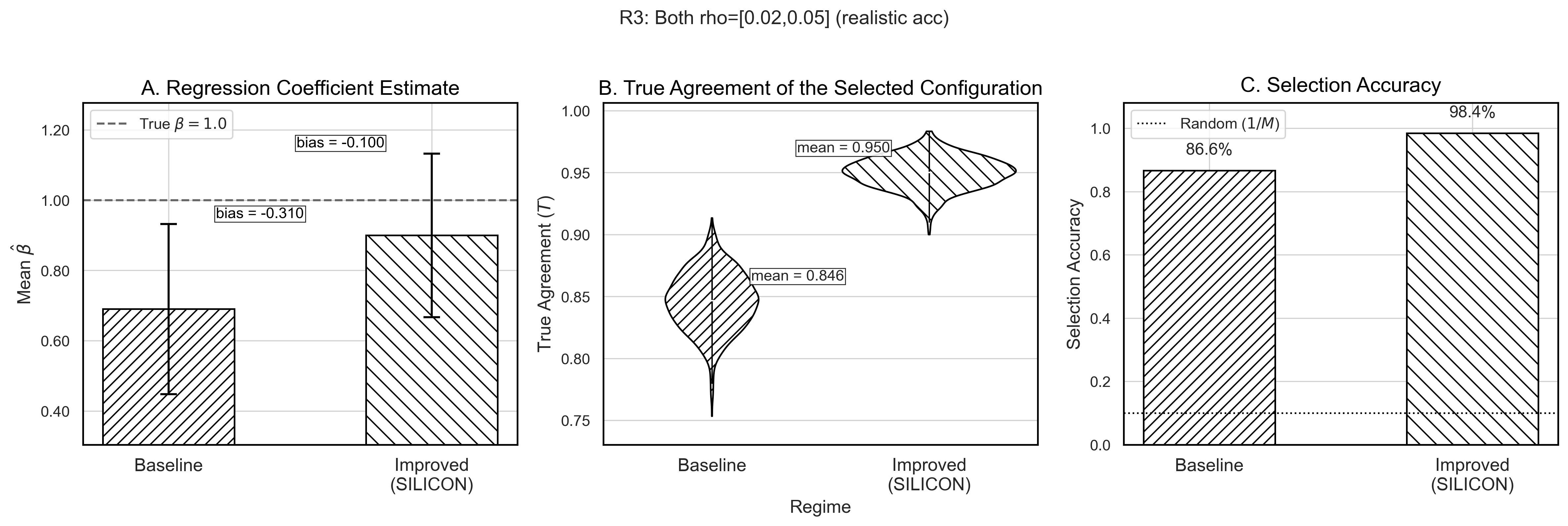}
    \begin{flushleft}
        \footnotesize \parbox{1.0\linewidth}{
        \textit{Notes.} Same format as Figure~\ref{fig:simulation_fig1}. Both regimes use $\rho \in [0.02, 0.05]$. Baseline: $e_h = 0.15$, accuracy $\in [0.65, 0.85]$. Improved: $e_h = 0.02$, accuracy $\in [0.75, 0.95]$.
        }
    \end{flushleft}
    \label{fig:sensitivity_r3}
\end{figure}

\end{APPENDICES}
		\putbib
	\end{bibunit}


\begin{thebibliography}{76}
\providecommand{\natexlab}[1]{#1}
\providecommand{\url}[1]{\texttt{#1}}
\providecommand{\urlprefix}{URL }

\bibitem[{Banerjee et~al.(2021)Banerjee, Dellarocas, \protect\BIBand{} Zervas}]{banerjee_interacting_2021}
Banerjee S, Dellarocas C, Zervas G (2021) Interacting {User}-{Generated} {Content} {Technologies}: {How} {Questions} and {Answers} {Affect} {Consumer} {Reviews}. \emph{Journal of Marketing Research} 58(4):742--761.

\bibitem[{Calderon et~al.(2025)Calderon, Reichart, \protect\BIBand{} Dror}]{calderon_alternative_2025}
Calderon N, Reichart R, Dror R (2025) The {Alternative} {Annotator} {Test} for {LLM}-as-a-{Judge}: {How} to {Statistically} {Justify} {Replacing} {Human} {Annotators} with {LLMs}. Che W, Nabende J, Shutova E, Pilehvar MT, eds., \emph{Proceedings of the 63rd {Annual} {Meeting} of the {Association} for {Computational} {Linguistics} ({Volume} 1: {Long} {Papers})}, 16051--16081 (Vienna, Austria: Association for Computational Linguistics), ISBN 979-8-89176-251-0.

\bibitem[{Cao et~al.(2019)Cao, Liang, \protect\BIBand{} Zhan}]{cao_peer_2019}
Cao J, Liang H, Zhan X (2019) Peer {Effects} of {Corporate} {Social} {Responsibility}. \emph{Management Science} 65(12):5487--5503.

\bibitem[{Carlson \protect\BIBand{} Burbano(2025)}]{carlson_use_2025}
Carlson NA, Burbano V (2025) The {Use} of {LLMs} to {Annotate} {Data} in {Management} {Research}: {Foundational} {Guidelines} and {Warnings}. \emph{Strategic Management Journal} forthcoming.

\bibitem[{Chakraborty et~al.(2022)Chakraborty, Kim, \protect\BIBand{} Sudhir}]{chakraborty_attribute_2022}
Chakraborty I, Kim M, Sudhir K (2022) Attribute sentiment scoring with online text reviews: {Accounting} for language structure and missing attributes. \emph{Journal of Marketing Research} 59(3):600--622.

\bibitem[{Chen et~al.(2023)Chen, Zaharia, \protect\BIBand{} Zou}]{chen_how_2023}
Chen L, Zaharia M, Zou J (2023) How is {ChatGPT}'s behavior changing over time?

\bibitem[{Chen \protect\BIBand{} Chan(2024)}]{chen_large_2024}
Chen Z, Chan J (2024) Large {Language} {Model} in {Creative} {Work}: {The} {Role} of {Collaboration} {Modality} and {User} {Expertise}. \emph{Management Science} 70(12):9101--9117.

\bibitem[{Cheng et~al.(2024)Cheng, Mayya, Shi, \protect\BIBand{} Ye}]{cheng_support_2024}
Cheng X, Mayya R, Shi L, Ye S (2024) Support or {Setback}? {Unpacking} the {Impact} of the {Small} {Business} {Badge} .

\bibitem[{Choudhury et~al.(2019)Choudhury, Wang, Carlson, \protect\BIBand{} Khanna}]{choudhury_machine_2019}
Choudhury P, Wang D, Carlson NA, Khanna T (2019) Machine learning approaches to facial and text analysis: {Discovering} {CEO} oral communication styles. \emph{Strategic Management Journal} 40(11):1705--1732.

\bibitem[{de~Kok(2025)}]{de_kok_chatgpt_2025}
de~Kok T (2025) {ChatGPT} for {Textual} {Analysis}? {How} to {Use} {Generative} {LLMs} in {Accounting} {Research}. \emph{Management Science} 71(9):7888--7906.

\bibitem[{Deng \protect\BIBand{} Ravichandran(2023)}]{deng_managerial_2023}
Deng C, Ravichandran T (2023) Managerial {Response} to {Online} {Positive} {Reviews}: {Helpful} or {Harmful}? \emph{Information Systems Research} 35(4):1802--1823.

\bibitem[{Doshi et~al.(2024)Doshi, Bell, Mirzayev, \protect\BIBand{} Vanneste}]{doshi_generative_2024}
Doshi AR, Bell JJ, Mirzayev E, Vanneste BS (2024) Generative artificial intelligence and evaluating strategic decisions. \emph{Strategic Management Journal} smj.3677.

\bibitem[{Fišar et~al.(2024)Fišar, Greiner, Huber, Katok, Ozkes, \protect\BIBand{} {and the Management Science Reproducibility Collaboration}}]{fisar_reproducibility_2024}
Fišar M, Greiner B, Huber C, Katok E, Ozkes AI, {and the Management Science Reproducibility Collaboration} (2024) Reproducibility in {Management} {Science}. \emph{Management Science} 70(3):1343--1356.

\bibitem[{Ghose \protect\BIBand{} Ipeirotis(2011)}]{ghose_estimating_2011}
Ghose A, Ipeirotis PG (2011) Estimating the {Helpfulness} and {Economic} {Impact} of {Product} {Reviews}: {Mining} {Text} and {Reviewer} {Characteristics}. \emph{IEEE Transactions on Knowledge and Data Engineering} 23(10):1498--1512.

\bibitem[{Gilardi et~al.(2023)Gilardi, Alizadeh, \protect\BIBand{} Kubli}]{gilardi_chatgpt_2023}
Gilardi F, Alizadeh M, Kubli M (2023) {ChatGPT} {Outperforms} {Crowd} {Workers} for {Text}-{Annotation} {Tasks}. \emph{Proceedings of the National Academy of Sciences} 120(30):e2305016120.

\bibitem[{Goli \protect\BIBand{} Singh(2024)}]{goli_frontiers_2024}
Goli A, Singh A (2024) Frontiers: {Can} {Large} {Language} {Models} {Capture} {Human} {Preferences}? \emph{Marketing Science} 43(4):709--722.

\bibitem[{Gregor \protect\BIBand{} Hevner(2013)}]{gregor_positioning_2013}
Gregor S, Hevner AR (2013) Positioning and {Presenting} {Design} {Science} {Research} for {Maximum} {Impact1}. \emph{MIS Quarterly} 37(2):337--355.

\bibitem[{Hashemi et~al.(2024)Hashemi, Eisner, Rosset, Van~Durme, \protect\BIBand{} Kedzie}]{hashemi_llm-rubric_2024}
Hashemi H, Eisner J, Rosset C, Van~Durme B, Kedzie C (2024) {LLM}-{Rubric}: {A} {Multidimensional}, {Calibrated} {Approach} to {Automated} {Evaluation} of {Natural} {Language} {Texts}. Ku LW, Martins A, Srikumar V, eds., \emph{Proceedings of the 62nd {Annual} {Meeting} of the {Association} for {Computational} {Linguistics} ({Volume} 1: {Long} {Papers})}, 13806--13834 (Bangkok, Thailand: Association for Computational Linguistics).

\bibitem[{Higashinaka et~al.(2016)Higashinaka, Funakoshi, Kobayashi, \protect\BIBand{} Inaba}]{higashinaka_dialogue_2016}
Higashinaka R, Funakoshi K, Kobayashi Y, Inaba M (2016) The dialogue breakdown detection challenge: {Task} description, datasets, and evaluation metrics. Calzolari N, Choukri K, Declerck T, Goggi S, Grobelnik M, Maegaard B, Mariani J, Mazo H, Moreno A, Odijk J, Piperidis S, eds., \emph{Proceedings of the {Tenth} {International} {Conference} on {Language} {Resources} and {Evaluation} ({LREC}'16)}, 3146--3150 (Portorož, Slovenia: European Language Resources Association (ELRA)).

\bibitem[{Homburg et~al.(2015)Homburg, Ehm, \protect\BIBand{} Artz}]{homburg_measuring_2015}
Homburg C, Ehm L, Artz M (2015) Measuring and {Managing} {Consumer} {Sentiment} in an {Online} {Community} {Environment}. \emph{Journal of Marketing Research} 52(5):629--641.

\bibitem[{Hong et~al.(2021)Hong, Peng, Burtch, \protect\BIBand{} Huang}]{hong_just_2021}
Hong Y, Peng J, Burtch G, Huang N (2021) Just {DM} {Me} ({Politely}): {Direct} {Messaging}, {Politeness}, and {Hiring} {Outcomes} in {Online} {Labor} {Markets}. \emph{Information Systems Research} 32(3):786--800.

\bibitem[{Hu \protect\BIBand{} Collier(2024)}]{hu_quantifying_2024}
Hu T, Collier N (2024) Quantifying the {Persona} {Effect} in {LLM} {Simulations}. Ku LW, Martins A, Srikumar V, eds., \emph{Proceedings of the 62nd {Annual} {Meeting} of the {Association} for {Computational} {Linguistics} ({Volume} 1: {Long} {Papers})}, 10289--10307 (Bangkok, Thailand: Association for Computational Linguistics).

\bibitem[{Huang et~al.(2024)Huang, Huang, \protect\BIBand{} Hong}]{huang_algorithm-enabled_2024}
Huang A, Huang N, Hong Y (2024) Algorithm-enabled {Workflow} {Automation} in {Open}-source {Development}: {Evidence} from {GitHub} {Actions}. \emph{SSRN Electronic Journal} .

\bibitem[{Humphreys \protect\BIBand{} Wang(2018)}]{humphreys_automated_2018}
Humphreys A, Wang RJH (2018) Automated {Text} {Analysis} for {Consumer} {Research}. \emph{Journal of Consumer Research} 44(6):1274--1306.

\bibitem[{Ipeirotis et~al.(2014)Ipeirotis, Provost, Sheng, \protect\BIBand{} Wang}]{ipeirotis_repeated_2014}
Ipeirotis PG, Provost F, Sheng VS, Wang J (2014) Repeated labeling using multiple noisy labelers. \emph{Data Mining and Knowledge Discovery} 28(2):402--441.

\bibitem[{Krippendorff(2004)}]{krippendorff_reliability_2004}
Krippendorff K (2004) Reliability in {Content} {Analysis}: {Some} {Common} {Misconceptions} and {Recommendations}. \emph{Human Communication Research} 30(3):411--433.

\bibitem[{Kwark et~al.(2021)Kwark, Lee, Pavlou, \protect\BIBand{} Qiu}]{kwark_spillover_2021}
Kwark Y, Lee GM, Pavlou PA, Qiu L (2021) On the {Spillover} {Effects} of {Online} {Product} {Reviews} on {Purchases}: {Evidence} from {Clickstream} {Data}. \emph{Information Systems Research} 32(3):895--913.

\bibitem[{Landis \protect\BIBand{} Koch(1977)}]{landis_measurement_1977}
Landis JR, Koch GG (1977) The {Measurement} of {Observer} {Agreement} for {Categorical} {Data}. \emph{Biometrics} 33(1):159.

\bibitem[{Lee \protect\BIBand{} Hosanagar(2021)}]{lee_how_2021}
Lee D, Hosanagar K (2021) How {Do} {Product} {Attributes} and {Reviews} {Moderate} the {Impact} of {Recommender} {Systems} {Through} {Purchase} {Stages}? \emph{Management Science} 67(1):524--546.

\bibitem[{Lee et~al.(2018)Lee, Hosanagar, \protect\BIBand{} Nair}]{lee_advertising_2018}
Lee D, Hosanagar K, Nair HS (2018) Advertising {Content} and {Consumer} {Engagement} on {Social} {Media}: {Evidence} from {Facebook}. \emph{Management Science} 64(11):5105--5131.

\bibitem[{Lee et~al.(2023)Lee, DeLucia, Nangia, Ganedi, Guan, Li, Ngaw, Singhal, Vaidya, Yuan, Zhang, \protect\BIBand{} Sedoc}]{lee_common_2023}
Lee S, DeLucia A, Nangia N, Ganedi P, Guan R, Li R, Ngaw B, Singhal A, Vaidya S, Yuan Z, Zhang L, Sedoc J (2023) Common {Law} {Annotations}: {Investigating} the {Stability} of {Dialog} {System} {Output} {Annotations}. Rogers A, Boyd-Graber J, Okazaki N, eds., \emph{Findings of the {Association} for {Computational} {Linguistics}: {ACL} 2023}, 12315--12349 (Toronto, Canada: Association for Computational Linguistics).

\bibitem[{Leek et~al.(2024)Leek, Bischl, \protect\BIBand{} Freier}]{leek_introducing_2024}
Leek LC, Bischl S, Freier M (2024) Introducing {Textual} {Measures} of {Central} {Bank} {Policy}-{Linkages} {Using} {ChatGPT}.

\bibitem[{Liu et~al.(2019)Liu, Lee, \protect\BIBand{} Srinivasan}]{liu_large-scale_2019}
Liu X, Lee D, Srinivasan K (2019) Large-{Scale} {Cross}-{Category} {Analysis} of {Consumer} {Review} {Content} on {Sales} {Conversion} {Leveraging} {Deep} {Learning}. \emph{Journal of Marketing Research} 56(6):918--943.

\bibitem[{Loughran \protect\BIBand{} Mcdonald(2014)}]{loughran_measuring_2014}
Loughran T, Mcdonald B (2014) Measuring {Readability} in {Financial} {Disclosures}. \emph{The Journal of Finance} 69(4):1643--1671.

\bibitem[{Lyu et~al.(2025)Lyu, Shridhar, Malaviya, Zhang, Elazar, Tandon, Apidianaki, Sachan, \protect\BIBand{} Callison-Burch}]{lyu_calibrating_2025}
Lyu Q, Shridhar K, Malaviya C, Zhang L, Elazar Y, Tandon N, Apidianaki M, Sachan M, Callison-Burch C (2025) Calibrating {Large} {Language} {Models} with {Sample} {Consistency}. \emph{Proceedings of the AAAI Conference on Artificial Intelligence} 39(18):19260--19268.

\bibitem[{Matook et~al.(2022)Matook, Dennis, \protect\BIBand{} Wang}]{matook_user_2022}
Matook S, Dennis AR, Wang YM (2022) User {Comments} in {Social} {Media} {Firestorms}: {A} {Mixed}-{Method} {Study} of {Purpose}, {Tone}, and {Motivation}. \emph{Journal of Management Information Systems} 39(3):673--705.

\bibitem[{Mayya et~al.(2021)Mayya, Ye, Viswanathan, \protect\BIBand{} Agarwal}]{mayya_who_2021}
Mayya R, Ye S, Viswanathan S, Agarwal R (2021) Who {Forgoes} {Screening} in {Online} {Markets} and {Why}? {Evidence} from {Airbnb}. \emph{MIS Quarterly} 45(4):1745--1776.

\bibitem[{Melumad et~al.(2019)Melumad, Inman, \protect\BIBand{} Pham}]{melumad_selectively_2019}
Melumad S, Inman JJ, Pham MT (2019) Selectively {Emotional}: {How} {Smartphone} {Use} {Changes} {User}-{Generated} {Content}. \emph{Journal of Marketing Research} 56(2):259--275.

\bibitem[{Mousavi et~al.(2024)Mousavi, Kitchens, Oliver, \protect\BIBand{} Abbasi}]{mousavi_lexicons_2024}
Mousavi R, Kitchens B, Oliver A, Abbasi A (2024) From {Lexicons} to {Large} {Language} {Models}: {A} {Holistic} {Evaluation} of {Psychometric} {Text} {Analysis} in {Social} {Science} {Research}.

\bibitem[{Natarajan et~al.(2013)Natarajan, Dhillon, Ravikumar, \protect\BIBand{} Tewari}]{natarajan_learning_2013}
Natarajan N, Dhillon IS, Ravikumar PK, Tewari A (2013) Learning with {Noisy} {Labels}. Burges CJ, Bottou L, Welling M, Ghahramani Z, Weinberger KQ, eds., \emph{Advances in {Neural} {Information} {Processing} {Systems}}, volume~26 (Curran Associates, Inc.).

\bibitem[{{National Academies of Sciences, Engineering, and Medicine}(2019)}]{national_academies_of_sciences_engineering_and_medicine_reproducibility_2019}
{National Academies of Sciences, Engineering, and Medicine} (2019) \emph{Reproducibility and {Replicability} in {Science}} (Washington, DC: The National Academies Press), ISBN 978-0-309-48616-3.

\bibitem[{Nickerson(1998)}]{nickerson_confirmation_1998}
Nickerson RS (1998) Confirmation {Bias}: {A} {Ubiquitous} {Phenomenon} in {Many} {Guises}. \emph{Review of General Psychology} 2(2):175--220.

\bibitem[{Obermeier \protect\BIBand{} Mayya(2024)}]{obermeier_decentralized_2024}
Obermeier D, Mayya R (2024) Decentralized {Governance} as an {Open} {Platform} {Strategy}: {An} {Empirical} {Study} of {Devolving} {Controls} through {DAOs} .

\bibitem[{Oh et~al.(2023)Oh, Goh, \protect\BIBand{} Phan}]{oh_are_2023}
Oh H, Goh KY, Phan TQ (2023) Are {You} {What} {You} {Tweet}? {The} {Impact} of {Sentiment} on {Digital} {News} {Consumption} and {Social} {Media} {Sharing}. \emph{Information Systems Research} 34(1):111--136.

\bibitem[{Oortwijn et~al.(2021)Oortwijn, Ossenkoppele, \protect\BIBand{} Betti}]{oortwijn_interrater_2021}
Oortwijn Y, Ossenkoppele T, Betti A (2021) Interrater {Disagreement} {Resolution}: {A} {Systematic} {Procedure} to {Reach} {Consensus} in {Annotation} {Tasks}. Belz A, Agarwal S, Graham Y, Reiter E, Shimorina A, eds., \emph{Proceedings of the {Workshop} on {Human} {Evaluation} of {NLP} {Systems} ({HumEval})}, 131--141 (Online: Association for Computational Linguistics).

\bibitem[{Paharia et~al.(2011)Paharia, Keinan, Avery, \protect\BIBand{} Schor}]{paharia_underdog_2011}
Paharia N, Keinan A, Avery J, Schor JB (2011) The {Underdog} {Effect}: {The} {Marketing} of {Disadvantage} and {Determination} through {Brand} {Biography}. \emph{Journal of Consumer Research} 37(5):775--790.

\bibitem[{Palmer et~al.(2023)Palmer, Smith, \protect\BIBand{} Spirling}]{palmer_using_2023}
Palmer A, Smith NA, Spirling A (2023) Using proprietary language models in academic research requires explicit justification. \emph{Nature Computational Science} 4(1):2--3.

\bibitem[{Pangakis et~al.(2023)Pangakis, Wolken, \protect\BIBand{} Fasching}]{pangakis_automated_2023}
Pangakis N, Wolken S, Fasching N (2023) Automated {Annotation} with {Generative} {AI} {Requires} {Validation} .

\bibitem[{Patterson \protect\BIBand{} Eggleston(2017)}]{patterson_intuitive_2017}
Patterson RE, Eggleston RG (2017) Intuitive {Cognition}. \emph{Journal of Cognitive Engineering and Decision Making} 11(1):5--22.

\bibitem[{Peng et~al.(2022)Peng, Romero, \protect\BIBand{} Horvat}]{peng_dynamics_2022}
Peng H, Romero DM, Horvat EA (2022) Dynamics of cross-platform attention to retracted papers. \emph{Proceedings of the National Academy of Sciences} 119(25):e2119086119.

\bibitem[{Pryzant et~al.(2023)Pryzant, Iter, Li, Lee, Zhu, \protect\BIBand{} Zeng}]{pryzant_automatic_2023}
Pryzant R, Iter D, Li J, Lee Y, Zhu C, Zeng M (2023) Automatic {Prompt} {Optimization} with “{Gradient} {Descent}” and {Beam} {Search}. Bouamor H, Pino J, Bali K, eds., \emph{Proceedings of the 2023 {Conference} on {Empirical} {Methods} in {Natural} {Language} {Processing}}, 7957--7968 (Singapore: Association for Computational Linguistics).

\bibitem[{Rathje et~al.(2024)Rathje, Mirea, Sucholutsky, Marjieh, Robertson, \protect\BIBand{} Van~Bavel}]{rathje_gpt_2024}
Rathje S, Mirea DM, Sucholutsky I, Marjieh R, Robertson C, Van~Bavel JJ (2024) {GPT} is an effective tool for multilingual psychological text analysis. \emph{Proceedings of the National Academy of Sciences} 121(34):121 (34) e2308950121.

\bibitem[{Roper et~al.(2022)Roper, Abdel-Rehim, Hubbard, Carpenter, Rzhetsky, Soldatova, \protect\BIBand{} King}]{roper_testing_2022}
Roper K, Abdel-Rehim A, Hubbard S, Carpenter M, Rzhetsky A, Soldatova L, King RD (2022) Testing the reproducibility and robustness of the cancer biology literature by robot. \emph{Journal of the Royal Society Interface} 19(189):20210821.

\bibitem[{Rosette et~al.(2013)Rosette, Carton, Bowes-Sperry, \protect\BIBand{} Hewlin}]{rosette_why_2013}
Rosette AS, Carton AM, Bowes-Sperry L, Hewlin PF (2013) Why {Do} {Racial} {Slurs} {Remain} {Prevalent} in the {Workplace}? {Integrating} {Theory} on {Intergroup} {Behavior}. \emph{Organization Science} 24(5):1402--1421.

\bibitem[{Rottger et~al.(2022)Rottger, Vidgen, Hovy, \protect\BIBand{} Pierrehumbert}]{rottger_two_2022}
Rottger P, Vidgen B, Hovy D, Pierrehumbert J (2022) Two {Contrasting} {Data} {Annotation} {Paradigms} for {Subjective} {NLP} {Tasks}. Carpuat M, de~Marneffe MC, Meza~Ruiz IV, eds., \emph{Proceedings of the 2022 {Conference} of the {North} {American} {Chapter} of the {Association} for {Computational} {Linguistics}: {Human} {Language} {Technologies}}, 175--190 (Seattle, United States: Association for Computational Linguistics).

\bibitem[{Ruggeri et~al.(2024)Ruggeri, Misino, Muti, Korre, Torroni, \protect\BIBand{} Barrón-Cedeño}]{ruggeri_let_2024}
Ruggeri F, Misino E, Muti A, Korre K, Torroni P, Barrón-Cedeño A (2024) Let {Guidelines} {Guide} {You}: {A} {Prescriptive} {Guideline}-{Centered} {Data} {Annotation} {Methodology}.

\bibitem[{Saha et~al.(2023)Saha, Garimella, Kalyan, Pandey, Meher, Mathew, \protect\BIBand{} Mukherjee}]{saha_rise_2023}
Saha P, Garimella K, Kalyan NK, Pandey SK, Meher PM, Mathew B, Mukherjee A (2023) On the {Rise} of {Fear} {Speech} in {Online} {Social} {Media}. \emph{Proceedings of the National Academy of Sciences} 120(11):e2212270120.

\bibitem[{Salehi et~al.(2017)Salehi, Teevan, Iqbal, \protect\BIBand{} Kamar}]{salehi_communicating_2017}
Salehi N, Teevan J, Iqbal S, Kamar E (2017) Communicating {Context} to the {Crowd} for {Complex} {Writing} {Tasks}. \emph{Proceedings of the 2017 {ACM} {Conference} on {Computer} {Supported} {Cooperative} {Work} and {Social} {Computing}}, 1890--1901, {CSCW} '17 (New York, NY, USA: Association for Computing Machinery), ISBN 978-1-4503-4335-0.

\bibitem[{Sap et~al.(2019)Sap, Card, Gabriel, Choi, \protect\BIBand{} Smith}]{sap_risk_2019}
Sap M, Card D, Gabriel S, Choi Y, Smith NA (2019) The {Risk} of {Racial} {Bias} in {Hate} {Speech} {Detection}. Korhonen A, Traum D, Màrquez L, eds., \emph{Proceedings of the 57th {Annual} {Meeting} of the {Association} for {Computational} {Linguistics}}, 1668--1678 (Florence, Italy: Association for Computational Linguistics).

\bibitem[{Sen et~al.(2020)Sen, Hartvigsen, Yin, Kong, \protect\BIBand{} Rundensteiner}]{sen_human_2020}
Sen C, Hartvigsen T, Yin B, Kong X, Rundensteiner E (2020) Human {Attention} {Maps} for {Text} {Classification}: {Do} {Humans} and {Neural} {Networks} {Focus} on the {Same} {Words}? \emph{Proceedings of the 58th {Annual} {Meeting} of the {Association} for {Computational} {Linguistics}}, 4596--4608 (Online: Association for Computational Linguistics).

\bibitem[{Shi et~al.(2022)Shi, Liu, \protect\BIBand{} Srinivasan}]{shi_hype_2022}
Shi ZJ, Liu X, Srinivasan K (2022) Hype {News} {Diffusion} and {Risk} of {Misinformation}: {The} {Oz} {Effect} in {Health} {Care}. \emph{Journal of Marketing Research} 59(2):327--352.

\bibitem[{Sibai et~al.(2024)Sibai, Luedicke, \protect\BIBand{} De~Valck}]{sibai_why_2024}
Sibai O, Luedicke MK, De~Valck K (2024) Why {Online} {Consumption} {Communities} {Brutalize}. \emph{Journal of Consumer Research} ucae022.

\bibitem[{Snow et~al.(2008)Snow, O'Connor, Jurafsky, \protect\BIBand{} Ng}]{snow_cheap_2008}
Snow R, O'Connor B, Jurafsky D, Ng A (2008) Cheap and {Fast} – {But} is it {Good}? {Evaluating} {Non}-{Expert} {Annotations} for {Natural} {Language} {Tasks}. Lapata M, Ng HT, eds., \emph{Proceedings of the 2008 {Conference} on {Empirical} {Methods} in {Natural} {Language} {Processing}}, 254--263 (Honolulu, Hawaii: Association for Computational Linguistics).

\bibitem[{Stolcke et~al.(2000)Stolcke, Ries, Coccaro, Shriberg, Bates, Jurafsky, Taylor, Martin, Van Ess-Dykema, \protect\BIBand{} Meteer}]{stolcke_dialogue_2000}
Stolcke A, Ries K, Coccaro N, Shriberg E, Bates R, Jurafsky D, Taylor P, Martin R, Van Ess-Dykema C, Meteer M (2000) Dialogue {Act} {Modeling} for {Automatic} {Tagging} and {Recognition} of {Conversational} {Speech}. \emph{Computational Linguistics} 26(3):339--374.

\bibitem[{Tan et~al.(2024)Tan, Li, Wang, Beigi, Jiang, Bhattacharjee, Karami, Li, Cheng, \protect\BIBand{} Liu}]{tan_large_2024}
Tan Z, Li D, Wang S, Beigi A, Jiang B, Bhattacharjee A, Karami M, Li J, Cheng L, Liu H (2024) Large {Language} {Models} for {Data} {Annotation} and {Synthesis}: {A} {Survey}. \emph{Proceedings of the 2024 {Conference} on {Empirical} {Methods} in {Natural} {Language} {Processing}}, 930--957 (Miami, Florida, USA: Association for Computational Linguistics).

\bibitem[{Tang et~al.(2024)Tang, Li, Ding, Gopal, \protect\BIBand{} Zhang}]{tang_racial_2024}
Tang C, Li SK, Ding Y, Gopal RD, Zhang G (2024) Racial {Discrimination} and {Anti}-discrimination: {The} {COVID}-19 {Pandemic}’s {Impact} on {Chinese} {Restaurants} in {North} {America}. \emph{Information Systems Research} 35(3):1274--1295.

\bibitem[{Törnberg(2024)}]{tornberg_best_2024}
Törnberg P (2024) Best {Practices} for {Text} {Annotation} with {Large} {Language} {Models}.

\bibitem[{Wei et~al.(2022)Wei, Wang, Schuurmans, Bosma, Ichter, Xia, Chi, Le, \protect\BIBand{} Zhou}]{wei_chain--thought_2022}
Wei J, Wang X, Schuurmans D, Bosma M, Ichter B, Xia F, Chi EH, Le QV, Zhou D (2022) Chain-of-thought prompting elicits reasoning in large language models. \emph{Proceedings of the 36th {International} {Conference} on {Neural} {Information} {Processing} {Systems}}, 24824--24837, {NIPS} '22 (Red Hook, NY, USA: Curran Associates Inc.), ISBN 978-1-71387-108-8.

\bibitem[{Xie et~al.(2024)Xie, Chen, Jia, Ye, Lai, Shu, Gu, Bibi, Hu, Jurgens, Evans, Torr, Ghanem, \protect\BIBand{} Li}]{xie_can_2024}
Xie C, Chen C, Jia F, Ye Z, Lai S, Shu K, Gu J, Bibi A, Hu Z, Jurgens D, Evans J, Torr P, Ghanem B, Li G (2024) Can {Large} {Language} {Model} {Agents} {Simulate} {Human} {Trust} {Behavior}? (38th Conference on Neural Information Processing Systems).

\bibitem[{Yang et~al.(2018)Yang, Adomavicius, Burtch, \protect\BIBand{} Ren}]{yang_mind_2018}
Yang M, Adomavicius G, Burtch G, Ren Y (2018) Mind the {Gap}: {Accounting} for {Measurement} {Error} and {Misclassification} in {Variables} {Generated} via {Data} {Mining}. \emph{Information Systems Research} 29(1):4--24.

\bibitem[{Yang et~al.(2019)Yang, Ren, \protect\BIBand{} Adomavicius}]{yang_understanding_2019}
Yang M, Ren Y, Adomavicius G (2019) Understanding {User}-{Generated} {Content} and {Customer} {Engagement} on {Facebook} {Business} {Pages}. \emph{Information Systems Research} 30(3):839--855.

\bibitem[{Ye et~al.(2024)Ye, Yoganarasimhan, \protect\BIBand{} Zheng}]{ye_lola_2024}
Ye Z, Yoganarasimhan H, Zheng Y (2024) {LOLA}: {LLM}-{Assisted} {Online} {Learning} {Algorithm} for {Content} {Experiments}.

\bibitem[{Yeverechyahu et~al.(2024)Yeverechyahu, Mayya, \protect\BIBand{} Oestreicher-Singer}]{yeverechyahu_impact_2024}
Yeverechyahu D, Mayya R, Oestreicher-Singer G (2024) The {Impact} of {Large} {Language} {Models} on {Open}-{Source} {Innovation}: {Evidence} from {GitHub} {Copilot}. \emph{SSRN Electronic Journal} .

\bibitem[{Zhang et~al.(2024)Zhang, D'Haro, Chen, Zhang, \protect\BIBand{} Li}]{zhang_comprehensive_2024}
Zhang C, D'Haro LF, Chen Y, Zhang M, Li H (2024) A {Comprehensive} {Analysis} of the {Effectiveness} of {Large} {Language} {Models} as {Automatic} {Dialogue} {Evaluators}.

\bibitem[{Zhang et~al.(2016)Zhang, Bhattacharyya, \protect\BIBand{} Ram}]{zhang_large-scale_2016}
Zhang K, Bhattacharyya S, Ram S (2016) Large-{Scale} {Network} {Analysis} for {Online} {Social} {Brand} {Advertising}. \emph{MIS Quarterly} 40(4):849--868.

\bibitem[{Zhang et~al.(2023)Zhang, Mille, Hou, Deutsch, Clark, Liu, Mahamood, Gehrmann, Clinciu, Chandu, \protect\BIBand{} Sedoc}]{zhang_needle_2023}
Zhang L, Mille S, Hou Y, Deutsch D, Clark E, Liu Y, Mahamood S, Gehrmann S, Clinciu M, Chandu KR, Sedoc J (2023) A {Needle} in a {Haystack}: {An} {Analysis} of {High}-{Agreement} {Workers} on {MTurk} for {Summarization}. \emph{Proceedings of the 61st {Annual} {Meeting} of the {Association} for {Computational} {Linguistics} ({Volume} 1: {Long} {Papers})}, 14944--14982 (Toronto, Canada: Association for Computational Linguistics).

\end{thebibliography}


\begin{thebibliography}{8}
\providecommand{\natexlab}[1]{#1}
\providecommand{\url}[1]{\texttt{#1}}
\providecommand{\urlprefix}{URL }

\bibitem[{Cohen(1968)}]{cohen_weighted_1968}
Cohen J (1968) Weighted kappa: {Nominal} scale agreement provision for scaled disagreement or partial credit. \emph{Psychological Bulletin} 70(4):213--220.

\bibitem[{Jaccard(1908)}]{jaccard_nouvelles_1908}
Jaccard P (1908) Nouvelles recherches sur la distribution florale. \emph{Bull. Soc. Vaud. Sci. Nat.} 44:223--270.

\bibitem[{Landis \protect\BIBand{} Koch(1977)}]{landis_measurement_1977}
Landis JR, Koch GG (1977) The {Measurement} of {Observer} {Agreement} for {Categorical} {Data}. \emph{Biometrics} 33(1):159.

\bibitem[{Lee et~al.(2023)Lee, DeLucia, Nangia, Ganedi, Guan, Li, Ngaw, Singhal, Vaidya, Yuan, Zhang, \protect\BIBand{} Sedoc}]{lee_common_2023}
Lee S, DeLucia A, Nangia N, Ganedi P, Guan R, Li R, Ngaw B, Singhal A, Vaidya S, Yuan Z, Zhang L, Sedoc J (2023) Common {Law} {Annotations}: {Investigating} the {Stability} of {Dialog} {System} {Output} {Annotations}. Rogers A, Boyd-Graber J, Okazaki N, eds., \emph{Findings of the {Association} for {Computational} {Linguistics}: {ACL} 2023}, 12315--12349 (Toronto, Canada: Association for Computational Linguistics).

\bibitem[{Pangakis et~al.(2023)Pangakis, Wolken, \protect\BIBand{} Fasching}]{pangakis_automated_2023}
Pangakis N, Wolken S, Fasching N (2023) Automated {Annotation} with {Generative} {AI} {Requires} {Validation} .

\bibitem[{Papke \protect\BIBand{} Wooldridge(1996)}]{papke_econometric_1996}
Papke LE, Wooldridge JM (1996) Econometric methods for fractional response variables with an application to 401(k) plan participation rates. \emph{Journal of Applied Econometrics} 11(6):619--632.

\bibitem[{Passonneau(2006)}]{passonneau_measuring_2006}
Passonneau R (2006) Measuring agreement on set-valued items ({MASI}) for semantic and pragmatic annotation. Calzolari N, Choukri K, Gangemi A, Maegaard B, Mariani J, Odijk J, Tapias D, eds., \emph{Proceedings of the fifth international conference on language resources and evaluation ({LREC}'06)} (Genoa, Italy: European Language Resources Association (ELRA)).

\bibitem[{Peng et~al.(2022)Peng, Romero, \protect\BIBand{} Horvat}]{peng_dynamics_2022}
Peng H, Romero DM, Horvat EA (2022) Dynamics of cross-platform attention to retracted papers. \emph{Proceedings of the National Academy of Sciences} 119(25):e2119086119.

\end{thebibliography}
\end{document}